\makeatletter
\let\my@xfloat\@xfloat
\makeatother

\documentclass{uicthesi}

\makeatletter
\def\@xfloat#1[#2]{
	\my@xfloat#1[#2]%
	\def\baselinestretch{1}%
	\@normalsize \normalsize
}
\makeatother

\usepackage{amsmath}
\usepackage{amssymb}
\usepackage{booktabs}
\usepackage{graphicx}
\usepackage{array}
\usepackage{url}
\usepackage{enumitem}
\usepackage{subfigure}
\usepackage{multirow}
\usepackage{longtable}
\usepackage{pdfpages}
\usepackage[bookmarks=false]{hyperref}
\usepackage{algorithm}
\usepackage{algorithmic}

\newtheorem{Theorem}{\textsc{Theorem}}
\newtheorem{Definition}{\textsc{Definition}}

\newcommand{\mvfs}{\texttt{tMVFS}}
\newcommand{\miqp}{\texttt{MIQP}}
\newcommand{\mkl}{\texttt{MKL}}
\newcommand{\trfe}{\texttt{STM-RFE}}
\newcommand{\vrfe}{\texttt{SVM-RFE}}
\newcommand{\tf}{\texttt{STM}}
\newcommand{\cf}{\texttt{SVM}}
\newcommand{\stm}{\texttt{STM}}
\newcommand{\svm}{\texttt{SVM}}

\newcommand{\gscore}{\texttt{gSide}}
\newcommand{\gmsv}{\texttt{gMSV}}
\newcommand{\gssc}{\texttt{gSSC}}
\newcommand{\duga}{\texttt{Conf}}
\newcommand{\dugb}{\texttt{Ratio}}
\newcommand{\dugc}{\texttt{Gtest}}
\newcommand{\dugd}{\texttt{HSIC}}
\newcommand{\topk}{\texttt{Freq}}
\newcommand{\side}{\texttt{MSV}}
\newcommand{\gspan}{\texttt{gSpan}}

\newcommand{\bne}{\texttt{tBNE}}
\newcommand{\cmtf}{\texttt{CMTF}}
\newcommand{\rubik}{\texttt{Rubik}}
\newcommand{\als}{\texttt{ALS}}
\newcommand{\cc}{\texttt{CC}}

\newcommand{\deepmood}{\texttt{DeepMood}}
\newcommand{\mvm}{\texttt{MVM}}
\newcommand{\fm}{\texttt{FM}}
\newcommand{\fc}{\texttt{FC}}
\newcommand{\dmvm}{\texttt{DeepMood-MVM}}
\newcommand{\dfm}{\texttt{DeepMood-FM}}
\newcommand{\dnn}{\texttt{DeepMood-FC}}
\newcommand{\xgb}{\texttt{XGBoost}}
\newcommand{\lr}{\texttt{LR}}
\newcommand{\rnn}{\texttt{RNN}}
\newcommand{\gru}{\texttt{GRU}}
\newcommand{\lstm}{\texttt{LSTM}}

\newcommand{\neu}{\emph{neuropsychological tests}}
\newcommand{\flo}{\emph{flow cytometry}}
\newcommand{\pla}{\emph{plasma luminex}}
\newcommand{\fre}{\emph{freesurfer}}
\newcommand{\ave}{\emph{overall brain microstructure}}
\newcommand{\loc}{\emph{localized brain microstructure}}
\newcommand{\seg}{\emph{brain volumetry}}

\newcommand{\dtsa}{\texttt{D2.1}}
\newcommand{\dtsb}{\texttt{D2.2}}
\newcommand{\dtsc}{\texttt{D3.1}}
\newcommand{\dtsd}{\texttt{D3.2}}
\newcommand{\dtse}{\texttt{D4.1}}
\newcommand{\dtsf}{\texttt{D4.2}}
\newcommand{\dtsg}{\texttt{D5.1}}
\newcommand{\dtsh}{\texttt{D5.2}}
\newcommand{\dtsi}{\texttt{D6.1}}
\newcommand{\dtsj}{\texttt{D6.2}}

\newcommand{\fmri}{\texttt{fMRI}}
\newcommand{\dti}{\texttt{DTI}}
\newcommand{\eeg}{\texttt{EEG}}

\newcommand{\nt}{\texttt{Neutral}}
\newcommand{\mt}{\texttt{Maintain}}
\newcommand{\rp}{\texttt{Reappraise}}

\newcommand{\ch}{\texttt{Alphanum}}
\newcommand{\nonch}{\texttt{Special}}
\newcommand{\accel}{\texttt{Accel}}

\newcommand{\hdrs}{\texttt{HDRS}}
\newcommand{\ymrs}{\texttt{YMRS}}

\newcommand{\libsvm}{\texttt{LIBSVM}}
\newcommand{\dparsf}{\texttt{DPARSF}}
\newcommand{\fsl}{\texttt{FSL}}
\newcommand{\aal}{\texttt{AAL}}
\newcommand{\eeglab}{\texttt{EEGLAB}}
\newcommand{\keras}{\texttt{Keras}}
\newcommand{\tensorflow}{\texttt{Tensorflow}}
\newcommand{\scikit}{\texttt{scikit-learn}}
\newcommand{\github}{\texttt{GitHub}}

\newcommand{\hiv}{\texttt{HIV}}
\newcommand{\aids}{\texttt{AIDS}}
\newcommand{\rbf}{\texttt{RBF}}
\newcommand{\ms}{\texttt{min\_sup}}
\newcommand{\dfs}{\texttt{DFS}}
\newcommand{\cp}{\texttt{CP}}
\newcommand{\admm}{\texttt{ADMM}}
\newcommand{\ccdf}{\texttt{CCDF}}
\newcommand{\rmsprop}{\texttt{RMSProp}}
\newcommand{\biaffect}{\texttt{BiAffect}}
\newcommand{\fone}{$\texttt{F}_1$}
\newcommand{\rmse}{\texttt{RMSE}}

\begin{document}


\title{Broad Learning for Healthcare}
\author{BOKAI CAO}
\pdegrees{B.E. and B.S., Renmin University of China, 2013}
\degree{Doctor of Philosophy in Computer Science}
\committee{
~~~~~~~~~~~~~~~~~~Philip S. Yu, Chair and Advisor \\
~~~~~~~~~~~~~~~~~~Bing Liu \\
~~~~~~~~~~~~~~~~~~Piotr Gmytrasiewicz \\
~~~~~~~~~~~~~~~~~~Alex D. Leow, Psychiatry \\
~~~~~~~~~~~~~~~~~~Olusola Ajilore, Psychiatry
}
\maketitle

\dedication
{\null\vfil
{\large
\begin{center}
This thesis is proudly dedicated\\\vspace{12pt}
to my beloved parents and grandparents.
\end{center}}
\vfil\null}

\acknowledgements
{
Firstly, I would like to express my sincere gratitude to my advisor Prof. Philip S. Yu for his patience, motivation, and immense knowledge. It would not be possible to conduct my doctoral study and related research without his continuous support and precious guidance.

Besides my advisor, I would like to thank the rest of my thesis committee: Prof. Bing Liu, Prof. Piotr Gmytrasiewicz, Dr. Alex D. Leow, and Dr. Olusola Ajilore, for their insightful comments and constructive suggestions which encouraged me to widen my research from various perspectives.

My sincere thanks also goes to Prof. Hongyan Liu at Tsinghua University, Prof. Jun He, Prof. Deying Li at Renmin University of China, and Dr. Ann B. Ragin at Northwestern University, who mentored me during the early stage of my research career, and Dr. Francine Chen, Dr. Dhiraj Joshi, Dr. Hucheng Zhou, Dr. Mia Mao, and Dr. Hauzhong Ning, who provided me with the opportunity to join their team as an intern.

I thank my colleagues and friends that I met in UIC for inspiring discussions and for all the happy time we have spent together in the last five years. In particular, I am grateful to Prof. Xiangnan Kong for enlightening me on the first glance of research.

Last but not the least, I would like to thank my family for the unconditional love and always supporting me spiritually throughout my doctoral study and my life in general.
\begin{flushright}
BC
\end{flushright}
}

\contributionofauthors

Chapter \ref{chapter:mvfs} presents a published manuscript \cite{cao2014tensor} for which I was the primary author. Dr. Lifang He contributed to optimization techniques and drafting a part of the manuscript. Prof. Xiangnan Kong, Prof. Philip S. Yu, Prof. Zhifeng Hao, and Dr. Ann B. Ragin contributed to discussions with respect to the work and revising the manuscript. 

Chapter \ref{chapter:subgraph} presents a published manuscript \cite{cao2015mining} for which I was the primary author. Prof. Xiangnan Kong, Dr. Jingyuan Zhang, Prof. Philip S. Yu, and Dr. Ann B. Ragin contributed to discussions with respect to the work and revising the manuscript. 

Chapter \ref{chapter:bne} presents a published manuscript \cite{cao2017tbne} for which I was the primary author. Dr. Lifang He contributed to optimization techniques. Mengqi Xing contributed to data preprocessing. Dr. Xiaokai Wei, Prof. Philip S. Yu, Dr. Heide Klumpp, and Dr. Alex D. Leow contributed to discussions with respect to the work and revising the manuscript. 

Chapter \ref{chapter:deepmood} presents a published manuscript \cite{cao2017deepmood} for which I was the primary author. Lei Zheng contributed to experiments. Andrea Piscitello contributed to software development. Chenwei Zhang, Prof. Philip S. Yu, Dr. John Zulueta, Dr. Olusola Ajilore, Dr. Kelly Ryan, and Dr. Alex D. Leow contributed to discussions with respect to the work and revising the manuscript. 

\tableofcontents
\listoftables
\listoffigures
\listofabbreviations
\begin{list}
{}
{\setlength
   {\labelwidth}{1in}
    \setlength{\leftmargin}{1.5in}
    \setlength{\labelsep}{.5in}
    \setlength{\rightmargin}{\leftmargin}}

\item[\admm\hfill] Alternating Direction Method of Multipliers
\item[\ccdf\hfill] Complementary Cumulative Distribution Function
\item[\dti\hfill] Diffusion Tensor Imaging
\item[\eeg\hfill] Electroencephalogram
\item[\fc\hfill] Fully Connected
\item[\fm\hfill] Factorization Machine
\item[\fmri\hfill] Functional Magnetic Resonance Imaging
\item[\gru\hfill] Gated Recurrent Unit
\item[\hdrs\hfill] Hamilton Depression Rating Scale
\item[\lr\hfill] Logistic Regression
\item[\lstm\hfill] Long Short-Term Memory
\item[\mvm\hfill] Multi-View Machine
\item[\rbf\hfill] Radial Basis Function
\item[\rnn\hfill] Recurrent Neural Network
\item[\stm\hfill] Support Tensor Machine
\item[\svm\hfill] Support Vector Machine
\item[\ymrs\hfill] Young Mania Rating Scale
\end{list}
 
\summary

A broad spectrum of data from different modalities are generated in the healthcare domain every day, including scalar data (\emph{e.g.}, clinical measures collected at hospitals), tensor data (\emph{e.g.}, neuroimages analyzed by research institutes), graph data (\emph{e.g.}, brain connectivity networks), and sequence data (\emph{e.g.}, digital footprints recorded on smart sensors). Capability for modeling information from these heterogeneous data sources is potentially transformative for investigating disease mechanisms and for informing therapeutic interventions.

Our works in this thesis attempt to facilitate healthcare applications in the setting of broad learning which focuses on fusing heterogeneous data sources for a variety of synergistic knowledge discovery and machine learning tasks. We are generally interested in computer-aided diagnosis, precision medicine, and mobile health by creating accurate user profiles which include important biomarkers, brain connectivity patterns, and latent representations. In particular, our works involve four different data mining problems with application to the healthcare domain: multi-view feature selection, subgraph pattern mining, brain network embedding, and multi-view sequence prediction.

\chapter{Introduction}
\label{chapter:intro}

\section{Thesis Outline}

Healthcare data are in heterogeneous forms. Diagnosis tools and methods have been developed to obtain many measurements from different medical examinations and laboratory tests (\emph{e.g.}, clinical, immunologic, serologic, and cognitive parameters). With rapid advances in neuroimaging techniques, brain networks (\emph{i.e.}, connectomes) can be constructed to map neural connections in the brain which provide us with another perspective of investigating neurological disorders. As a more unobtrusive manner for mental health monitoring, mobile devices present new opportunities to investigate the manifestations of psychiatric diseases in patients’ daily lives. It is critical to model information from these heterogeneous healthcare data.

In this thesis, we attempt to facilitate healthcare applications in the setting of \emph{broad learning} which focuses on fusing heterogeneous data sources for a variety of synergistic knowledge discovery and machine learning tasks. Broad learning provides machine learning problems with a new research dimension that is somewhat orthogonal to pure deep learning approaches. Because the performance of a data-driven model is bounded by either model capacity or data capacity, merely creating a deeper model to increase the model capacity would certainly lead to overfitting given a certain amount of data. Therefore, in order to achieve a further improvement on model performance, we should consider how to effectively make use of a broad spectrum of data sources that are relevant to a learning task. In general, broad learning includes (1) \emph{multi-view learning} that aims to learn from different data sources representing the same entity \cite{cao2014tensor,cao2016multi,lu2017multilinear,lu2018learning}, (2) \emph{transfer learning} that leverages knowledge from a source entity to help with learning a similar target entity \cite{dai2007co,mihalkova2007mapping,mihalkova2009transfer,shi2012transfer,zhang2015deep}, and (3) \emph{learning in heterogeneous information networks} that formulates a learning task around multiple linked entities \cite{cao2014collective,kong2013meta,kong2013multi}.

Our works are mostly related to multi-view learning, where data in different views can exist in different data structures or modalities, including scalars, tensors, graphs, and sequences. In particular, we propose a tensor-based approach to selecting discriminative biomarkers by exploring feature interactions across different data sources. We further utilize auxiliary measures to identify connectivity patterns in the brain that are associated with brain injury and learn effective brain network representations. Moreover, we develop a deep learning framework for mood detection by modeling typing dynamics data that are collected on smartphones.

\section{Multi-View Feature Selection}

(Part of the section was previously published in \cite{cao2014tensor}.)

With the development of disease diagnosis and treatment, many tools and methods have been developed to obtain a large number of measurements from medical examinations and laboratory tests. Different groups of measures characterize the health state of a subject from different aspects. Conventionally, such a type of data is referred to as \emph{multi-view data}. In medical studies, a critical problem is that there are usually a small number of subjects available yet introducing a large number of measurements, some of which may be irrelevant to the diagnosis. These irrelevant features can bring noise into the decision process and potentially result in a wrong judgment. Therefore, feature selection is desirable in order to improve the computer-aided diagnosis as well as interpretability. The selected features can also be used by researchers to find biomarkers for brain diseases which are clinically imperative for detecting brain injury at an early stage before it is irreversible. Valid biomarkers are useful for aiding diagnosis, monitoring disease progression and evaluating effects of intervention \cite{kong2013discriminative}.

In Chapter~\ref{chapter:mvfs}, the task of identifying important biomarkers from multiple groups of medical examinations is formulated as a \emph{multi-view feature selection} problem. We utilize the tensor product operation to model feature interactions across different data sources, factorization techniques to reduce the optimization, and recursive feature elimination to select discriminative biomarkers. The proposed method can work efficiently with many views and effectively with both linear and nonlinear kernels.

\section{Subgraph Pattern Mining}

(Part of the section was previously published in \cite{cao2015mining,cao2015identifying}.)

In cases where there are only a limited number of labeled instances for brain network analysis, information from the graph view alone may be insufficient for finding important subgraph patterns, and we should consider to leverage side information that is available with the graph data. For example, hundreds of clinical, immunologic, serologic, and cognitive measures are usually documented for each subject in medical studies \cite{cao2014tensor,cao2015determinants}. These measures compose multiple side views which contain a tremendous amount of supplemental information in addition to brain networks themselves. It is desirable to find meaningful subgraph patterns in brain networks by utilizing the side views as guidance.

In Chapter~\ref{chapter:subgraph}, the task of finding connectivity patterns in brain networks is formulated as a \emph{subgraph pattern mining} problem. We treat side information as a label proxy and propose to identify subgraph patterns in the brain that are consistent with the side information and associated with brain injury. In contrast to existing subgraph mining approaches that focus on graph instances alone, the proposed method explores multiple vector-based side views to find an optimal set of subgraph features for graph classification. Based on the side views and some available label information, we design an evaluation criterion for subgraph patterns and derive its lower bound. This allows us to develop a branch-and-bound algorithm to efficiently search for optimal subgraph patterns with pruning, thereby avoiding exhaustive enumeration of all subgraph patterns.

\section{Brain Network Embedding}

(Part of the section was previously published in \cite{cao2016semi,cao2017tbne}.)

In order to apply the conventional machine learning algorithms that take vector data as input to brain networks, or graph data in general, one can first compute graph-theoretical measures \cite{wee2012identification,jie2014integration} or extract subgraph patterns \cite{kong2013discriminative,cao2015mining}. However, the expressive power of these explicit features is limited. To explore a larger space of potentially informative features to represent the brain networks, it is desirable to learn the representations in an implicit manner.

In Chapter~\ref{chapter:bne}, the task of converting brain network data from graph structures to vectorial representations is formulated as a \emph{brain network embedding} problem. We leverage tensor factorization techniques to obtain latent representations of brain networks which can be further utilized to facilitate downstream tasks. In particular, undirected brain networks are stacked as a partially symmetric tensor before conducting factorization. The self-report data are incorporated as guidance in the tensor factorization procedure to learn latent factors that are consistent with the side information. Furthermore, the representation learning and classifier training are blended into a unified optimization framework to obtain discriminative representations, by allowing the classifier parameters to interact with the original brain network data via latent factors and the representation learning process to be aware of the supervision information. The formulated optimization problem can be interpreted as partially coupled matrix and tensor factorization with constraints.

\section{Multi-View Sequence Prediction}

(Part of the section was previously published in \cite{cao2017deepmood}.)

The wide use of mobile phones presents new opportunities in the treatment of psychiatric illness by allowing us to study the manifestations of psychiatric illness in an unobtrusive manner and at a level of detail that was not previously possible. Continuous collection of automatically generated smartphone data that reflect illness activity could facilitate early intervention \cite{ankers2009objective,bopp2010longitudinal,faurholt2016behavioral}. The sensor readings are essentially sequence data, and the user data collected from multiple sensors on a mobile device can be considered as multi-view times series data. Modeling the multi-view time series data is challenging because time series in different views may have different timestamps, and thus it is usually non-trivial to align them together. Moreover, they may also have different densities, and thus dense views could dominate after concatenation and potentially override the effects of sparse but important views.

In Chapter~\ref{chapter:deepmood}, the task of inferring mood disturbance from mobile phone typing dynamics metadata is formulated as a \emph{multi-view sequence prediction} problem. We develop a deep learning architecture for mood detection using the collected features about alphanumeric characters, special characters, and accelerometer values. Specifically, it is an end-to-end approach based on late fusion to modeling the multi-view time series data. In the first stage, each view of the time series is separately modeled by a recurrent network. The multi-view information is then fused in the second stage through three alternative layers that concatenate and explore interactions across the output vectors from each view.

\chapter{Multi-view feature selection}
\label{chapter:mvfs}

(This chapter was previously published as ``Tensor-based Multi-view Feature Selection with Applications to Brain Diseases \cite{cao2014tensor}'', in \textit{Proceedings of the 2014 IEEE International Conference on Data Mining (ICDM)}, 2014, IEEE. DOI: \url{https://doi.org/10.1109/ICDM.2014.26}.)

\section{Introduction}

Many neurological disorders are characterized by ongoing injury that is clinically silent for prolonged periods and irreversible by the time symptoms first present. New approaches for detection of early changes in subclinical periods would afford powerful tools for aiding clinical diagnosis, clarifying underlying mechanisms and informing neuroprotective interventions to slow or reverse neural injury for a broad spectrum of brain disorders, including {\hiv} infection on brain \cite{he3dusk,kong2014brain}, Alzheimer's disease \cite{ye2008heterogeneous}, Parkinson's Disease, Schizophrenia, Depression, \emph{etc.} Early diagnosis has the potential to greatly alleviate the burden of brain disorders and the ever increasing costs to families and society. For example, total healthcare costs for those 65 and older, are more that three times higher in those with Alzheimer's and other dementia \cite{mebane20092009}.

\begin{figure}[t]
\centering
\begin{minipage}[l]{0.6\columnwidth}
  \centering
  \includegraphics[width=1\textwidth]{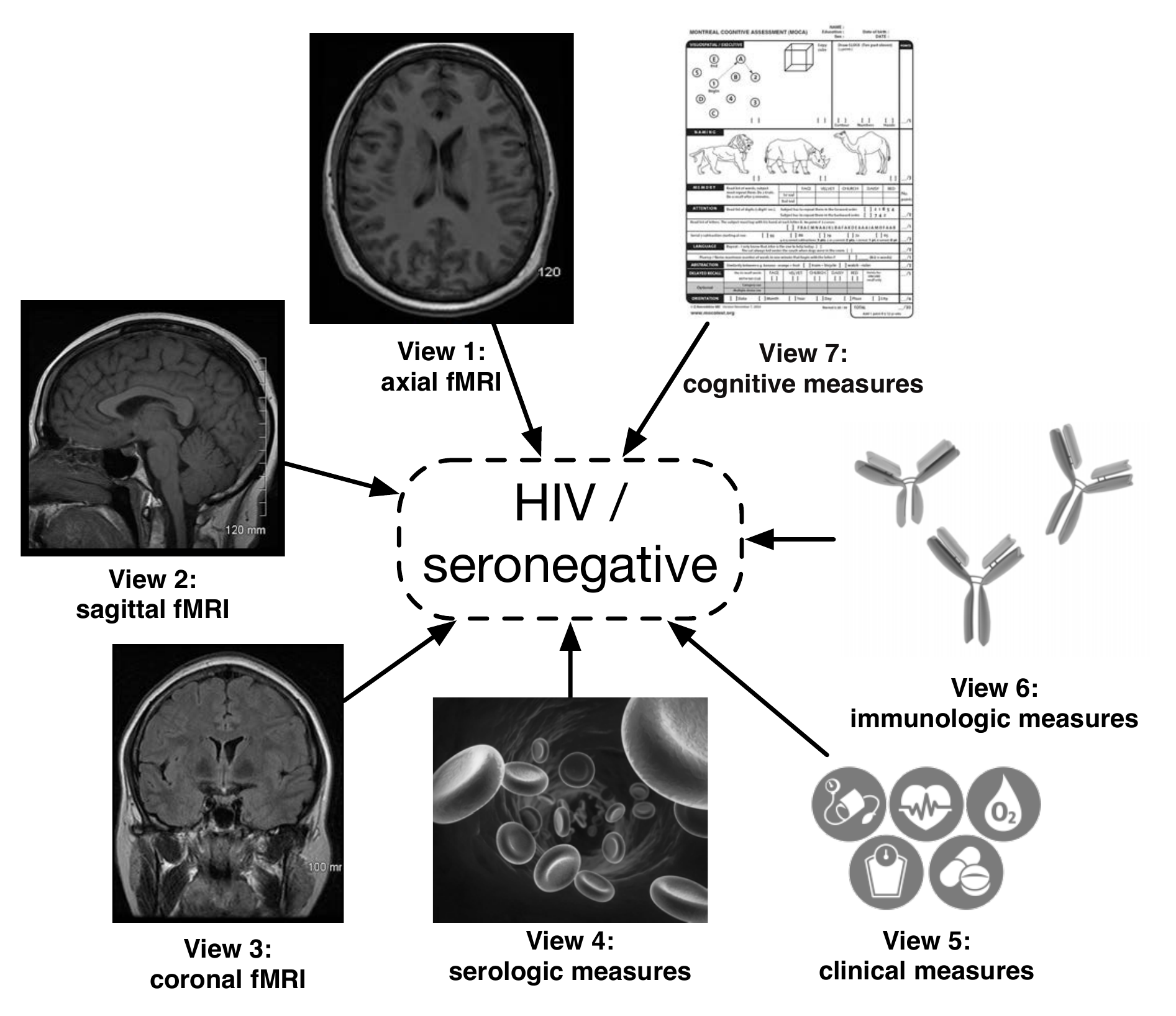}
\end{minipage}
\caption{Example of multi-view data in medical studies.}
\label{fig:multiview}
\end{figure}

As diagnosis of neurological disorder is extremely challenging, tools and methods have been developed to obtain many measurements from different examinations and laboratory tests. As shown in \ref{fig:multiview}, there are measurements from a series of medical examinations documented for each subject, including imaging, clinical, immunologic, serologic, and cognitive measures. Different groups of measures characterize the health state of a subject from different aspects. Conventionally, such a type of data is referred to as \emph{multi-view data}.

In many medical studies, a critical problem is that there are usually a small number of subjects available yet introducing a large number of measurements. However, some of the features in the multi-view data may be irrelevant to the learning task. Moreover, the combination of multiple views can potentially incur redundant and even conflicting information which is unfavorable for classifier learning. Therefore, feature selection should be conducted before or within a machine learning procedure. The selected features can also be used by researchers to find biomarkers for brain diseases which are clinically imperative for detecting brain injury at an early stage before it is irreversible. Valid biomarkers are useful for aiding diagnosis, monitoring disease progression and evaluating effects of intervention \cite{kong2013discriminative}.

A straightforward solution to the multi-view feature selection problem is to handle each view separately and conduct feature selection independently. This paradigm is based on the assumption that each view is sufficient on its own to learn the target concept \cite{xu2013survey}. However, different views can often provide complementary information to each other, thus leading to improved performance. Existing feature selection approaches can generally be categorized as filter models \cite{peng2005feature,robnik2003theoretical} and embedded models based on sparsity regularization \cite{feng2012adaptive,fang2013discriminative,wang2013multi,wang2013heterogeneous}. While in this work, we focus on wrapper models for feature selection. We propose a dual method of tensor-based multi-view feature selection ({\mvfs}), by exploiting the underlying correlations between the input space and the constructed tensor product space. In addition, the proposed method can naturally extend to more than two views and work in conjunction with nonlinear kernels. Empirical studies on an {\hiv} dataset \cite{ragin2012structural} demonstrate that the proposed method can achieve superior classification accuracy. While the experiments are conducted on medical data from a clinical application in {\hiv} infection on brain, the techniques developed for detecting important biomarkers have considerable promise for early diagnosis of other neurological disorders.

\section{Problem Formulation}
\label{sec:mvfs_problem}
In this section, we introduce the problem of multi-view feature selection for classification. 
A multi-view classification problem with $n$ labeled instances represented from $m$ different views can be formulated as: $\mathcal{D} = \left\{ \left (\mathbf{x}_i^{(1)}, \cdots, \mathbf{x}_i^{(m)}, y_i \right)\right\}_{i=1}^n$, where $\mathbf{x}_{i}^{(v)} \in \mathbb{R}^{I_{v} }$, $i \in \{ 1, \cdots, n \}$, $v \in \{ 1, \cdots, m \}$, $I_{v}$ is the dimensionality of the $v$-th view, and $y_{i}\in\{-1,1\}$ is the class label of the $i$-th instance. In this manner, $\mathcal{X}_{i} = \{ \mathbf{x}_{i}^{(1)}, \cdots, \mathbf{x}_{i}^{(m)} \}$ denotes the multi-view features of the $i$-th instance, $\mathcal{X}^{(v)} = \{ \mathbf{x}_{1}^{(v)}, \cdots, \mathbf{x}_{n}^{(v)} \}$ denotes all the instances in the $v$-th view, and $\mathcal{Y} = \left\{ y_{1}, \cdots, y_{n} \right\}$ denotes labels of all the instances.

Considering feature selection, $J_v$ denotes the number of features to be selected in the $v$-th view, and $\mathbf{s}^{(v)}$ denotes the indices of selected features in the $v$-th view.
The task of multi-view feature selection for classification is to determine $\{\mathbf{s}^{(v)}\}_{v=1}^m$ as well as to find a classifier function
$f: \mathbb{R}^{J_1 + \cdots + J_m} \rightarrow \{-1,1\}$ that correctly predicts the label of an unseen instance $\mathcal{X}^* = \{ \mathbf{x}^{(1)*}, \cdots, \mathbf{x}^{(m)*} \}$ where $\mathbf{x}^{(v)*}$ contains features only in $\mathbf{s}^{(v)}$.

\section{Preliminaries} \label{sec:tensor}

Tensors are higher order arrays that generalize the notions of vectors (first-order tensors) and matrices (second-order tensors), whose elements are indexed by more than two indices. Each index expresses a \emph{mode} of variation of the data and corresponds to a coordinate direction. The number of variables in a mode indicates the dimensionality of the mode. The order of a tensor is determined by the number of its modes. The use of this data structure has been advocated in virtue of certain favorable properties. A key to this work is to utilize the tensor structure to capture all the possible feature interactions across different views. 

\begin{Definition}[Tensor product]
The tensor product of two vectors $\mathbf{x} \in \mathbb{R}^{I_{1}}$ and $\mathbf{y} \in \mathbb{R}^{I_{2}}$, denoted by $\mathbf{x} \circ \mathbf{y}$, represents a matrix with the elements $\left(\mathbf{x} \circ \mathbf{y} \right)_{i_1, i_2}\ =\ x_{i_1}y_{i_2}$.
\end{Definition}


The tensor product (\emph{i.e.}, outer product) of vector spaces forms an elegant algebraic structure for the theory of tensors. Such a structure endows tensors with an inherent advantage in representing the real-world data that result from the interaction of multiple factors, where each mode of a tensor corresponds to one factor \cite{geng2011face}. Therefore, the use of tensorial representations is a reasonable choice for adequately capturing the possible relationships among multiple views of data. Another advantage in representing the multi-view information in a tensor data structure is that we can flexibly explore those useful knowledge in the tensor product space by virtue of tensor-based techniques.

Based on the definition of tensor product of two vectors, we can then express $\mathbf{x} \circ \mathbf{y} \circ \mathbf{z}$ as a third-order tensor in $\mathbb{R}^{I_{1} \times I_{2} \times I_{3}}$, of which the elements are defined by $\left(\mathbf{x} \circ \mathbf{y} \circ \mathbf{z} \right)_{i_1, i_2, i_3}\ =\ x_{i_1}y_{i_2}z_{i_3}$. $\mathcal{X}=\left(x_{i_1,\ldots,i_m}\right)$ is used to denote an $m$th-order tensor $\mathcal{X} \in \mathbb{R}^{I_{1} \times \cdots \times I_{m}}$ and its elements, where for $v \in \{ 1, \cdots, m \}$, $I_{v}$ is the dimensionality of $\mathcal{X}$ along the $v$-th mode. $\mathcal{X}_{:, \ldots, :, i_{v}, :, \ldots, : }$ is used to denote the object resulting from fixing the index in $v$-th mode of $\mathcal{X}$ to be $i_v$.


\begin{Definition}[Inner product] The inner product of two same-sized tensors $\mathcal{X}, \mathcal{Y} \in \mathbb{R}^{I_{1} \times \cdots \times I_{m}}$ is defined as the sum of the products of their elements
\begin{equation}
\begin{aligned}
\label{eq:innerproduct1}
\left\langle \mathcal {X}, \mathcal {Y}\right\rangle=\sum_{i_{1}=1}^{I_1} \cdots \sum_{i_{m}=1}^{I_m} x_{i_1,\ldots,i_m} y_{i_1,\ldots,i_m}
\end{aligned}
\end{equation}
\end{Definition}

For tensors $\mathcal{X}=\mathbf{x}^{(1)} \circ \cdots \circ \mathbf{x}^{(m)}$ and $\mathcal{Y}=\mathbf{y}^{(1)} \circ \cdots \circ \mathbf{y}^{(m)}$, it holds that
\begin{equation}
\begin{aligned}
\label{eq:innerproduct2}
\left\langle \mathcal {X}, \mathcal {Y}\right\rangle=\langle \mathbf{x}^{(1)}, \mathbf{y}^{(1)}\rangle \cdots \langle \mathbf{x}^{(m)}, \mathbf{y}^{(m)}\rangle
\end{aligned}
\end{equation}

For the sake of brevity, in the following we use the notations $\prod_{i=1}^{m}\circ \mathbf{x}^{(i)}$ and $ \prod_{i=1}^{m} \langle \mathbf{x}^{(i)}, \mathbf{y}^{(i)} \rangle$ to denote $\mathbf{x}^{(1)} \circ \cdots \circ \mathbf{x}^{(m)}$ and $\langle \mathbf{x}^{(1)}, \mathbf{y}^{(1)} \rangle \cdots \langle \mathbf{x}^{(m)}, \mathbf{y}^{(m)} \rangle$, respectively.

\begin{Definition}[Tensor norm]
The norm of a tensor $\mathcal {X} \in \mathbb{R}^{I_{1} \times \cdots \times I_{m}}$ is defined to be the square root of the sum of the squared value of all elements in the tensor
\begin{equation}
\begin{aligned}
\label{eq:norm}
\left\|\mathcal {X}\right\|_\mathrm{F}=\sqrt{\left\langle \mathcal {X}, \mathcal {X}\right\rangle}=\sqrt{\sum_{i_1=1}^{I_1}\cdots \sum_{i_m=1}^{I_m}x_{i_1,\ldots,i_m}^{2}}
\end{aligned}
\end{equation}
\end{Definition}

As can be seen, the norm of a tensor is a straightforward generalization of the Frobenius norm for matrices and the Euclidean or $l_{2}$ norm for vectors.

\section{Proposed Method}
\label{sec:mvfs_method}
The concept of tensor serves as a backbone for incorporating multi-view features into a consensus representation by means of tensor product, where the complex relationships among views are embedded within the tensor structure. By mining structural information contained in the tensor, knowledge of multi-view features can be extracted and used to establish a predictive model. In this work, we propose {\mvfs} as a dual method of tensor-based multi-view feature selection, inspired by the idea of using Support Vector Machine ({\svm}) for recursive feature elimination ({\vrfe}) \cite{guyon2002gene}. The general idea is to select useful features in conjunction with the classifier by exploiting the feature interactions across multiple views.

\subsection{Tensor Modeling}
Following the introduction to the concepts of tensors in Section~\ref{sec:tensor}, we now describe how multi-view classification can be formulated and implemented in the framework of {\svm}.

By utilizing the tensor product operation, we can bring the multi-view feature vectors of each instance into a tensorial representation. This allows us to transform the multi-view classification task from an independent domain of each view $\{ (\mathcal{X}^{(1)}, \cdots, \mathcal{X}^{(m)}), \mathcal{Y} \}$ to a consensus domain $\{ \mathcal{X}^{(1)} \times \cdots \times \mathcal{X}^{(m)}, \mathcal{Y} \}$ as a tensor classification problem.

For the sake of simplicity, $\mathcal{X}_{i}$ is used to denote $\prod_{v=1}^{m} \circ \mathbf{x}_{i}^{(v)}$, and the dataset of labeled multi-view instances can be represented as $\mathcal{D} = \{ (\mathcal{X}_{1}, y_1), \cdots, (\mathcal{X}_{n}, y_n) \}$. Note that each multi-view instance $\mathcal{X}_{i}$ is an $m$th-order tensor that lies in the tensor product space $\mathbb{R}^{I_{1} \times \cdots \times I_{m}}$, and each element in $\mathcal{X}_{i}$ is the tensor product of multi-view features in the input space, which is denoted by $x_{i (i_1,\ldots,i_m)}$. Based on the definitions of inner product and tensor norm, we can formulate multi-view classification as a global convex optimization problem in the framework of Support Tensor Machine ({\stm}) as
\begin{equation}
\begin{aligned}
\label{eq:svm1}
\operatornamewithlimits{min}_{\mathcal{W},b,\xi} & \frac{1}{2}\left\|\mathcal {W}\right\|^2_\mathrm{F}+C \sum_{i=1}^{n} \xi_{i} \\
\text{s.t.  } & y_{i} (\langle \mathcal{W}, \mathcal{X}_{i} \rangle + b ) \geq 1 - \xi_{i}\\
& \xi_{i} \geq 0, \forall i = 1,\cdots,n.
\end{aligned}
\end{equation}
where $\mathcal{W}$ is the weight tensor which can be regarded as a separating hyperplane in the tensor product space $\mathbb{R}^{I_{1} \times \cdots \times I_{m}}$, $b$ is the bias, $\xi_{i}$ is the error of the $i$-th training instance, and $C$ is the trade-off between the margin and empirical loss. \ref{eq:svm1} can be solved with the use of optimization techniques developed for {\stm} and {\svm}, and the weight tensor $\mathcal{W}$ can be obtained from
\begin{equation}
\begin{aligned}
\label{eq:w1}
\mathcal{W} = \sum_{i=1}^n \alpha_i y_i \mathcal{X}_{i}
\end{aligned}
\end{equation}
where $\alpha_i$ is the dual variable corresponding to each instance. The decision function is
\begin{equation}
\begin{aligned}
\label{eq:f1}
f \left (\mathcal{X}\right)= \texttt{sign} \left( \langle \mathcal{W}, \mathcal{X} \rangle + b \right)
\end{aligned}
\end{equation}
where $\mathcal{X}$ denotes a test instance given by the tensor product of its multi-view features $\mathbf{x}^{(v)}$ for all $v \in \{1, \cdots, m \}$.

However, there are two major drawbacks incurred by the combination of multiple views if directly solve the problem in the {\stm} framework. Firstly, the constructed tensor may contain much redundant and irrelevant higher order features which will degrade the learning performance. Secondly, the dimensionality of the constructed tensor can be extremely large, which grows at an exponential rate with respect to the number of views. Optimizing for the {\stm} problem will suffer from the curse of dimensionality. Therefore, it is necessary to perform feature selection to concentrate the multi-view information and improve the tensorial representation.

\begin{figure}[t]
\centering
\begin{minipage}[l]{0.8\columnwidth}
  \centering
  \includegraphics[width=1\textwidth]{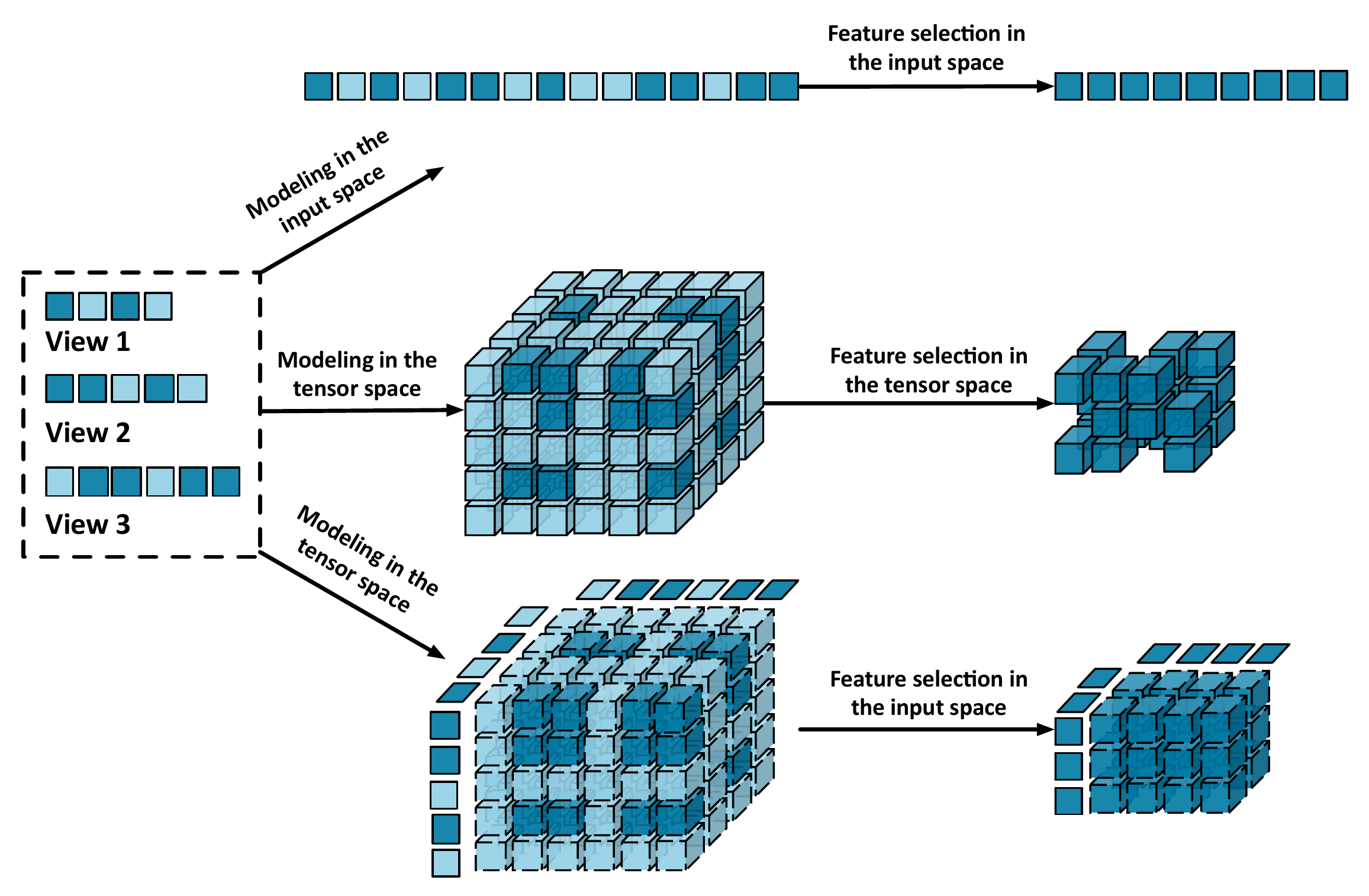}
\end{minipage}
\caption{Three strategies for multi-view feature selection.}
\label{fig:method}
\end{figure}

\subsection{Dual Feature Selection}\label{sec:dual}

{\vrfe} \cite{guyon2002gene} performs {\svm}-based feature selection in a vector space, as the first method shown in \ref{fig:method}. Inspired by {\vrfe}, we can see from \ref{eq:f1} that the inner product of the weight tensor $\mathcal{W} = \left(w_{i_1,\ldots,i_m}\right)$ and the input tensor $\mathcal{X} = \left(x_{i_1,\ldots,i_m}\right)$ determines the value of $f\left( \mathcal{X} \right)$. Intuitively, the tensor product features that are weighted by the largest absolute values influence the most on the classification decision, thus corresponding to the most informative features. Therefore, the absolute weights $\left|w_{i_1,\ldots,i_m}\right|$ or the square of the weights $\left( w_{i_1,\ldots,i_m} \right)^{2}$ can be used as a criterion to select the most discriminative feature subset. Based on this observation, we can conduct recursive feature elimination in the {\stm} framework by
\begin{equation}
\begin{aligned}
\label{eq:arg1}
\operatornamewithlimits{argmin}_{i_1, \cdots, i_m} \left( r_{i_1,\ldots,i_m} \right)
\end{aligned}
\end{equation}
where $r_{i_1,\ldots,i_m}$ denotes the ranking score of each tensor product feature $x_{i_1,\ldots,i_m}$. A straightforward extension of {\vrfe} to the tensor product space is to use the feature ranking criterion
\begin{equation}
\begin{aligned}
\label{eq:r1}
r_{i_1,\ldots,i_m}=\left( w_{i_1,\ldots,i_m} \right)^2
\end{aligned}
\end{equation}

As the second method shown in \ref{fig:method}, it converts multiple views into a tensor and directly performs feature selection in the tensor product space. However, the number of elements in $\mathcal{W}$ is equivalent to the dimensionality of the constructed tensor in the tensor product space. It is usually computationally intractable to enumerate all the elements in $\mathcal{W}$ in such a high-dimensional tensor product space. On the other hand, noise in the original multi-view features may be further exaggerated over the manipulation of tensor product, thereby degrading the generalization performance.

In order to overcome these problems, it is desirable to remove irrelevant features in the input space. In particular, we maintain an independent ranking of features in each view, because each view usually has different statistical properties and intrinsic physical meanings. For each view $v \in \{1, \cdots, m \}$, we can perform recursive feature elimination by
\begin{equation}
\begin{aligned}
\label{eq:arg2}
\operatornamewithlimits{argmin}_{i_v} \left( r_{i_v}^{(v)} \right)
\end{aligned}
\end{equation}
where $r_{i_v}^{(v)}$ denotes the ranking score of the feature $x_{i_v}^{(v)}, i_v \in \{ 1, \cdots, I_v \}$ in the input space.

A general idea is to leverage the weight coefficients $\mathcal{W}$ in the tensor product space to facilitate the implementation of feature selection in the input space. In particular, we evaluate the value of $r_{i_v}^{(v)}$ from $w_{i_1,\ldots,i_m}$ by virtue of the relationship between the input space and the tensor product space. Based on the definition of the tensor product, we can see that the feature $x_{i_v}^{(v)}$ in the input space will diffuse to $\mathcal{X}_{:, \ldots, :, i_{v}, :, \ldots, : }$ in the tensor product space, thus to $\mathcal{W}_{:, \ldots, :, i_{v}, :, \ldots, : }$. Intuitively, it means that the influence of $x_{i_v}^{(v)}$ on the decision function $f \left (\mathcal{X}\right)$ transfers to $\mathcal{X}_{:, \ldots, :, i_{v}, :, \ldots, : }$. Therefore, the ranking score of $x_{i_v}^{(v)}$ can be estimated from the elements in $\mathcal{W}_{:, \ldots, :, i_{v}, :, \ldots, : }$ as
\begin{equation}
\begin{aligned}
\label{eq:r2}
r_{i_v}^{(v)}=\sum_{i_1=1}^{I_1}\cdots\sum_{i_{v-1}=1}^{I_{v-1}}\sum_{i_{v+1}=1}^{I_{v+1}}\cdots\sum_{i_m=1}^{I_m}(w_{i_1,\cdots, i_m})^2
\end{aligned}
\end{equation}


Compared with performing feature selection in the tensor product space, the efficiency is largely improved by removing the irrelevant and redundant features in the input space. In addition, it provides better interpretability by maintaining the physical meanings of the original features without any manipulation. However, it is still prone to overfitting, because the number of elements in $\mathcal{W}$ grows at an exponential rate as the number of views increases. Therefore, the problem reduces to improving the generalization capability of the {\stm} framework. Following the low-rank assumption in the supervised tensor learning framework \cite{tao2007supervised}, here we assume that $\mathcal{W}$ can be decomposed as $\mathcal{W} = \prod_{v=1}^{m} \circ \mathbf{w}^{(v)}$, and then we can rewrite the optimization problem in \ref{eq:svm1} as
\begin{equation}
\begin{aligned}
\label{eq:svm2}
\operatornamewithlimits{min}_{ \mathbf{w}^{(v)} ,b,\xi} & \frac{1}{2} \prod_{v=1}^{m} \left\| \mathbf{w}^{(v)} \right\|^2_\mathrm{F} +C \sum_{i=1}^{n} \xi_{i}\\
\text{s.t.  } & y_{i} \left( \prod_{v=1}^{m} \left \langle \mathbf{w}^{(v)}, \mathbf{x}_{i}^{(v)} \right \rangle + b \right) \geq 1 - \xi_{i}\\
& \xi_{i} \geq 0, \forall i = 1,\cdots,n.
\end{aligned}
\end{equation}
and the optimal decision function is
\begin{equation}
\begin{aligned}
\label{eq:f2}
f \left (\mathcal{X}\right)= \texttt{sign} \left( \prod_{v=1}^{m} \left \langle \mathbf{w}^{(v)}, \mathbf{x}^{(v)} \right \rangle + b \right)
\end{aligned}
\end{equation}

In this manner, the number of variables with respect to $\mathcal{W}$ is greatly reduced from $\prod_{v=1}^{m} I_{v}$ to $\sum_{v=1}^{m} I_{v}$. From \ref{eq:f2}, we can see that the influence of the input feature $x_{i_v}^{(v)}$ on the decision function $f \left (\mathcal{X}\right)$ is determined only by its corresponding weight coefficient $w_{i_v}^{(v)}$. Hence, \ref{eq:r2} can be simplified as
\begin{equation}
\begin{aligned}
\label{eq:r4}
r_{i_v}^{(v)}=\left (w_{i_v}^{(v)} \right)^2
\end{aligned}
\end{equation}

\begin{Theorem}
The ranking criteria, \ref{eq:r2} and \ref{eq:r4} are equivalent per view.
\end{Theorem}
\textsc{Proof}.
Based on the low-rank assumption of the weight tensor, $w_{i_1,\ldots,i_m} = w_{i_1}^{(1)} \cdots w_{i_m}^{(m)}$, we can rewrite \ref{eq:r2} as
\begin{equation}
\begin{aligned}
\label{eq:r5}
r_{i_v}^{(v)} = & \sum_{i_1} \cdots \sum_{i_{v-1}} \sum_{i_{v+1}} \cdots \sum_{i_m} \left( w_{i_1,\cdots, i_m} \right)^2 \\
= & \sum_{i_1}\cdots\sum_{i_{v-1}} \sum_{i_{v+1}} \cdots \sum_{i_m} \left( w_{i_1}^{(1)} \cdots w_{i_m}^{(m)} \right)^2 \\
= & \left (w_{i_v}^{(v)} \right)^2 \prod_{1 \leq j \leq m}^{j \neq v} \left\| \mathbf{w}^{(j)} \right\|^2_\mathrm{F} \\
= & P^{(-v)} \left (w_{i_v}^{(v)} \right)^2
\end{aligned}
\end{equation}
where $P^{(-v)}=\prod_{1 \leq j \leq m}^{j \neq v} \| \mathbf{w}^{(j)} \|^2_\mathrm{F}$. For the $v$-th mode, the multiplier $P^{(-v)}$ is a non-negative constant, thereby having no effects on the ranking order.
$\square$

Now we introduce how to solve the optimization problem in \ref{eq:svm2}. In an iterative manner, we can update the variables associated with one mode while fixing others during each iteration
\begin{equation}
\begin{aligned}
\label{eq:svm3}
\operatornamewithlimits{min}_{ \mathbf{w}^{(v)} ,b^{(v)},\xi^{(v)}} & \frac{P^{(-v)}}{2} \left\| \mathbf{w}^{(v)} \right\|^2_\mathrm{F} +C \sum_{i=1}^{n} \xi_{i}^{(v)}\\
\text{s.t.  } & y_{i} \left( Q_{i}^{(-v)} \left\langle \mathbf{w}^{(v)}, \mathbf{x}_{i}^{(v)} \right\rangle + b^{(v)} \right) \geq 1 - \xi_{i}^{(v)}\\
& \xi_{i}^{(v)} \geq 0, \forall i = 1,\cdots,n.
\end{aligned}
\end{equation}
where $P^{(-v)}$ and $Q_{i}^{(-v)}$ are constants that denote $P^{(-v)}=\prod_{1 \leq j \leq m}^{j \neq v} \| \mathbf{w}^{(j)} \|^2_\mathrm{F}$ and $Q_{i}^{(-v)}=\prod_{1 \leq j \leq m}^{j \neq v} \langle \mathbf{w}^{(j)}, \mathbf{x}_{i}^{(j)}\rangle$.

Let $\mathbf{x}_{i}^{(v)'}= (Q_{i}^{(-v)}/\sqrt{P^{(-v)}})\mathbf{x}_{i}^{(v)}$ and $\mathbf{w}^{(v)'}=\sqrt{P^{(-v)}}\mathbf{w}^{(v)}$, then the optimization problem in \ref{eq:svm3} is equivalent to
\begin{equation}
\begin{aligned}
\label{eq:svm4}
\operatornamewithlimits{min}_{ \mathbf{w}^{(v)'} ,b^{(v)},\xi^{(v)}} & \frac{1}{2} \left\| \mathbf{w}^{(v)'} \right\|^2_\mathrm{F} +C \sum_{i=1}^{n} \xi_{i}^{(v)} \\
\text{s.t.  } & y_{i} \left( \left\langle \mathbf{w}^{(v)'}, \mathbf{x}_{i}^{(v)'} \right\rangle + b^{(v)} \right) \geq 1 - \xi_{i}^{(v)} \\
& \xi_{i}^{(v)} \geq 0, \forall i = 1,\cdots,n.
\end{aligned}
\end{equation}
which reduces to a standard linear {\svm}. It can be efficiently solved by available algorithms, and we can obtain $\mathbf{w}^{(v)}$ as
\begin{equation}
\begin{aligned}
\label{eq:w2}
\mathbf{w}^{(v)} = \frac{1}{P^{(-v)}} \sum_{i=1}^n Q_{i}^{(-v)} \alpha_i^{(v)} y_i \mathbf{x}_{i}^{(v)}
\end{aligned}
\end{equation}
where $\alpha_i^{(v)}$ is the dual variable corresponding to each instance in the $v$-th view.

Algorithm~\ref{algo:mvfs} outlines the proposed {\mvfs} approach which is also illustrated as the third method in \ref{fig:method}. {\mvfs} effectively exploits the relationship between the input space and the tensor product space by optimizing \ref{eq:svm4} for each view in alternation, obtaining feature weights as in \ref{eq:w2}, and eliminating features as in \ref{eq:r4}.
The implementation has been made available at {\github}\footnote{\url{https://github.com/caobokai/tMVFS}}.

\begin{algorithm}[t]
\caption{\mvfs}
\label{algo:mvfs}
\begin{algorithmic}[1]
\REQUIRE
$\{\mathbf{X}^{(v)}\}_{v=1}^m$ (multi-view training samples), $\mathbf{y}$ (class labels), $\{J_v\}_{v=1}^m$ (number of features to be selected in each view)
\ENSURE $\{\mathbf{s}^{(v)}\}_{v=1}^m$ (selected multi-view features)
\FOR{$v = 1~\text{to}~m$}
\STATE Initialize the subset of surviving features: $\mathbf{s}^{(v)}=[1,\cdots,I_{v}]$
\REPEAT
\STATE Restrict training samples to good feature indices: $\mathbf{X}^{(v)*}=\mathbf{X}^{(v)}(\mathbf{s}^{(v)},:)$
\STATE Train the classifier: $\mathbf{\alpha}=\texttt{SVM-train}(\mathbf{X}^{(v)*},\mathbf{y})$ as in \ref{eq:svm4}
\STATE Compute the weight vector $\mathbf{w}^{(v)}$ according to \ref{eq:w2}
\STATE Compute the ranking criteria $\mathbf{r}^{(v)}$ according to \ref{eq:r4}
\STATE Find the bad feature index: $f=\operatornamewithlimits{argmin}(\mathbf{r}^{(v)})$
\STATE Eliminate the feature: $\mathbf{s}^{(v)}(f)=[]$
\UNTIL{length($\mathbf{s}^{(v)}$) $\leq J_v$}
\ENDFOR
\end{algorithmic}
\end{algorithm}

\subsection{Extension to Nonlinear Kernels}\label{sec:nonlinear}

Although applying the tensor product is an effective approach to capturing feature interactions across multiple views, interactions between features within the same view are not considered. To achieve this purpose, we should replace the linear kernel with a nonlinear kernel. Through implicitly projecting features into a high dimensional space within each view, a nonlinear kernel can work in conjunction with tensor product to exploit feature interactions across different views as well as those within each view.

In the case of nonlinear {\svm}s, we first represent the optimization problem in \ref{eq:svm4} in the dual form as
\begin{equation}
\begin{aligned}
\label{eq:svm5}
\operatornamewithlimits{min}_{\alpha} & \frac{1}{2}{\alpha^{(v)}}^\top\mathbf{H}\alpha^{(v)}-{\alpha^{(v)}}^\top\mathbf{1} \\
\text{s.t.  } & \sum\limits_{i=1}^n\alpha_i^{(v)} y_i=0 \\
& 0\leq\alpha_i^{(v)}\leq C, \forall i = 1,\cdots,n. 
\end{aligned}
\end{equation}
where $\mathbf{H}$ is the matrix with elements $y_h y_k \kappa(\mathbf{x}_h^{(v)'},\mathbf{x}_k^{(v)'})$, and $\kappa(\cdot,\cdot)$ is a nonlinear kernel.

To measure the change in the cost function caused by removing an input feature $x_{i_v}^{(v)}$, we can fix $\alpha$ variables and re-compute the matrix $\mathbf{H}$ with the feature removed. This corresponds to computing $\kappa(\mathbf{x}_h^{(v)'}(-i_v),\mathbf{x}_k^{(v)'}(-i_v))$, yielding matrix $\mathbf{H}(-i_v)$, where the notation $\mathbf{x}_h^{(v)'}(-i_v)$ means that the input feature $x_{i_v}^{(v)}$ is removed from $\mathbf{x}_h^{(v)'}$. Therefore, the feature ranking criterion for nonlinear {\svm}s is
\begin{equation}
\begin{aligned}
\label{eq:r6}
r_{i_v}^{(v)}={\alpha^{(v)}}^\top\mathbf{H}\alpha^{(v)}-{\alpha^{(v)}}^\top\mathbf{H}(-i_v)\alpha^{(v)}
\end{aligned}
\end{equation}

The input feature corresponding to the smallest difference $r_{i_v}^{(v)}$ shall be removed. In the linear case, $\kappa(\mathbf{x}_h^{(v)'},\mathbf{x}_k^{(v)'})=\langle \mathbf{x}_{h}^{(v)'} ,\mathbf{x}_{k}^{(v)'} \rangle$ and ${\alpha^{(v)}}^\top\mathbf{H}\alpha^{(v)}=\|\mathbf{w}^{(v)'}\|_F^2$. Therefore, in \ref{eq:r6}, $r_{i_v}^{(v)}=P^{(-v)} \left (w_{i_v}^{(v)} \right)^2\propto\left (w_{i_v}^{(v)} \right)^2$, which is equivalent to the criteria in \ref{eq:r2} and \ref{eq:r4} for the linear kernel.

\section{Experiments}
\label{sec:mvfs_exp}

In this section, we evaluate the compared methods on a classification task with two views, more than two views, and nonlinear kernels.

\subsection{Data Collection}

In order to evaluate the performance on multi-view feature selection for classification, we compare methods on a dataset collected from the Chicago Early HIV Infection Study \cite{ragin2012structural}, which includes 56 {\hiv} and 21 seronegative control subjects.
There are seven groups of features investigated in the data collection, including {\neu}, {\flo}, {\pla}, {\fre}, {\ave}, {\loc}, {\seg}. Each group can be regarded as a distinct view that partially reflects the health state of a subject, and measurements from different medical examinations usually provide complementary information. Different views are sampled to form multiple compositions. The datasets used in the experiments are summarized in \ref{tab:dataset} where ``$\blacksquare$" indicates that the view is selected in the dataset, while ``$\square$" indicates not selected, and the number in braces indicates the number of features in a view. Additionally, features are normalized within $[0,1]$.

\begin{table}[t]
\centering
\scriptsize
\caption{View composition of the {\hiv} dataset.}
\label{tab:dataset}
\begin{tabular}{||l|c|c|c|c|c|c|c|c|c|c||}
\hline
Views		 &{\dtsa} &{\dtsb} &{\dtsc} &{\dtsd} &{\dtse} &{\dtsf} &{\dtsg} &{\dtsh} &{\dtsi} &{\dtsj}\\
\hline\hline
{\neu} (36)&$\square$&$\square$&$\blacksquare$&$\square$&$\square$&$\square$&$\blacksquare$&$\square$&$\blacksquare$&$\blacksquare$\\
{\flo} (65)&$\square$&$\square$&$\square$&$\square$&$\blacksquare$&$\blacksquare$&$\blacksquare$&$\blacksquare$&$\blacksquare$&$\blacksquare$\\
{\pla} (45)&$\blacksquare$&$\square$&$\blacksquare$&$\blacksquare$&$\blacksquare$&$\square$&$\square$&$\blacksquare$&$\blacksquare$&$\blacksquare$\\
{\fre} (28)&$\square$&$\blacksquare$&$\blacksquare$&$\square$&$\square$&$\blacksquare$&$\square$&$\blacksquare$&$\square$&$\blacksquare$\\
{\ave} (21)&$\blacksquare$&$\blacksquare$&$\square$&$\square$&$\blacksquare$&$\blacksquare$&$\blacksquare$&$\square$&$\blacksquare$&$\blacksquare$\\
{\dti} (54)&$\square$&$\square$&$\square$&$\blacksquare$&$\square$&$\blacksquare$&$\blacksquare$&$\blacksquare$&$\blacksquare$&$\blacksquare$\\
{\seg} (12)&$\square$&$\square$&$\square$&$\blacksquare$&$\blacksquare$&$\square$&$\blacksquare$&$\blacksquare$&$\blacksquare$&$\square$\\
\hline
\end{tabular}
\end{table}

\subsection{Compared Methods}

In order to demonstrate the effectiveness of the proposed multi-view feature selection approach, we compare the following methods:
\begin{itemize}[leftmargin=*,noitemsep,topsep=0pt]
\item\textbf{\mvfs}: the proposed dual method of tensor-based multi-view feature selection. It effectively exploits the feature interactions across multiple views in the tensor product feature space and efficiently performs feature selection in the input space.
\item\textbf{\miqp}: iterative tensor product feature selection with mixed-integer quadratic programming \cite{smalter2009feature}. It explicitly considers the cross-domain interactions between two views in the tensor product feature space. The bipartite feature selection problem is formulated as an integer quadratic programming problem. A subset of features is selected that maximizes the sum over the submatrix of the original weight matrix.
\item\textbf{\trfe}: recursive feature elimination with the tensor product features \cite{smalter2009feature}.
\item\textbf{\vrfe}: recursive feature elimination with the concatenated multi-view features \cite{guyon2002gene}.
\item\textbf{\mkl}: multi-kernel {\svm} \cite{fan2008liblinear}. Each kernel corresponds to one of the multiple views, and multiple kernels are combined linearly.
\item\textbf{\tf}: direct modeling with the tensor product features \cite{kolda2009tensor}.
\item\textbf{\cf}: direct modeling with the concatenated multi-view features \cite{libsvm}.
\end{itemize}

\begin{table}[t]
\centering
\footnotesize
\caption{Comparison of multi-view feature selection methods.}
\label{tab:baseline}
\newcolumntype{x}[1]{>{\centering\arraybackslash}p{#1}}
\begin{tabular}{||l|x{1cm}|x{1cm}|x{1cm}|x{1cm}|x{1cm}|x{1cm}|x{1cm}||}
\hline
\multirow{2}*{Properties}&{\cf}&{\tf}&{\mkl}&{\vrfe}&{\trfe}&{\miqp}&\multirow{2}*{\mvfs}     \\
&\cite{libsvm}&\cite{kolda2009tensor}&\cite{fan2008liblinear}&\cite{guyon2002gene}&\cite{smalter2009feature}&\cite{smalter2009feature}&     \\
\hline\hline
Feature selection        &$\times$   &$\times$   &$\times$   &$\surd$    &$\surd$    &$\surd$    &$\surd$    \\
View discrimination      &$\times$   &$\surd$    &$\surd$    &$\times$   &$\surd$    &$\surd$    &$\surd$    \\
Applicability to many views         &$\surd$    &$\surd$    &$\surd$    &$\surd$    &$\times$   &$\times$   &$\surd$    \\
Compatibility with nonlinear kernels&$\surd$    &$\times$   &$\times$    &$\surd$    &$\times$   &$\times$   &$\surd$    \\
\hline
\end{tabular}
\end{table}

In \ref{tab:baseline}, these methods are compared with respect to four properties: whether it performs feature selection, whether it discriminates different views, whether it is applicable to many views, and whether it is compatible with nonlinear kernels. Note that sparsity regularization models \cite{friedman2010note,wang2013heterogeneous} are not considered as we focus on wrapper models in this work.

For a fair comparison, we use {\libsvm} \cite{libsvm} with a linear kernel as the base classifier for all the compared methods. In the experiments, 3-fold cross validation is performed on balanced datasets. The soft margin parameter $C$ is selected through a validation set. For all the feature selection approaches, 50\% of the original features in each view are selected.

\begin{table}[t]
\centering
\caption{Classification performance with two views in the linear case.}
\label{tab:mvfs_mainresult_two}
\begin{tabular}{||c|l|c|c|c|c||}
\hline
\multirow{2}*{Datasets} & \multirow{2}*{Methods} & \multicolumn{4}{c||}{Evaluation metrics} \\
\cline{3-6}
&   &Accuracy	&Precision	&Recall	&{\fone} \\
\hline\hline
\multirow{5}*{{\dtsa}}
&{\mvfs} &0.762 \tiny{(1)} &0.784 \tiny{(1)} &0.792 \tiny{(4)} &0.786 \tiny{(1)} \\
&{\miqp}&0.667 \tiny{(4)} &0.667 \tiny{(5)} &0.833 \tiny{(1)} &0.737 \tiny{(3)} \\
&{\trfe}  &0.643 \tiny{(5)} &0.657 \tiny{(6)} &0.792 \tiny{(4)} &0.713 \tiny{(6)} \\
&{\vrfe}  &0.690 \tiny{(3)} &0.730 \tiny{(3)} &0.750 \tiny{(7)} &0.736 \tiny{(4)} \\
\cline{2-6}
&{\mkl} &0.643 \tiny{(5)} &0.710 \tiny{(4)} &0.792 \tiny{(4)} &0.730 \tiny{(5)} \\
&{\tf} &0.595 \tiny{(7)} &0.623 \tiny{(7)} &0.833 \tiny{(1)} &0.698 \tiny{(7)} \\
&{\cf}  &0.738 \tiny{(2)} &0.747 \tiny{(2)} &0.833 \tiny{(1)} &0.782 \tiny{(2)} \\
\hline
\multirow{5}*{{\dtsb}}
&{\mvfs} &0.714 \tiny{(1)} &0.769 \tiny{(1)} &0.792 \tiny{(1)} &0.759 \tiny{(1)}	\\
&{\miqp}&0.595 \tiny{(5)} &0.638 \tiny{(6)} &0.667 \tiny{(4)} &0.648 \tiny{(5)}	\\
&{\trfe}  &0.619 \tiny{(3)} &0.656 \tiny{(4)} &0.750 \tiny{(2)} &0.690 \tiny{(3)}	\\
&{\vrfe}  &0.524 \tiny{(7)} &0.607 \tiny{(7)} &0.500 \tiny{(6)} &0.540 \tiny{(7)}	\\
\cline{2-6}
&{\mkl} &0.667 \tiny{(2)} &0.730 \tiny{(2)} &0.667 \tiny{(4)} &0.692 \tiny{(2)}	\\
&{\tf} &0.619 \tiny{(3)} &0.639 \tiny{(5)} &0.750 \tiny{(2)} &0.683 \tiny{(4)}	\\
&{\cf}  &0.548 \tiny{(6)} &0.657 \tiny{(3)} &0.500 \tiny{(6)} &0.553 \tiny{(6)}	\\
\hline
\end{tabular}
\end{table}

\subsection{Performance on Two Views}
\label{sec:two}

We first study the effectiveness of our proposed method on the task of learning from two views. The average performance of the compared methods with standard deviations is reported with respect to four evaluation metrics: accuracy, precision, recall and {\fone} score. Results on {\dtsa} and {\dtsb} are shown in \ref{tab:mvfs_mainresult_two} where for each dataset the four methods at the top perform feature selection.

Considering feature selection, {\mvfs} significantly improves the accuracy over other methods by effectively pruning redundant and irrelevant features. On the other hand, {\miqp} conducts feature selection by maximizing the sum over the weight submatrix in the tensor product feature space, and simply applying recursive feature elimination on either the input space (\emph{i.e.}, {\vrfe}) or the tensor product feature space (\emph{i.e.}, {\trfe}) cannot achieve a better performance. For {\vrfe}, feature interactions between different views are not exploited when selecting features; while for {\trfe}, features are directly selected in the tensor product space, resulting in the potential of overfitting.

In comparison of the three methods at the bottom that perform no feature selection, none of them demonstrates a clear advantage. The performance varies a lot that depends on the given access to different views, and the redundancy in each view is not resolved. Therefore, it is necessary to select discriminative features and eliminate redundant ones when combining multiple views.

\begin{table}[t]
\centering
\tiny
\caption{Classification performance with many views in the linear case.}
\label{tab:mvfs_mainresult_many}
\begin{tabular}{||c|l|c|c|c|c||}
\hline
\multirow{2}*{Datasets} & \multirow{2}*{Methods} & \multicolumn{4}{c||}{Evaluation metrics} \\
\cline{3-6}
&   &Accuracy	&Precision	&Recall	&{\fone} \\
\hline\hline
\multirow{5}*{{\dtsc}}
&{\mvfs} &0.833 \tiny{(1)} &0.926 \tiny{(1)} &0.792 \tiny{(2)} &0.846 \tiny{(1)}	\\
&{\vrfe}  &0.714 \tiny{(4)} &0.741 \tiny{(4)} &0.792 \tiny{(2)} &0.761 \tiny{(4)}	\\
\cline{2-6}
&{\mkl} &0.738 \tiny{(3)} &0.783 \tiny{(3)} &0.750 \tiny{(5)} &0.763 \tiny{(3)}	\\
&{\tf} &0.690 \tiny{(5)} &0.692 \tiny{(5)} &0.833 \tiny{(1)} &0.753 \tiny{(5)}	\\
&{\cf}  &0.762 \tiny{(2)} &0.795 \tiny{(2)} &0.792 \tiny{(2)} &0.791 \tiny{(2)}	\\
\hline
\multirow{5}*{{\dtsd}}
&{\mvfs} &0.810 \tiny{(1)} &0.820 \tiny{(2)} &0.875 \tiny{(1)} &0.839 \tiny{(1)}	\\
&{\vrfe}  &0.667 \tiny{(5)} &0.692 \tiny{(5)} &0.750 \tiny{(3)} &0.718 \tiny{(4)}	\\
\cline{2-6}
&{\mkl} &0.714 \tiny{(2)} &0.822 \tiny{(1)} &0.667 \tiny{(5)} &0.709 \tiny{(5)}	\\
&{\tf} &0.714 \tiny{(2)} &0.711 \tiny{(4)} &0.833 \tiny{(2)} &0.767 \tiny{(2)}	\\
&{\cf}  &0.690 \tiny{(4)} &0.727 \tiny{(3)} &0.750 \tiny{(3)} &0.734 \tiny{(3)}	\\
\hline
\multirow{5}*{{\dtse}}
&{\mvfs} &0.929 \tiny{(1)} &0.926 \tiny{(1)} &0.958 \tiny{(1)} &0.939 \tiny{(1)}	\\
&{\vrfe}  &0.881 \tiny{(3)} &0.852 \tiny{(3)} &0.958 \tiny{(1)} &0.902 \tiny{(3)}	\\
\cline{2-6}
&{\mkl} &0.905 \tiny{(2)} &0.917 \tiny{(2)} &0.917 \tiny{(3)} &0.917 \tiny{(2)}	\\
&{\tf} &0.833 \tiny{(5)} &0.838 \tiny{(5)} &0.875 \tiny{(5)} &0.855 \tiny{(5)}	\\
&{\cf}  &0.857 \tiny{(4)} &0.847 \tiny{(4)} &0.917 \tiny{(3)} &0.880 \tiny{(4)}	\\
\hline
\multirow{5}*{{\dtsf}}
&{\mvfs} &0.929 \tiny{(1)} &0.958 \tiny{(1)} &0.917 \tiny{(1)} &0.936 \tiny{(1)}	\\
&{\vrfe}  &0.833 \tiny{(4)} &0.878 \tiny{(4)} &0.833 \tiny{(5)} &0.852 \tiny{(4)}	\\
\cline{2-6}
&{\mkl} &0.905 \tiny{(2)} &0.917 \tiny{(2)} &0.917 \tiny{(1)} &0.917 \tiny{(2)}	\\
&{\tf} &0.810 \tiny{(5)} &0.792 \tiny{(5)} &0.917 \tiny{(1)} &0.847 \tiny{(5)}	\\
&{\cf}  &0.857 \tiny{(3)} &0.886 \tiny{(3)} &0.875 \tiny{(4)} &0.874 \tiny{(3)}	\\
\hline
\multirow{5}*{{\dtsg}}
&{\mvfs} &0.952 \tiny{(1)} &0.963 \tiny{(1)} &0.958 \tiny{(1)} &0.958 \tiny{(1)}	\\
&{\vrfe}  &0.905 \tiny{(2)} &0.963 \tiny{(1)} &0.875 \tiny{(3)} &0.911 \tiny{(3)}	\\
\cline{2-6}
&{\mkl} &0.905 \tiny{(2)} &0.917 \tiny{(4)} &0.917 \tiny{(2)} &0.917 \tiny{(2)}	\\
&{\tf} &0.810 \tiny{(5)} &0.812 \tiny{(5)} &0.875 \tiny{(3)} &0.837 \tiny{(5)}	\\
&{\cf}  &0.905 \tiny{(2)} &0.963 \tiny{(1)} &0.875 \tiny{(3)} &0.911 \tiny{(3)}	\\
\hline
\multirow{5}*{{\dtsh}}
&{\mvfs} &0.905 \tiny{(1)} &0.917 \tiny{(1)} &0.917 \tiny{(1)} &0.917 \tiny{(1)}	\\
&{\vrfe}  &0.857 \tiny{(4)} &0.847 \tiny{(4)} &0.917 \tiny{(1)} &0.880 \tiny{(4)}	\\
\cline{2-6}
&{\mkl} &0.905 \tiny{(1)} &0.917 \tiny{(1)} &0.917 \tiny{(1)} &0.917 \tiny{(1)}	\\
&{\tf} &0.714 \tiny{(5)} &0.719 \tiny{(5)} &0.833 \tiny{(5)} &0.771 \tiny{(5)}	\\
&{\cf}  &0.881 \tiny{(3)} &0.915 \tiny{(3)} &0.875 \tiny{(4)} &0.892 \tiny{(3)}	\\
\hline
\multirow{5}*{{\dtsi}}
&{\mvfs} &0.952 \tiny{(1)} &1.000 \tiny{(1)} &0.917 \tiny{(1)} &0.956 \tiny{(1)}	\\
&{\vrfe}  &0.905 \tiny{(2)} &0.952 \tiny{(2)} &0.875 \tiny{(3)} &0.911 \tiny{(3)}	\\
\cline{2-6}
&{\mkl} &0.905 \tiny{(2)} &0.917 \tiny{(3)} &0.917 \tiny{(1)} &0.917 \tiny{(2)}	\\
&{\tf} &0.833 \tiny{(5)} &0.838 \tiny{(5)} &0.875 \tiny{(3)} &0.855 \tiny{(5)}	\\
&{\cf}  &0.881 \tiny{(4)} &0.915 \tiny{(4)} &0.875 \tiny{(3)} &0.892 \tiny{(4)}	\\
\hline
\multirow{5}*{{\dtsj}}
&{\mvfs} &0.929 \tiny{(1)} &1.000 \tiny{(1)} &0.875 \tiny{(3)} &0.930 \tiny{(1)}	\\
&{\vrfe}  &0.905 \tiny{(2)} &0.952 \tiny{(2)} &0.875 \tiny{(3)} &0.911 \tiny{(4)}	\\
\cline{2-6}
&{\mkl} &0.905 \tiny{(2)} &0.917 \tiny{(4)} &0.917 \tiny{(1)} &0.917 \tiny{(2)}	\\
&{\tf} &0.810 \tiny{(5)} &0.810 \tiny{(5)} &0.875 \tiny{(3)} &0.841 \tiny{(5)}	\\
&{\cf}  &0.905 \tiny{(2)} &0.921 \tiny{(3)} &0.917 \tiny{(1)} &0.917 \tiny{(2)}	\\
\hline
\end{tabular}
\end{table}

\begin{figure}[t]
\centering
\begin{minipage}[l]{0.6\columnwidth}
  \centering
  \includegraphics[width=1\textwidth]{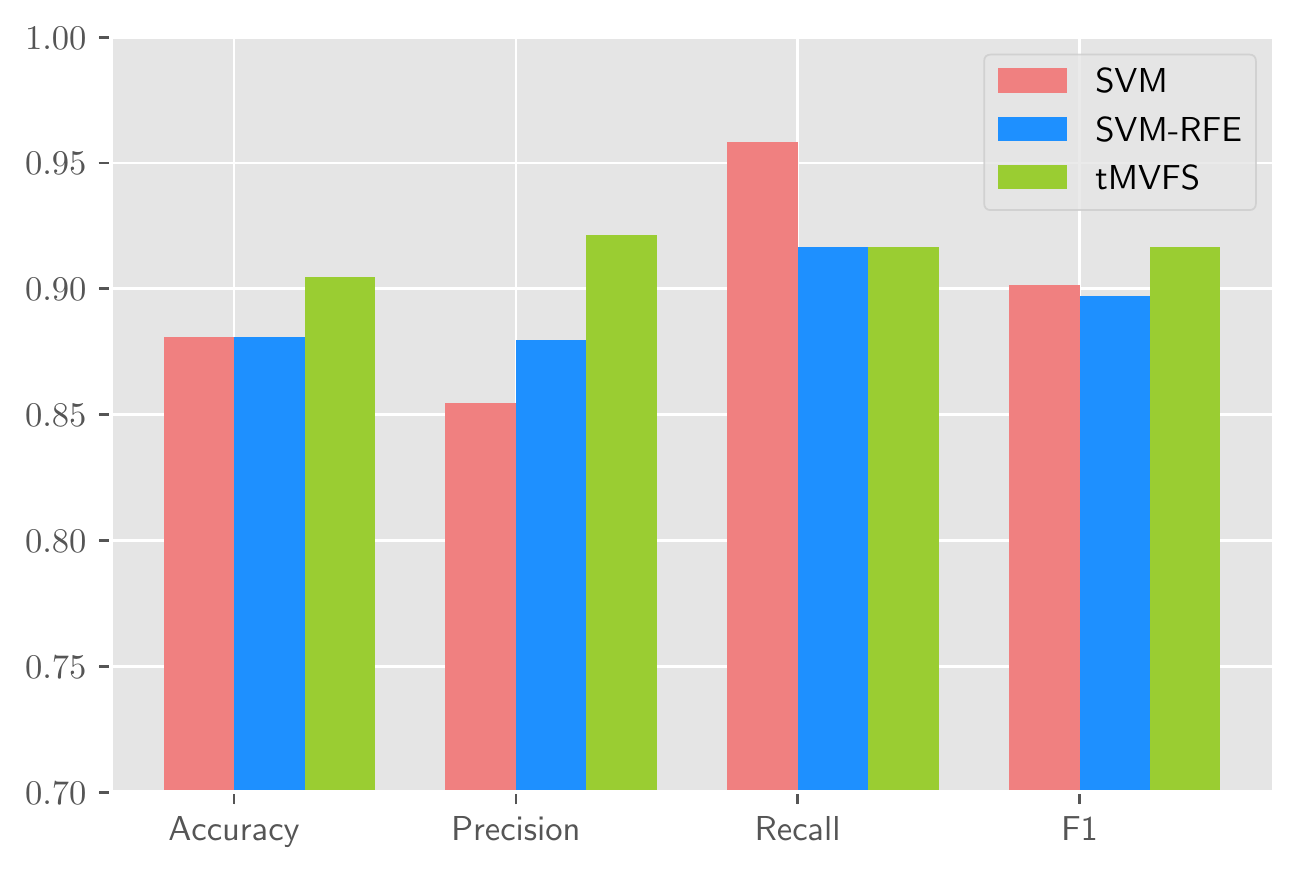}
\end{minipage}
\caption{Classification performance in the nonlinear case.}
\label{fig:rbf}
\end{figure}

\subsection{Performance on Many Views}
\label{sec:many}

In many real-world applications, there are usually more than two views. It is desirable to leverage all of them simultaneously. However, {\miqp} and {\trfe} need to explicitly compute the tensor product feature space, resulting in space complexity that is exponential to the number of views. They are therefore no longer feasible in the case of many views, due to high dimensionality of the tensor product feature space. However, {\mvfs} implicitly exploits the feature interactions across multiple views in the tensor product feature space and efficiently performs feature selection in the input space. Thus, the time complexity and space complexity of {\mvfs} are linear with respect to the number of views, and thus {\mvfs} can naturally extend to more than two views. The experimental results are presented in \ref{tab:mvfs_mainresult_many}.

We can observe that neither {\cf} nor {\vrfe} performs well by simply concatenating the multi-view features. {\tf} performs the worst in most cases as it computes the tensor product feature space where some potentially irrelevant features are included. In general, {\mkl} performs well by modeling each view with a kernel and combining them linearly. By conducting feature selection, {\mvfs} is able to achieve a significant improvement over other methods, especially in terms of accuracy and {\fone} score. It indicates that compared with approaches not distinguishing different views or not conducting feature selection, a better subset of discriminative features can be selected for classification by exploring the feature interactions across multiple views using tensor techniques.

\subsection{Performance on Nonlinear Kernels}
\label{sec:exp_nonlinear}

Here we replace the linear kernel with the radial basis function ({\rbf}) kernel for {\mvfs}, {\vrfe}, and {\cf}. Note that {\miqp}, {\trfe}, and {\tf} are not applicable, because they need to explicitly compute the high dimensional feature space which is intractable when we apply a nonlinear kernel. The experimental results are shown in \ref{fig:rbf}. It illustrates that {\mvfs} still outperforms other methods in the nonlinear case, in terms of accuracy and {\fone} score.

\begin{table}[t]
\centering
\caption{{\hiv}-related features selected in each view.}
\label{tab:mvfs_features}
\begin{tabular}{||l|l||}
\hline
Views & Selected features\\
\hline\hline
{\neu} & \emph{Karnofsky Performance Scale, NART FSIQ, Rey Trial}\\
{\flo} & \emph{Tcells 4+8-, 3+56-16+NKT Cells 4+8-, Lymphocytes}\\
{\pla} & \emph{MMP-2, GRO, TGFa}\\
{\fre} & \emph{Cerebral Cortex, Thalamus Proper, CC\_Mid\_Posterior}\\
{\ave} & \emph{MTR-CC, MTR-Hippocampus, MD-Cerebral-White-Matter}\\
{\loc} & \emph{MTR-CC\_Mid\_Anterior, FA-CC\_Anterior, MTR-CC\_Central}\\
{\seg} & \emph{Norm Peripheral Gray Volume, BPV, Norm Brain Volume}\\
\hline
\end{tabular}
\end{table}

\subsection{Feature Evaluation}
\label{sec:exp_feature}

\ref{tab:mvfs_features} lists the most discriminative features selected by {\mvfs} which can be validated by medical literature on {\hiv} and brain injury. The Karnofsky Performance Status is the most widely used health status measure in {\hiv} medicine and research \cite{o1995validity}. \texttt{CD4+ T} cell depletion is observed during all stages of {\hiv} disease \cite{brenchley2004cd4+}. Mycoplasma membrane protein (\texttt{MMP}) is identified as a possible cofactor responsible for the progression of {\aids} \cite{chowdhury1990mycoplasma}. The fronto-orbital cortex, one of the cerebral cortical areas, is mainly damaged in {\aids} brains \cite{weis1993neuronal}. Magnetization transfer ratio (\texttt{MTR}) is reduced in patients with {\hiv} and correlated with dementia severity \cite{ragin2004disease}. {\hiv} dementia is associated with specific gray matter volume reduction, as well as with generalized volume reduction of white matter \cite{aylward1995magnetic}.

\section{Related Work}

Representative methods for multi-view learning can be categorized into three groups \cite{xu2013survey}: co-training, multi-kernel learning, and subspace learning. Generally, the co-training algorithms are a classic approach for semi-supervised learning, which train on different views in alternation to maximize the mutual agreement. Multi-kernel learning algorithms use a separate kernel for each view and combine them either linearly \cite{lanckriet2004learning} or nonlinearly \cite{varma2009more,cortes2009learning} to improve learning performance. Subspace learning algorithms learn a latent subspace, from which multiple views are generated. Multi-kernel learning and subspace learning are generalized as co-regularization algorithms \cite{sun2013survey}, where the disagreement between the functions of different views is included in the objective function to be minimized. Overall, by exploring the consistency and complementary properties of different views, multi-view learning is more effective than single-view learning.

One of the key challenges for multi-view classification comes from the fact that the incorporation of multiple views will bring much redundant and even conflicting information which is unfavorable for supervised learning. In order to tackle this problem, feature selection is a promising solution. Most of the existing studies can be categorized as filter models \cite{peng2005feature,robnik2003theoretical} and embedded models based on sparsity regularization \cite{feng2012adaptive,fang2013discriminative,wang2013multi,wang2013heterogeneous}. While in this work, we focus on wrapper models for feature selection. 

Guyon et al.~developed a wrapper model by utilizing the weight vector produced from {\svm} and performing recursive feature elimination ({\vrfe}) \cite{guyon2002gene}. Smalter et al.~formulated the problem of feature selection in the tensor product space as an integer quadratic programming problem \cite{smalter2009feature}. However, this method is limited to the interaction between two views, and it is hard to have it extended to many views, since it directly selects features in the tensor product space which leads to the curse of dimensionality. Tang et al.~studied multi-view feature selection in the unsupervised setting by constraining that similar instances from each view should have similar pseudo-class labels \cite{tang2013unsupervised}.

In this work, we use tensor product to organize multi-view features and solve the problem of multi-view feature selection based on recursive feature elimination. Different from the existing methods, we develop a wrapper feature selection model by leveraging the relationship between the original data and the constructed tensor.

\chapter{Subgraph pattern mining}
\label{chapter:subgraph}

(This chapter was previously published as ``Mining Brain Networks using Multiple Side Views for Neurological Disorder Identification \cite{cao2015mining}'', in \textit{Proceedings of the 2015 IEEE International Conference on Data Mining (ICDM)}, 2015, IEEE. DOI: \url{https://doi.org/10.1109/ICDM.2015.50}. An extended version was previously published as ``Identifying HIV-induced Subgraph Patterns in Brain Networks with Side Information \cite{cao2015identifying}'', in \textit{Brain Informatics}, 2015, Springer. DOI: \url{https://doi.org/10.1007/s40708-015-0023-1}. This work is licensed under a Creative Commons Attribution 4.0 International License.)

\section{Introduction}

Recent years have witnessed an increasing amount of data in the form of graph representations which involve complex structures, \emph{e.g.}, brain networks and social networks. Such data are inherently represented as a set of nodes and links, instead of feature vectors as traditional data. For example, brain networks are composed of brain regions as nodes (\emph{e.g.}, \emph{insula} and \emph{hippocampus}) and functional or structural connectivity between the brain regions as links. The linkage structures in brain networks encode tremendous information about the mental health of a subject \cite{ajilore2013constructing,craddock2012whole}. For example, in brain networks derived from functional magnetic resonance imaging ({\fmri}), links represent the correlations between functional activities of brain regions. Functional brain networks provide a graph-theoretical viewpoint to investigate the collective patterns of functional activities across all brain regions \cite{jie2014integration}, with potential applications to the early detection of brain diseases \cite{wang2011abnormalities}. On the other hand, links in diffusion tensor imaging ({\dti}) brain networks indicate the number of neural fibers that connect different brain regions. 

A typical approach to modeling graph instances is to first extract a set of subgraph patterns for downstream tasks. Conventional approaches focus on mining subgraph patterns from the graph view alone which is usually feasible for applications such as molecular graph analysis, where a large set of labeled graph instances are available. For brain network analysis, however, we usually have only a small number of graph instances \cite{kong2013discriminative}, and thus the information from the graph view alone may be insufficient for finding important subgraphs. We notice that the side information is usually available with the graph data for brain disorder identification. For example, hundreds of clinical, immunologic, serologic, and cognitive measures are usually documented for each subject in medical studies \cite{cao2014tensor,cao2015determinants}, in addition to the brain network data. These measures compose multiple side views that contain a tremendous amount of supplemental information for diagnostic purposes. It is desirable to extract valuable information from these side views to guide the procedure of subgraph mining in brain networks.

\begin{figure}[t]
\centering
\subfigure[Late fusion.]{
\label{fig:method1}
\begin{minipage}[l]{0.45\columnwidth}
  \centering
  \includegraphics[width=1\textwidth]{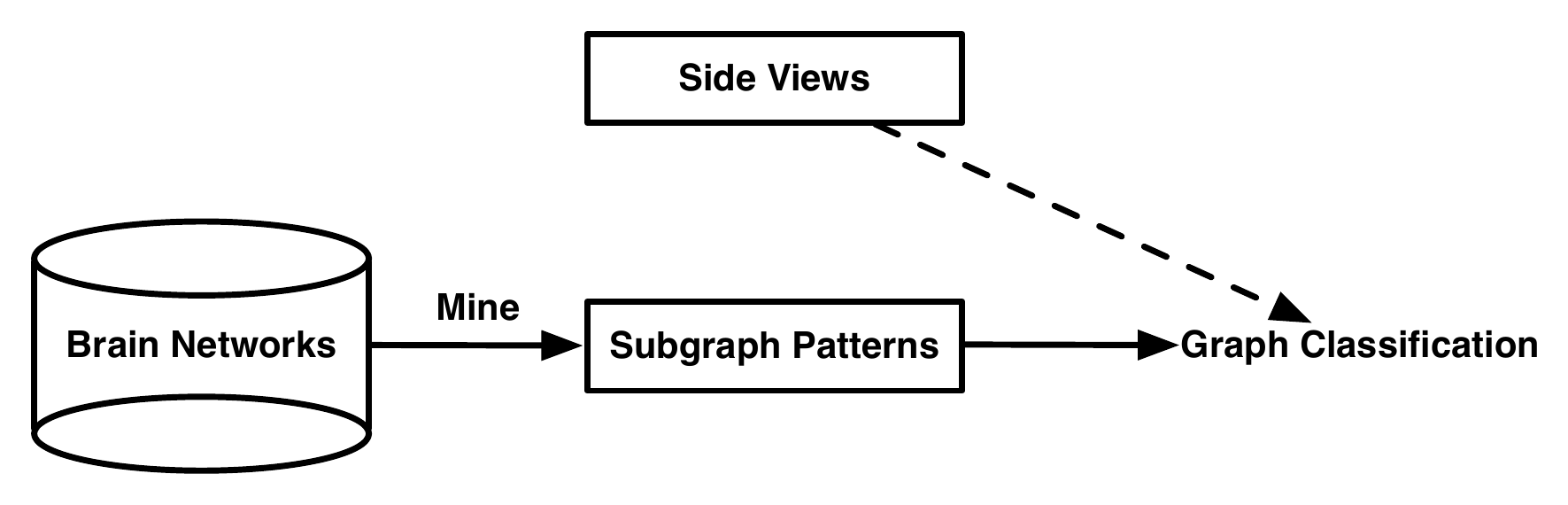}
\end{minipage}
}
\subfigure[Early fusion.]{
\label{fig:method2}
\begin{minipage}[l]{0.45\columnwidth}
  \centering
  \includegraphics[width=1\textwidth]{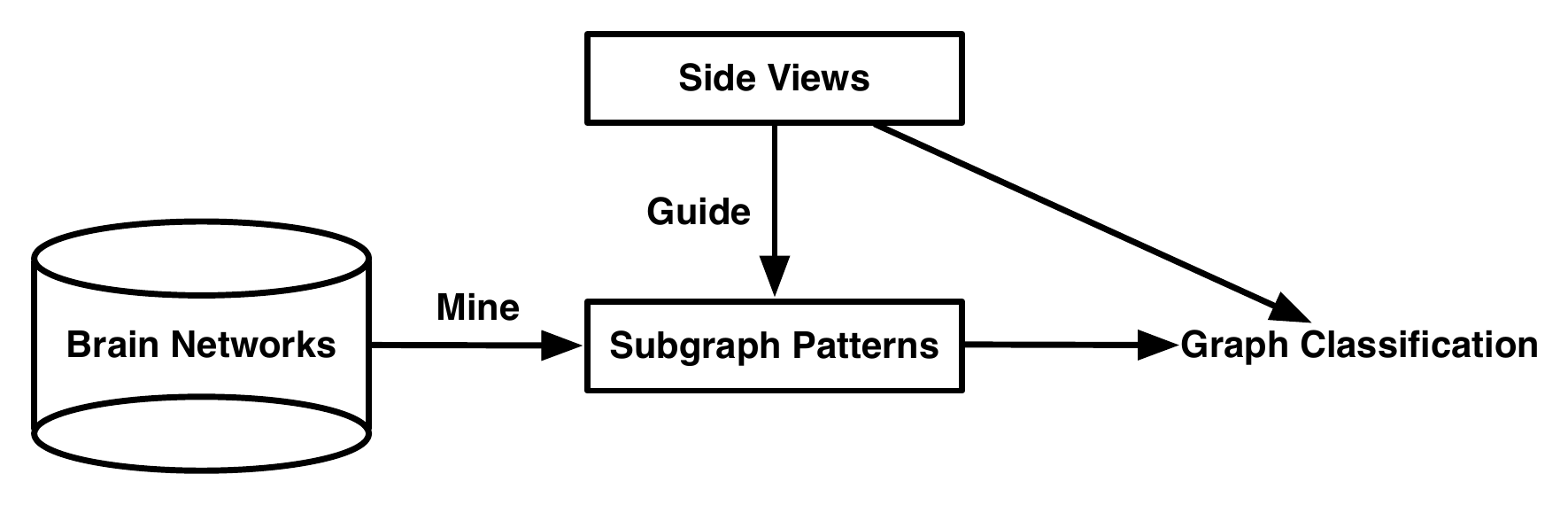}
\end{minipage}
}
\caption{Two strategies for using side information in subgraph selection.}
\end{figure}

Figure~\ref{fig:method1} illustrates that conventionally side views and subgraph patterns are modeled separately and they may only be concatenated in the final step for graph classification. In this manner, the valuable information embedded in side views is not fully leveraged in the subgraph selection procedure. For brain network analysis with a small sample size, it is critical to learn knowledge from other data sources. We notice that transfer learning can borrow supervision knowledge from the source domain to help with learning in the target domain, \emph{e.g.}, finding a good feature representation \cite{dai2007co}, mapping relational knowledge \cite{mihalkova2007mapping,mihalkova2009transfer}, and learning across graph databases \cite{shi2012transfer}. However, existing approaches cannot transfer complementary information from vector-based side views to graph instances with complex structures.

As an attempt to tackle this problem, we introduce a novel framework for discriminative subgraph selection using multiple side views, as illustrated in Figure~\ref{fig:method2}. In contrast to existing subgraph mining approaches that focus on the graph view alone, the proposed method can explore multiple vector-based side views to find an optimal set of subgraph features for graph classification. Based on the side views and some available label information, we design an evaluation criterion for subgraph features, named {\gscore}. By further deriving its lower bound, we develop a branch-and-bound algorithm, named {\gmsv}, to avoid exhaustive enumeration of all subgraph features and efficiently search for optimal subgraph features with pruning. We use real-world {\fmri} and {\dti} brain network datasets to evaluate the proposed method. The experimental results demonstrate that the subgraph selection method using multiple side views can effectively boost the graph classification performance. Moreover, we show that {\gmsv} is more efficient by pruning the subgraph search space via {\gscore}.

\section{Problem Formulation}
\label{sec:subgraph_problem}

Before presenting the subgraph feature selection model, we first introduce the notations that will be used throughout this work. A graph dataset is denoted as $\mathcal{D}=\{G_1,\cdots,G_n\}$ where there are $n$ graph instances, and each of them $G_i$ is labeled by $y_i\in\{-1,1\}$.

\begin{Definition}[Side view]
A side view is a set of vector-based features $\mathbf{z}_i=[z_1,\cdots,z_d]^\top$ associated with each graph instance $G_i$, where $d$ is the dimensionality of this view. A side view is denoted as $\mathbf{Z}=[\mathbf{z}_1,\cdots,\mathbf{z}_n]$.
\end{Definition}

It is assumed that there are multiple side views $\mathcal{Z}=\{\mathbf{Z}^{(1)},\cdots,\mathbf{Z}^{(v)}\}$ along with the graph dataset $\mathcal{D}$, where $v$ is the number of side views. We employ kernels $\kappa^{(p)}$ on $\mathbf{Z}^{(p)}$, such that $\kappa^{(p)}(i,j)$ represents the similarity between $G_i$ and $G_j$ from the perspective of the $p$-th side view.
The {\rbf} kernel is used as the default kernel in this work, unless otherwise specified
\begin{equation}
\begin{aligned}
\label{eq:rbf}
\kappa^{(p)}(i,j)=\text{exp}\left(-\frac{\|\mathbf{z}_i^{(p)}-\mathbf{z}_j^{(p)}\|_2^2}{d^{(p)}}\right)
\end{aligned}
\end{equation}

\begin{Definition}[Graph] A graph is represented as $G =(V,E)$, where $V$ is the set of nodes, and $E\subseteq V\times V$ is the set of links.
\end{Definition}

\begin{Definition}[Subgraph]
Let $G'=(V',E')$ and $G=(V,E)$ be two graphs. $G'$ is a subgraph of $G$ (denoted as $G'\subseteq G$) if $V'\subseteq V$ and $E'\subseteq E$. If $G'$ is a subgraph of $G$, then $G$ is supergraph of $G'$.
\end{Definition}

In this work, we adopt the idea from subgraph-based graph classification approaches, which assume that each graph instance $G_j$ is represented as a binary vector $\mathbf{x}_j=[x_{1j},\cdots,x_{mj}]^\top$ associated with the full set of subgraph patterns $\mathcal{S}=\{g_1,\cdots,g_m\}$ for the graph dataset $\mathcal{D}$. Here $x_{ij}\in\{0,1\}$ is the binary feature of $G_j$ corresponding to the subgraph pattern $g_i$, and $x_{ij}=1$ if $g_i$ is a subgraph of $G_j$ ($g_i\subseteq G_j$), otherwise $x_{ij}=0$. Let $\mathbf{X}=[x_{ij}]^{m\times n}$ denote the matrix consisting of binary feature vectors using $\mathcal{S}$ to represent the graph dataset $\mathcal{D}$, then $\mathbf{X}=[\mathbf{x}_1,\cdots,\mathbf{x}_n]=[\mathbf{f}_1,\cdots,\mathbf{f}_m]^\top\in\{0,1\}^{m\times n}$, where $\mathbf{f}_i=[f_{i1},\cdots,f_{in}]^\top\in\{0,1\}^n$ is the indicator vector for subgraph pattern $g_i$, and $f_{ij}=1$ if $g_i\subseteq G_j$, otherwise $f_{ij}=0$.

The full set $\mathcal{S}$ is usually too large to be enumerated, and only a subset of subgraph patterns $\mathcal{T}\subseteq\mathcal{S}$ is relevant to the graph classification task. 
The problem of discriminative subgraph selection using multiple side views is to find an optimal set of subgraph patterns $\mathcal{T}$ for graph classification by utilizing the side views $\mathcal{Z}$. This is non-trivial due to the following problems:
\begin{description}[leftmargin=*,noitemsep,topsep=0pt]
\item[(P1)] How can we leverage the valuable information embedded in multiple side views to evaluate the usefulness of a set of subgraph patterns?
\item[(P2)] How can we efficiently search for the optimal subgraph patterns without exhaustive enumeration in the graph space?
\end{description}

\section{Data Analysis}
\label{sec:subgraph_data}


In this work, we use the dataset collected from the Chicago Early HIV Infection Study at Northwestern University \cite{ragin2012structural}. The clinical cohort includes 56 {\hiv} (positive) and 21 seronegative controls (negative). This dataset contains brain networks that are constructed from {\fmri} and {\dti}, respectively.

For {\fmri} data, we used the {\dparsf} toolbox\footnote{\url{http://rfmri.org/DPARSF}} and applied the same preprocessing steps following \cite{kong2013discriminative}. For {\dti} data, we used the {\fsl} toolbox\footnote{\url{http://fsl.fmrib.ox.ac.uk/fsl/fslwiki}} to extract the brain networks. The processing pipeline consists of the following steps: (1) correct the distortions induced by eddy currents in the gradient coils and use affine registration to a reference volume for head motion, (2) delete non-brain tissue from the image of the whole head \cite{smith2002fast,jenkinson2005bet2}, (3) fit the diffusion tensor model at each voxel, (4) build up distributions on diffusion parameters at each voxel, and (5) repetitively sample from the distributions of voxel-wise principal diffusion directions. As with the {\fmri} data, the Automated Anatomical Labeling ({\aal}) was propagated to parcellate the {\dti} images into 90 regions (45 for each hemisphere) \cite{tzourio2002automated}. We preprocessed the link weights by min-max normalization.

In addition, for each subject, seven groups of features were documented, including {\neu}, {\flo}, {\pla}, {\fre}, {\ave}, {\loc}, {\seg}. Each group can be regarded as a distinct side view that partially reflects the health state of a subject, and measurements from different medical examinations usually provide complementary information. We preprocessed these auxiliary features by min-max normalization before employing the {\rbf} kernel on each side view.

Next, we study the potential impact of side information on selecting subgraph patterns. Side information consistency suggests that the similarity of side view features between instances with the same label should have higher probability to be larger than that with different labels. We use statistical hypothesis testing to validate whether this statement holds in our data collection.

For each side view, we first construct two vectors $\mathbf{a}_s^{(p)}$ and $\mathbf{a}_d^{(p)}$ with an equal number of elements sampled from the sets $\mathcal{A}_s^{(p)}=\{\kappa^{(p)}(i,j)|y_iy_j=1\}$ and $\mathcal{A}_d^{(p)}=\{\kappa^{(p)}(i,j)|y_iy_j=-1\}$, respectively. Then, we form a two-sample one-tail t-test to validate the existence of side information consistency. Let $\mu_s^{(p)}$ and $\mu_d^{(p)}$ denote the sample means of similarity scores in the two groups, respectively. The null hypothesis is $H_0: \mu_s^{(p)}-\mu_d^{(p)}\leq0$, and the alternative hypothesis is $H_1: \mu_s^{(p)}-\mu_d^{(p)}>0$. In this manner, we test whether there is sufficient evidence to support the hypothesis that the similarity score in $\mathbf{a}_s^{(p)}$ is larger than that in $\mathbf{a}_d^{(p)}$. 

\begin{table}[t]
\centering
\caption{Hypothesis testing for side information consistency.}
\label{tab:result_ttest}
\begin{tabular}{||l|c|c||}
\hline
Side views	&{\fmri} dataset &{\dti} dataset\\
\hline\hline
{\neu}  &1.3220e-20	&3.6015e-12 \\
{\flo}  &5.9497e-57	&5.0346e-75 \\
{\pla}  &9.8102e-06	&7.6090e-06 \\
{\fre}  &2.9823e-06	&1.5116e-03 \\
{\ave}  &1.0403e-02	&8.1027e-03 \\
{\loc}  &3.1108e-04	&5.7040e-04 \\
{\seg}  &2.0024e-04	&1.2660e-02 \\
\hline
\end{tabular}
\end{table}

The t-test results, p-values, are summarized in \ref{tab:result_ttest}. The results show that there is strong evidence, with significance level $\alpha=0.05$, to reject the null hypothesis on both of the {\fmri} and {\dti} samples. In other words, we validate the existence of side information consistency in our data collection, thereby paving the way for our next study of leveraging multiple side views for discriminative subgraph selection.

\section{Proposed Method}
\label{sec:subgraph_method}

In this section, we first introduce the optimization framework for selecting discriminative subgraph features using multiple side views. Next, we describe our subgraph mining strategy using the evaluation criterion derived from the optimization solution.

\subsection{Modeling Side Views}

We address the problem (P1) discussed in Section~\ref{sec:subgraph_problem} by formulating the discriminative subgraph selection problem as a general optimization framework
\begin{equation}
\begin{aligned}
\label{eq:argmin1}
\mathcal{T}^*=\operatornamewithlimits{argmin}_{\mathcal{T}\subseteq\mathcal{S}}\mathcal{F}(\mathcal{T})~~~\text{s.t.}~|\mathcal{T}|\le k
\end{aligned}
\end{equation}
where $|\cdot|$ denotes the cardinality, and $k$ is the maximum number of selected subgraph patterns. $\mathcal{F}(\mathcal{T})$ is the evaluation criterion to estimate the score (the lower the better) of a subset of subgraph patterns $\mathcal{T}\subseteq\mathcal{S}$, and $\mathcal{T}^*$ denotes the optimal set of subgraph patterns.

Following the observation in Section~\ref{sec:subgraph_data} that the side view information is clearly correlated with the given label information, we assume that the set of optimal subgraph patterns should have the following property: the distance between instances in the space of subgraph features should be consistent with that in the space of a side view. That is to say, if two instances are similar in the space of the $p$-th view (\emph{i.e.}, a high $\kappa^{(p)}(i,j)$ value), they should also be close to each other in the space of subgraph features (\emph{i.e.}, a small distance between subgraph feature vectors). On the other hand, if two instances are dissimilar in the space of the $p$-th view (\emph{i.e.}, a low $\kappa^{(p)}(i,j)$ value), they should be far away from each other in the space of subgraph features (\emph{i.e.}, a large distance between subgraph feature vectors). Therefore, our objective function could be to minimize the distance between subgraph features of similar instances in each side view and maximize the distance between dissimilar instances. This idea is formulated as
\begin{equation}
\begin{aligned}
\label{eq:J1}
\operatornamewithlimits{argmin}_{\mathcal{T}\subseteq\mathcal{S}}\frac{1}{2}\sum_{p=1}^v\lambda^{(p)}\sum_{i,j=1}^n
\|\mathbf{I}_\mathcal{T}\mathbf{x}_i-\mathbf{I}_\mathcal{T}\mathbf{x}_j\|^2_\mathrm{F}\Theta^{(p)}(i,j)
\end{aligned}
\end{equation}
where $\mathbf{I}_\mathcal{T}$ is a diagonal matrix indicating which subgraph features are selected into $\mathcal{T}$ from $\mathcal{S}$, and $(\mathbf{I}_\mathcal{T})_{ii}=1$ if $g_i\in\mathcal{T}$, otherwise $(\mathbf{I}_\mathcal{T})_{ii}=0$. The parameters $\lambda^{(p)}\ge0$ are employed to control the contribution per side view.
\begin{equation}
\begin{aligned}
\label{eq:theta}
\Theta^{(p)}(i,j)
=\left\{
\begin{array}{ll}
    \frac{1}{|\mathcal{H}^{(p)}|}~&~(i,j)\in\mathcal{H}^{(p)}\\
    -\frac{1}{|\mathcal{L}^{(p)}|}~&~(i,j)\in\mathcal{L}^{(p)}
\end{array}
\right.
\end{aligned}
\end{equation}
where $\mathcal{H}^{(p)}=\{(i,j)|\kappa^{(p)}(i,j)\ge\mu^{(p)}\}$, $\mathcal{L}^{(p)}=\{(i,j)|\kappa^{(p)}(i,j)<\mu^{(p)}\}$, and $\mu^{(p)}=\frac{1}{n^2}\sum_{i,j=1}^n\kappa^{(p)}(i,j)$ is the mean value of $\kappa^{(p)}(i,j)$. This normalization is to balance the effect of similar instances and dissimilar instances.

Intuitively, \ref{eq:J1} will minimize the distance between subgraph features of similar instances with $\kappa^{(p)}(i,j)\ge\mu^{(p)}$ and maximize the distance between dissimilar instances with $\kappa^{(p)}(i,j)<\mu^{(p)}$ for each side view. In this manner, the side view information is effectively used to guide the procedure of discriminative subgraph selection. It can be used in a semi-supervised setting, for graph instances with or without labels, as long as the side views are available.

For the labeled graphs only, we further consider that the optimal set of subgraph patterns should satisfy the following constraint: labeled graphs in the same class should be close to each other in the subgraph space, and labeled graphs in different classes should be far away from each other in the subgraph space. Intuitively, this constraint tends to select the most discriminative subgraph patterns based on the graph labels. Such an idea has been well explored in the context of dimensionality reduction and feature selection \cite{bar2005learning,tang2006pairwise}. It can be mathematically formulated as minimizing the loss function
\begin{equation}
\begin{aligned}
\label{eq:J2}
\operatornamewithlimits{argmin}_{\mathcal{T}\subseteq\mathcal{S}}\frac{1}{2}\sum_{i,j=1}^n\|\mathbf{I}_\mathcal{T}\mathbf{x}_i-\mathbf{I}_\mathcal{T}\mathbf{x}_j\|^2_\mathrm{F}\Omega(i,j)
\end{aligned}
\end{equation}
where
\begin{equation}
\begin{aligned}
\label{eq:omega}
\Omega(i,j)
=\left\{
\begin{array}{ll}
    \frac{1}{|\mathcal{M}|}~&~(i,j)\in\mathcal{M}\\
    -\frac{1}{|\mathcal{C}|}~&~(i,j)\in\mathcal{C}\\
    0~&~\text{otherwise}
\end{array}
\right.
\end{aligned}
\end{equation}
and $\mathcal{M}=\{(i,j)|y_i y_j=1\}$ denotes the set of pairwise constraints between graphs with the same label, and $\mathcal{C}=\{(i,j)|y_i y_j=-1\}$ denotes the set of pairwise constraints between graphs with different labels.

By defining the matrix $\Phi\in\mathbb{R}^{n\times n}$ as
\begin{equation}
\begin{aligned}
\label{eq:W}
\Phi(i,j)=\Omega(i,j)+\sum_{p=1}^v\lambda^{(p)}\Theta^{(p)}(i,j)
\end{aligned}
\end{equation}
we can combine and rewrite the function in \ref{eq:J1} and \ref{eq:J2} as
\begin{equation}
\begin{aligned}
\label{eq:J}
\mathcal{F}(\mathcal{T})&=\frac{1}{2}\sum_{i=1}^n\sum_{j=1}^n\|\mathbf{I}_\mathcal{T}\mathbf{x}_i-\mathbf{I}_\mathcal{T}\mathbf{x}_j\|^2_\mathrm{F}\Phi(i,j)\\
&=\text{tr}(\mathbf{I}^{\top}_\mathcal{T}\mathbf{X}(\mathbf{D}-\Phi)\mathbf{X}^{\top}\mathbf{I}_\mathcal{T})\\
&=\text{tr}(\mathbf{I}^{\top}_\mathcal{T}\mathbf{X}\mathbf{L}\mathbf{X}^{\top}\mathbf{I}_\mathcal{T})\\
&=\sum_{g_i\in\mathcal{T}}\mathbf{f}_i^\top \mathbf{L}\mathbf{f}_i
\end{aligned}
\end{equation}
where $\text{tr}(\cdot)$ is the trace of a matrix, $\mathbf{D}$ is a diagonal matrix whose elements are column sums of $\Phi$, \emph{i.e.}, $\mathbf{D}(i,i)=\sum_{j}\Phi(i,j)$, and $\mathbf{L}=\mathbf{D}-\Phi$ is a Laplacian matrix.

\begin{Definition}[\gscore]\label{def:q}
Let $\mathcal{D}=\{G_1,\cdots,G_n\}$ denote a graph dataset with multiple side views. Suppose $\Phi$ is a matrix defined as \ref{eq:W}, $\mathbf{D}$ is a diagonal matrix with $\mathbf{D}(i,i)=\sum_{j}\Phi(i,j)$, and $\mathbf{L}$ is a Laplacian matrix defined as $\mathbf{L}=\mathbf{D}-\Phi$. We define an evaluation criterion $q$, named {\gscore}, for a subgraph pattern $g_i$ as
\begin{equation}
\begin{aligned}
\label{eq:q}
q(g_i)=\mathbf{f}_i^\top \mathbf{L}\mathbf{f}_i
\end{aligned}
\end{equation}
\end{Definition}

Since the Laplacian matrix $\mathbf{L}$ is positive semi-definite, for any subgraph pattern $g_i$, $q(g_i)\ge0$. Based on \ref{eq:J} and \ref{eq:q}, the optimization problem in \ref{eq:argmin1} can be rewritten as
\begin{equation}
\begin{aligned}
\label{eq:argmin2}
\mathcal{T}^*=\operatornamewithlimits{argmin}_{\mathcal{T}\subseteq\mathcal{S}}\sum_{g_i\in\mathcal{T}}q(g_i)~~~\text{s.t.}~|\mathcal{T}|\le k
\end{aligned}
\end{equation}

Suppose the {\gscore} values for all the subgraph patterns in $\mathcal{S}$ are sorted as $q(g_1)\le\cdots\le q(g_m)$, then the optimal solution to the problem in \ref{eq:argmin2} is
\begin{equation}
\begin{aligned}
\label{eq:T}
\mathcal{T}^*=\operatornamewithlimits{\cup}_{i=1}^k\{g_i\}
\end{aligned}
\end{equation}

\subsection{Pruning the Search Space}

Now we address the problem (P2) discussed in Section~\ref{sec:subgraph_problem} and propose an efficient method to find the optimal set of subgraph patterns from a graph dataset with multiple side views.

A straightforward solution to the goal of finding an optimal feature set is exhaustive enumeration. That is to say, we need to first enumerate all subgraph patterns from a graph dataset and then calculate the {\gscore} value for each subgraph pattern. In the context of graph data, however, it is usually not feasible to enumerate the full set of subgraph patterns before feature selection. Actually, the number of subgraph patterns grows exponentially with the size of graphs. Inspired by the recent advances in graph classification approaches \cite{cao2015identification,kong2010multi,kong2010semi,yan2008mining} which nest their evaluation criteria into the subgraph mining process and use constraints to prune the search space, we adopt such an idea and develop a different constraint based upon {\gscore}. In particular, we can enumerate all the subgraph patterns from a graph dataset using the {\gspan} algorithm \cite{yan2002gspan}. It recursively searches for the next subgraph in a canonical search space, called {\dfs} code trees. In order to prune the subgraph search space, we can derive a lower bound of the {\gscore} value:
\begin{Theorem}
\label{theorem:bound}
Given any two subgraph patterns $g_i,g_j\in\mathcal{S}$, $g_j$ is a supergraph of $g_i$, i.e., $g_i\subseteq g_j$, and $\hat{q}(g_i)$ is defined as
\begin{equation}
\begin{aligned}
\label{eq:bound}
\hat{q}(g_i)=\mathbf{f}_i^\top\hat{\mathbf{L}}\mathbf{f}_i
\end{aligned}
\end{equation}
where the matrix $\hat{\mathbf{L}}$ is defined as $\hat{\mathbf{L}}(p,q)=\min(0,\mathbf{L}(p,q))$, then the {\gscore} value of $g_j$ is bounded by $\hat{q}(g_i)$. i.e., $q(g_j)\ge\hat{q}(g_i)$.
\end{Theorem}
\textsc{Proof}.
According to Definition~\ref{def:q},
\begin{equation}
\begin{aligned}
\label{eq:proof1}
q(g_j)=\mathbf{f}_j^\top \mathbf{L}\mathbf{f}_j=\sum_{p,q:G_p,G_q\in\mathcal{G}(g_j)}\mathbf{L}(p,q)
\end{aligned}
\end{equation}
where $\mathcal{G}(g_j)=\{G_k|g_j\subseteq G_k,1\le k\le n\}$. Because $g_i\subseteq g_j$, according to anti-monotonic property, we have $\mathcal{G}(g_j)\subseteq\mathcal{G}(g_i)$. Because $\hat{\mathbf{L}}(p,q)=\min(0,\mathbf{L}(p,q))$, we have $\hat{\mathbf{L}}(p,q)\le \mathbf{L}(p,q)$ and $\hat{\mathbf{L}}(p,q)\le0$. Therefore,
\begin{equation}
\begin{aligned}
\label{eq:proof2}
q(g_j)=\sum_{p,q:G_p,G_q\in\mathcal{G}(g_j)}\mathbf{L}(p,q)\ge\sum_{p,q:G_p,G_q\in\mathcal{G}(g_j)}\hat{\mathbf{L}}(p,q)\ge\sum_{p,q:G_p,G_q\in\mathcal{G}(g_i)}\hat{\mathbf{L}}(p,q)=\hat{q}(g_i)
\end{aligned}
\end{equation}
Thus, for any $g_i\subseteq g_j$, $q(g_j)\ge\hat{q}(g_i)$. $\square$

\begin{algorithm}[t]
\caption{\gmsv}
\label{alg:subgraph}
\begin{algorithmic}[1]
\REQUIRE $\mathcal{D}$ (graph dataset), $\mathcal{Z}$ (side views), $\{\lambda^{(p)}\}_{p=1}^{v}$ (view coefficients), $k$ (number of subgraph patterns to be selected), $\ms$ (minimum support value)
\ENSURE $\mathcal{T}$ (selected subgraph patterns)
\STATE Initialize $\mathcal{T}=\emptyset,~\theta=\texttt{Inf}$
\WHILE{$\text{unexplored nodes in {\dfs} code trees}\ne\emptyset$}
\STATE $g=\text{currently explored node in {\dfs} code trees}$
\IF{$\text{frequency}(g)\ge \ms$}
\IF{$|\mathcal{T}|<k~\text{or}~q(g)<\theta$}
\STATE $\mathcal{T}=\mathcal{T}\cup\{g\}$
\IF{$|\mathcal{T}|>k$}
\STATE $g_{max}=\operatornamewithlimits{argmax}_{g'\in\mathcal{T}}q(g')$
\STATE $\mathcal{T}=\mathcal{T}/\{g_{max}\}$
\ENDIF
\STATE $\theta=\max_{g'\in\mathcal{T}}q(g')$
\ENDIF
\IF{$\hat{q}(g)<\theta$}\label{line:if}
\STATE $\text{Depth-first search the subtree rooted from}~g$
\ENDIF
\ENDIF
\ENDWHILE
\end{algorithmic}
\end{algorithm}
We can use this lower bound of {\gscore} in the subgraph mining steps of {\gspan} to efficiently prune the subgraph search space. During the depth-first search of subgraph patterns, we always maintain the top-$k$ best subgraph patterns according to {\gscore} denoted as $\mathcal{T}$ and the temporally suboptimal {\gscore} value denoted as $\theta=\max_{g'\in\mathcal{T}}q(g')$. If $\hat{q}(g_i)\ge\theta$, the {\gscore} value of any supergraph $g_j$ of $g_i$ should be no less than $\hat{q}(g_i)$ according to Theorem~\ref{theorem:bound}, \emph{i.e.}, $q(g_j)\ge\hat{q}(g_i)\ge\theta$. Thus, we can safely prune the subtree rooted from $g_i$ in the search space. If $\hat{q}(g_i)<\theta$, we cannot prune this subtree since there might exist a supergraph $g_j$ of $g_i$ such that $q(g_j)<\theta$. As long as a subgraph $g_i$ can improve the {\gscore} value of any subgraph in $\mathcal{T}$, it is added into $\mathcal{T}$ and the least best subgraph is removed from $\mathcal{T}$. The branch-and-bound algorithm, named {\gmsv}, is summarized in Algorithm~\ref{alg:subgraph}. 
The implementation has been made available at {\github}\footnote{\url{https://github.com/caobokai/gMSV}}.

\section{Experiments}
\label{sec:subgraph_exp}

In order to evaluate the performance of the proposed solution to the problem of subgraph selection for graph classification using multiple side views, we conduct experiments on the {\fmri} and {\dti} brain network datasets introduced in Section~\ref{sec:subgraph_data}.

\subsection{Compared Methods}

We compare the proposed method with other methods that use different statistical measures and discrimination functions to evaluate subgraph features. For all the compared methods, {\gspan} \cite{yan2002gspan} is used as the underlying searching strategy. Note that although alternative algorithms are available \cite{yan2008mining,jin2009graph,jin2010gaia}, the search step efficiency is not the focus of this work. The compared methods are summarized as follows:
\begin{itemize}[leftmargin=*,noitemsep,topsep=0pt]
\item\textbf{\gmsv}: the proposed discriminative subgraph selection method using multiple side views. Following the observation in Section~\ref{sec:subgraph_data} that side information consistency is verified to be significant in all the side views, the view coefficients in {\gmsv} are simply set to $\lambda^{(1)}=\cdots=\lambda^{(v)}=1$. In the case where some side views are suspect to be redundant, we can adopt an alternating optimization strategy to iteratively select discriminative subgraph patterns and update view coefficients.
\item\textbf{\gssc}: a semi-supervised subgraph selection method based upon both labeled and unlabeled graphs \cite{kong2010semi}. The parameters in gSSC are set to $\alpha=\beta=1$.
\item\textbf{\duga}, \textbf{\dugb}, \textbf{\dugc}, and \textbf{\dugd}: supervised subgraph selection methods based upon confidence \cite{gao2010direct}, frequency ratio \cite{jin2011lts,jin2010gaia,jin2009graph}, G-test score \cite{yan2008mining}, and Hilbert-Schmidt Independence Criterion \cite{kong2010multi}, respectively. 
\item \textbf{\topk}: an unsupervised subgraph selection method based upon frequency. 
\end{itemize}

For a fair comparison, we append the side view data to the subgraph-based graph representations produced from the compared methods before feeding the concatenated features to a standard classifier which is {\libsvm} \cite{libsvm} with a linear kernel. Another baseline that uses only side view features is denoted as {\side}. In the experiments, 3-fold cross validation is performed on balanced datasets. To obtain binary link weights, a threshold of 0.9 and 0.3 is applied on the {\fmri} and {\dti} datasets, respectively.

\subsection{Performance on Graph Classification}

\begin{figure}[t]
\centering
\begin{minipage}[l]{\columnwidth}
  \centering
  \includegraphics[width=1\textwidth]{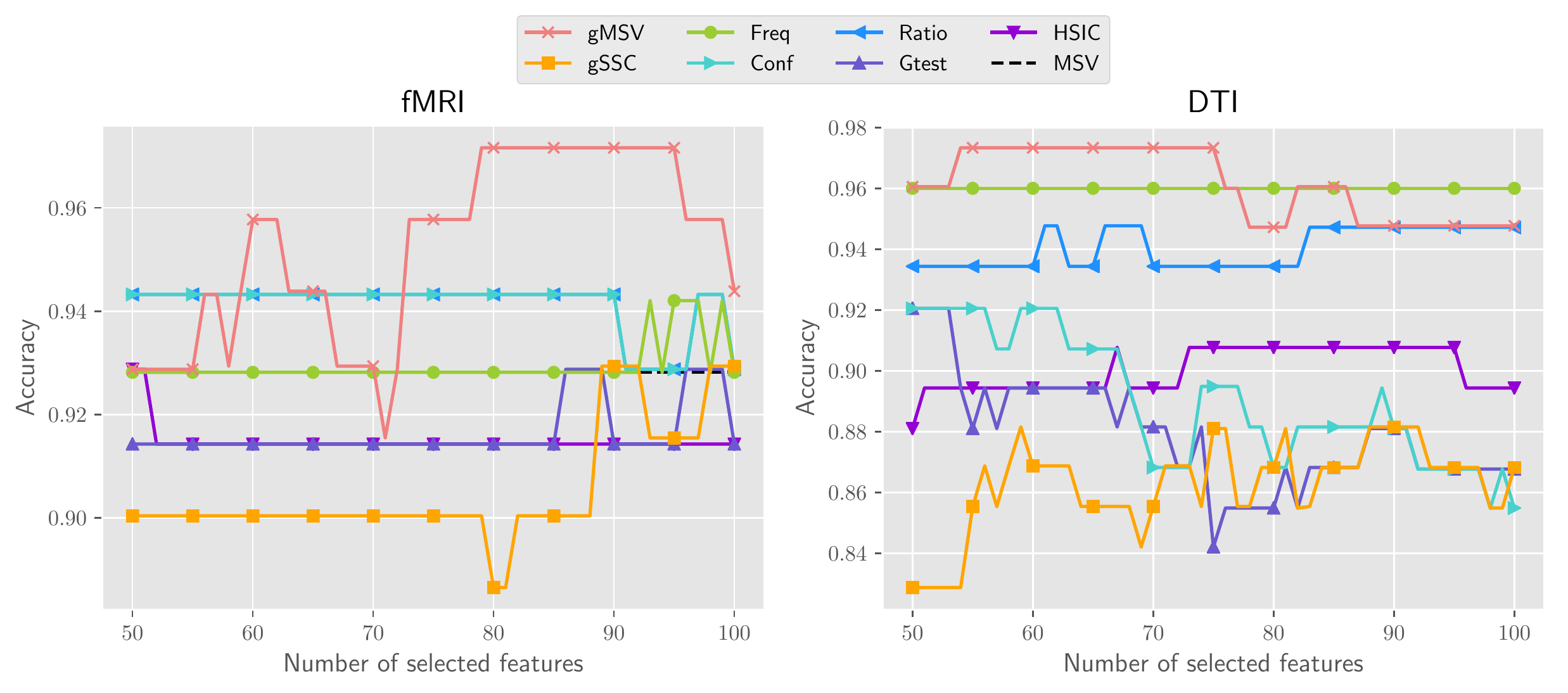}
\end{minipage}
\caption{Classification performance with varying number of subgraphs.}
\label{fig:subgraph_acc}
\end{figure}

Experimental results on the {\fmri} and {\dti} datasets are shown in \ref{fig:subgraph_acc}. The average performance with different number of selected subgraph features is reported for each method. Classification accuracy is used as the evaluation metric.

On the {\fmri} dataset, the proposed method {\gmsv} can achieve the classification accuracy as high as 97.16\% which is significantly better than the union of other subgraph features and side view features. The black dashed line denotes the baseline {\side} that uses only side view data. {\duga} and {\dugb} perform slightly better than {\side}. {\topk} has a comparable performance with {\side}, which indicates that there is no additional information added from the selected frequent subgraphs. Other methods that use different discrimination functions without leveraging the guidance from side views perform even worse than {\side} on graph classification, because they evaluate the usefulness of subgraph patterns solely based on the limited label information from a small sample size of graph instances. The selected subgraph patterns can potentially be redundant or irrelevant, thereby compromising the effects of side view data. Importantly, {\gmsv} outperforms the semi-supervised approach {\gssc} which explores the unlabeled graphs based on the separability property. This indicates that rather than enforcing unlabeled graphs to be separated from each other in the subgraph space, it would be better to regularize such separability to be consistent with the available side views.
 
Similar observations can be found on the {\dti} dataset where {\gmsv} outperforms other baselines by achieving a good performance as high as 97.33\% accuracy. We notice that only {\gmsv} is able to outperform {\side} by adding complementary subgraph features to the side view features.
It supports our premise that leveraging multiple side views can boost the performance of graph classification, and the evaluation criterion {\gscore} used in {\gmsv} can find more informative subgraph patterns for graph classification than subgraphs based on frequency or other discrimination scores.

\begin{figure}[t]
\centering
\begin{minipage}[l]{\columnwidth}
  \centering
  \includegraphics[width=1\textwidth]{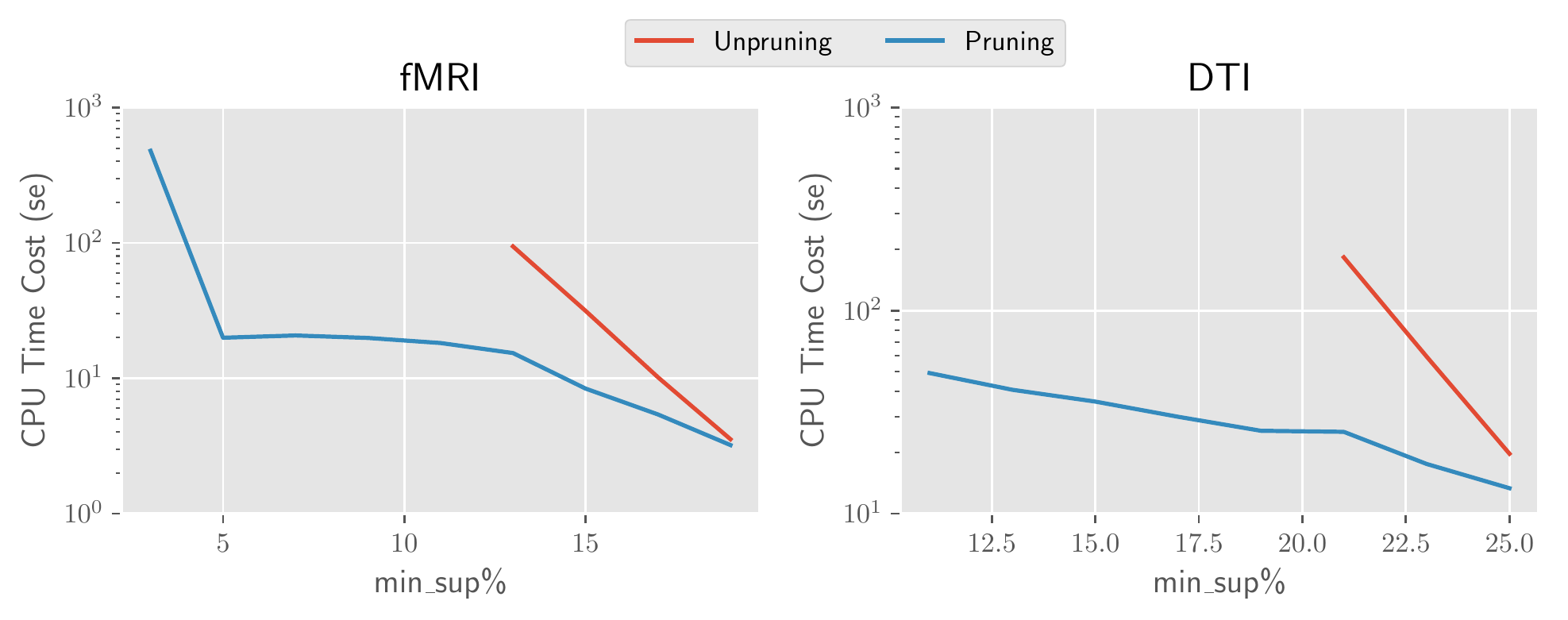}
\end{minipage}
\caption{CPU time with varying support value.}
\label{fig:time}
\end{figure}

\begin{figure}[t]
\centering
\begin{minipage}[l]{\columnwidth}
  \centering
  \includegraphics[width=1\textwidth]{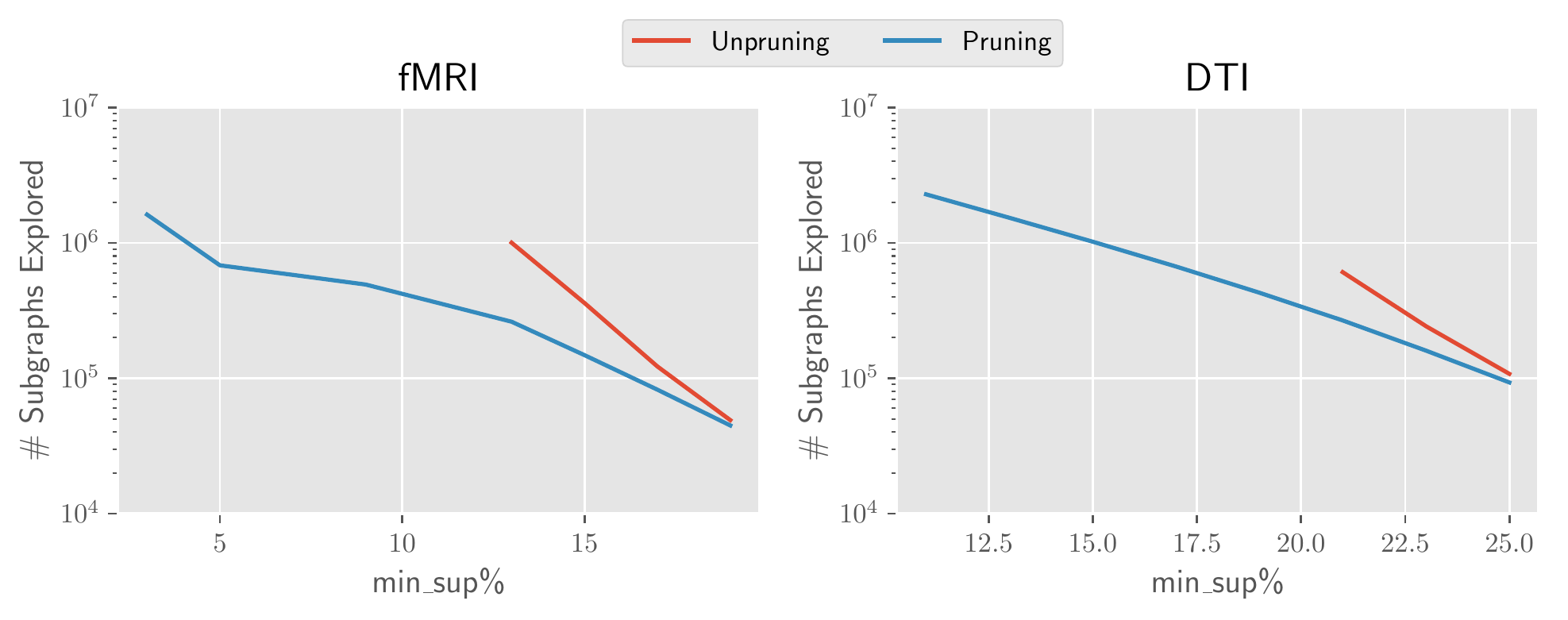}
\end{minipage}
\caption{Number of enumerated subgraphs with varying support value.}
\label{fig:num_fea}
\end{figure}

\begin{table}[t]
\centering
\small
\caption{Classification performance with different side views.}
\label{tab:subgraph_result}
\newcolumntype{x}[1]{>{\centering\arraybackslash}p{#1}}
\begin{tabular}{||c|l|x{1.5cm}|x{1.5cm}|x{1.5cm}|x{1.5cm}||}
\hline
Brain networks &
Side views &Accuracy &Precision &Recall &{\fone} \\
\hline\hline
\multirow{8}*{\fmri}
&{\neu}	&0.743 &0.851 &0.679 &0.734 \\
&{\flo}	&0.887 &0.919 &0.872 &0.892 \\
&{\pla}	&0.715 &0.769 &0.682 &0.710 \\
&{\fre}	&0.786 &0.851 &0.737 &0.785 \\
&{\ave}	&0.672 &0.824 &0.500 &0.618 \\
&{\loc}	&0.628 &0.686 &0.605 &0.637 \\
&{\seg}	&0.701 &0.739 &0.737 &0.731 \\
\cline{2-6}
&All side views	&0.972 &1.000 &0.949 &0.973 \\
\hline\hline
\multirow{8}*{\dti}
&{\neu} &0.616 &0.630 &0.705 &0.662\\
&{\flo}	&0.815 &0.847 &0.808 &0.822\\
&{\pla}	&0.736 &0.801 &0.705 &0.744\\
&{\fre}	&0.631 &0.664 &0.632 &0.644\\
&{\ave}	&0.604 &0.626 &0.679 &0.647\\
&{\loc}	&0.723 &0.717 &0.775 &0.741\\
&{\seg}	&0.605 &0.616 &0.679 &0.644\\
\cline{2-6}
&All side views	&0.973 &1.000 &0.951 &0.974 \\
\hline
\end{tabular}
\end{table}

\subsection{Time and Space Complexity}

We evaluate the efficiency of pruning the subgraph search space by using the lower bound of {\gscore} in {\gmsv}. In particular, we compare the runtime performance of two implementation versions of {\gmsv}: the pruning one uses the lower bound of {\gscore} to prune the search space when enumerating subgraph patterns, as shown in Algorithm~\ref{alg:subgraph}; the unpruning one performs no pruning in the subgraph mining process, \emph{i.e.}, removing the \texttt{if} statement in Line~\ref{line:if} of Algorithm~\ref{alg:subgraph}. We test both approaches and report the average CPU time used and the average number of subgraph patterns explored during the procedure of subgraph mining and feature selection.

The comparisons with respect to the time complexity and the space complexity are shown in \ref{fig:time} and \ref{fig:num_fea}, respectively. On both datasets, the unpruning {\gmsv} needs to explore an exponentially larger subgraph search space as we decrease the value of {\ms}. When {\ms} is too low, the subgraph enumeration step in the unpruning {\gmsv} runs out of the memory. However, the pruning {\gmsv} is still effective and efficient when {\ms} goes to very low, because its running time and space requirement do not increase as much as the unpruning {\gmsv} by reducing the subgraph search space via the lower bound of {\gscore}.

The focus of this work is to investigate side information consistency and explore multiple side views in discriminative subgraph selection. As potential alternatives to the {\gspan}-based branch-and-bound algorithm, we could employ other more sophisticated searching strategies with our proposed multi-side-view evaluation criterion, {\gscore}. For example, we can substitute {\gscore} for the G-test score in \texttt{LEAP} \cite{yan2008mining}, the log ratio in \texttt{COM} \cite{jin2009graph} and \texttt{GAIA} \cite{jin2010gaia}, \emph{etc.} However, as shown in \ref{fig:time} and \ref{fig:num_fea}, our proposed pruning solution can already survive at $\ms=4\%$. Considering the small sample size problem for brain network analysis, {\gmsv} is efficient enough because subgraph patterns with too few supported graphs are not desired.

\subsection{Effects of Side Views}

We investigate the contributions from different side views. \ref{tab:subgraph_result} shows the performance of {\gmsv} by considering only one side view each time. In general, the best performance is achieved by simultaneously leveraging all the side views. Specifically, we observe that the side view {\flo} can independently provide the most useful information for selecting discriminative subgraph patterns, which might imply that {\hiv} brain injury is most likely to express in measurements from this side view. It is consistent with our finding in Section~\ref{sec:subgraph_data} that the side view {\flo} is the most significantly correlated with the diagnosis.

\begin{figure}[t]
\centering
\subfigure[{\fmri} 1]{
\label{fig:fMRI_1}
\begin{minipage}[l]{0.22\columnwidth}
  \centering
  \includegraphics[width=1\textwidth]{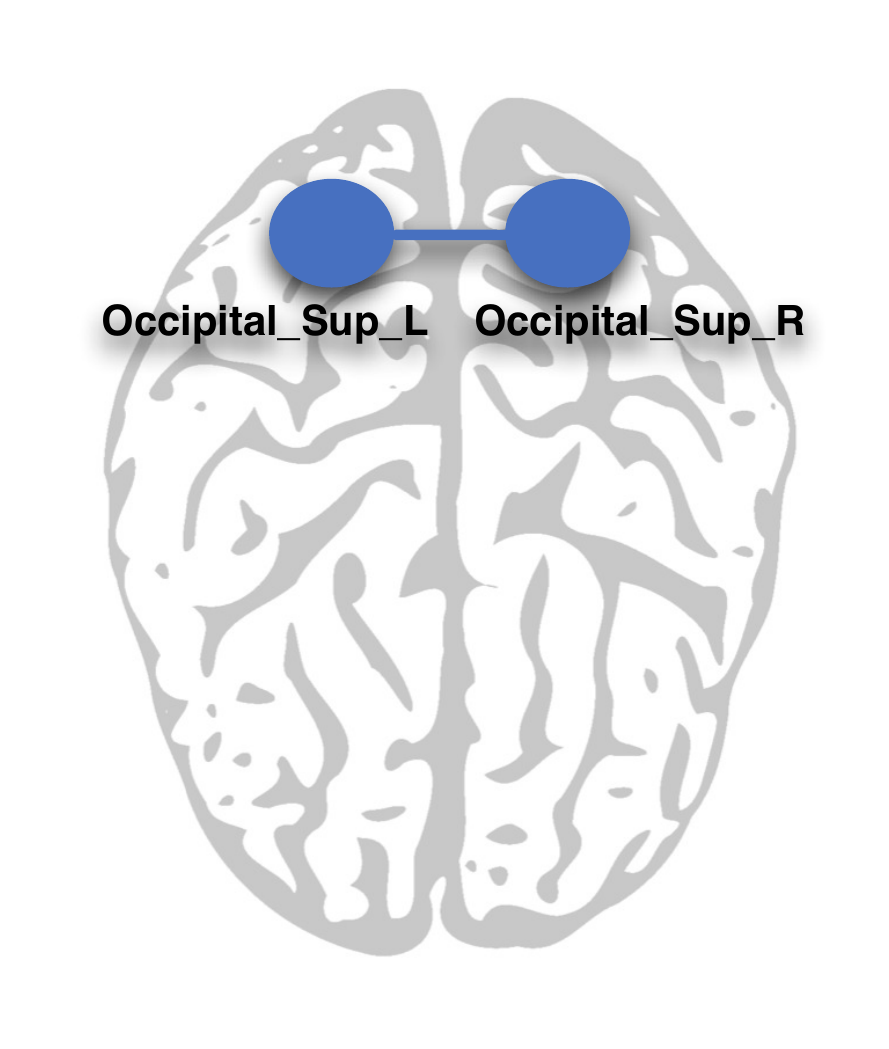}
\end{minipage}
}
\subfigure[{\fmri} 2]{
\label{fig:fMRI_2}
\begin{minipage}[l]{0.22\columnwidth}
  \centering
  \includegraphics[width=1\textwidth]{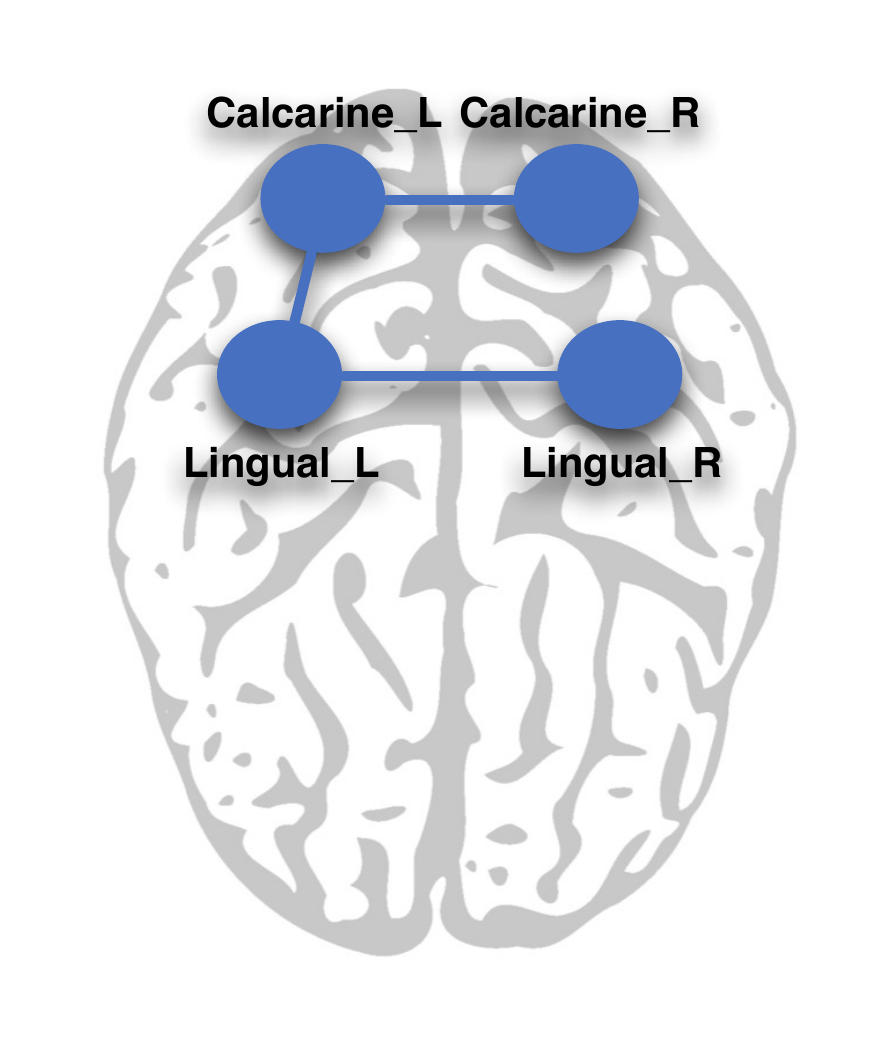}
\end{minipage}
}
\subfigure[{\dti} 1]{
\label{fig:DTI_1}
\begin{minipage}[l]{0.22\columnwidth}
  \centering
  \includegraphics[width=1\textwidth]{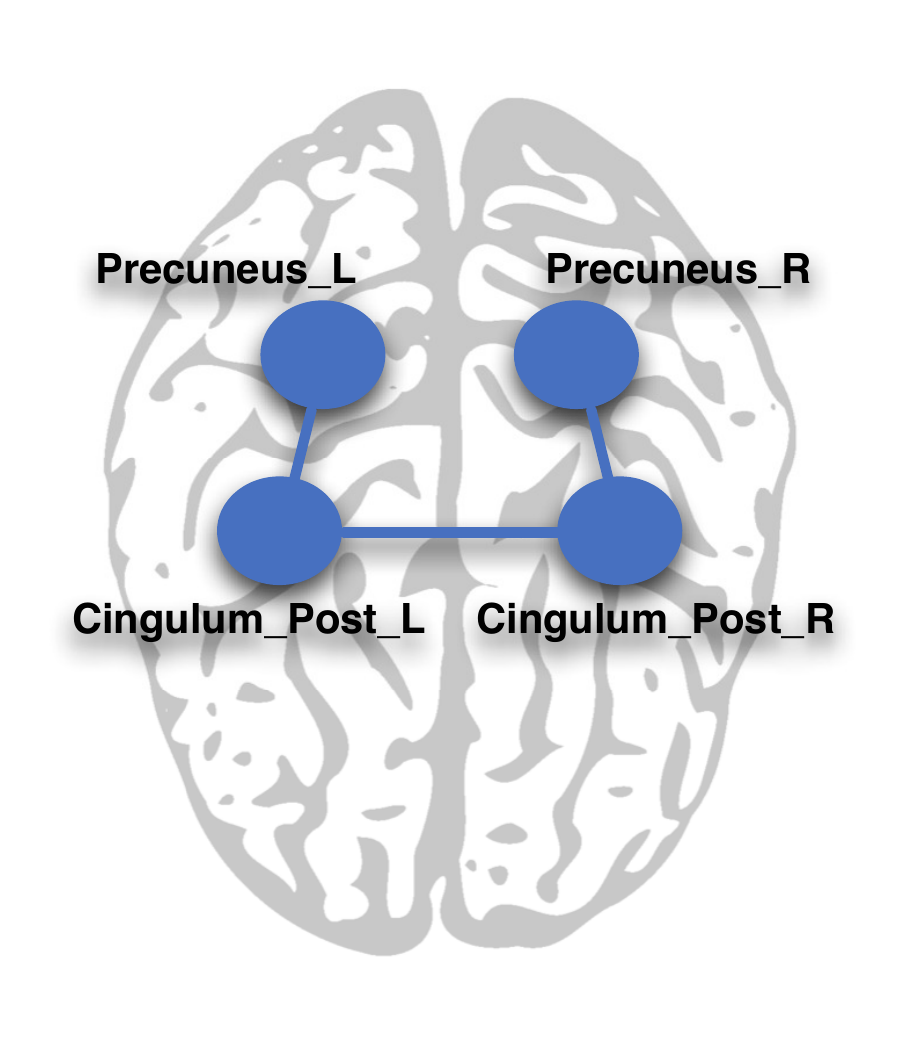}
\end{minipage}
}
\subfigure[{\dti} 2]{
\label{fig:DTI_2}
\begin{minipage}[l]{0.22\columnwidth}
  \centering
  \includegraphics[width=1\textwidth]{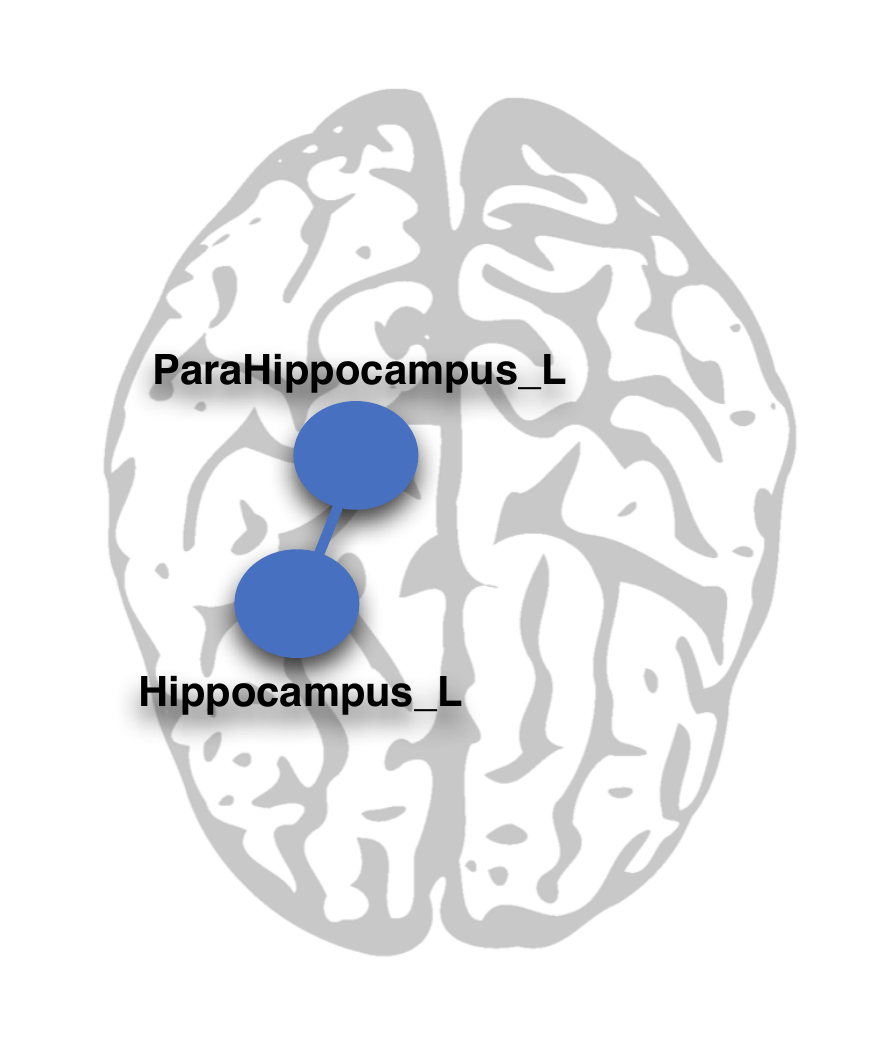}
\end{minipage}
}
\caption{{\hiv}-related subgraph patterns in the human brain.}
\label{fig:subgraph_fea}
\end{figure}

\subsection{Feature Evaluation}

\ref{fig:subgraph_fea} displays the most discriminative subgraph patterns selected by {\gmsv} from the {\fmri} dataset and the {\dti} dataset, respectively. These findings examining functional and structural networks are consistent with other in vivo studies \cite{castelo2006altered,wang2011abnormalities} and with the pattern of brain injury at autopsy \cite{everall1993neuronal,langford2003changing} in {\hiv} infection. With the approach presented in this analysis, alterations in the brain can be detected in initial stages of injury and in the context of clinically meaningful information, such as host immune status and immune response ({\flo}), immune mediators ({\pla}) and cognitive function ({\neu}). This approach optimizes the valuable information inherent in complex clinical datasets. Strategies for combining various sources of clinical information have promising potential for informing an understanding of disease mechanisms, for identification of new therapeutic targets, and for discovery of biomarkers to assess risk and to evaluate response to treatment.

\section{Related Work}

Mining subgraph patterns from graph data has been extensively studied. A typical evaluation criterion is frequency which aims at searching for frequently appearing subgraph patterns in a graph dataset that satisfy a minimum support value. Most of the frequent subgraph mining approaches are unsupervised, \emph{e.g.}, {\gspan} \cite{yan2002gspan}, \texttt{AGM} \cite{inokuchi2000apriori}, \texttt{FSG} \cite{kuramochi2001frequent}, \texttt{MoFa} \cite{borgelt2002mining}, \texttt{FFSM} \cite{huan2003efficient}, and \texttt{Gaston} \cite{nijssen2004quickstart}. In particular, {\gspan} which builds a lexicographic order among graphs is a depth-first search strategy proposed by Yan and Han \cite{yan2002gspan}.

Moreover, other filtering criteria have been proposed in the setting of supervised subgraph mining which examines how to improve the efficiency of identifying discriminative subgraph patterns for graph classification. Yan et al.~introduced two concepts, \emph{structural leap search} and \emph{frequency-descending mining}, and proposed \texttt{LEAP} \cite{yan2008mining} which is one of the first works on discriminative subgraph mining. Thoma et al.~proposed \texttt{CORK} which can yield a near-optimal solution using greedy feature selection \cite{thoma2009near}. Ranu and Singh proposed a scalable approach, called \texttt{GraphSig}, that is capable of mining discriminative subgraphs with a low frequency threshold \cite{ranu2009graphsig}. Jin et al.~proposed \texttt{COM} which takes into account the co-occurrence of subgraph patterns, thereby facilitating the mining process \cite{jin2009graph}. Jin et al.~further proposed an evolutionary computation method, called \texttt{GAIA}, to mine discriminative subgraph patterns using a randomized searching strategy \cite{jin2010gaia}. Our proposed criterion {\gscore} can be combined with these efficient searching algorithms to speed up the procedure of mining discriminative subgraph patterns by substituting it for the G-test score in \texttt{LEAP} \cite{yan2008mining}, the log ratio in \texttt{COM} \cite{jin2009graph} and \texttt{GAIA} \cite{jin2010gaia}, \emph{etc.} Zhu et al.~designed a diversified discrimination score based on the log ratio which can reduce the overlap between selected features \cite{zhu2012graph}. Such an idea may also be integrated into {\gscore} to improve feature diversity. Wu et al.~considered the scenario where one object can be described by multiple graphs generated from different feature views and proposed an evaluation metric to estimate the usefulness and the redundancy of subgraph features across all views \cite{wu2014multi}. In contrast, in this work, we assume that one object can have other data representations as side views in addition to the graph representation.

\chapter{Brain network embedding}
\label{chapter:bne}

(This chapter was partially published as ``Semi-supervised Tensor Factorization for Brain Network Analysis \cite{cao2016semi}'', in \textit{Joint European Conference on Machine Learning and Knowledge Discovery in Databases (ECML/PKDD)}, 2016, Springer. DOI: \url{https://doi.org/10.1007/978-3-319-46128-1_2}. 
This chapter was partially published as ``t-BNE: Tensor-based Brain Network Embedding \cite{cao2017tbne}'', in \textit{Proceedings of the 2017 SIAM International Conference on Data Mining (SDM)}, 2017, SIAM. DOI: \url{https://doi.org/10.1137/1.9781611974973.22}.)

\section{Introduction}

For brain networks analysis, one can first derive features from the graph data and then use conventional machine learning algorithms for downstream tasks. In general, two types of features are usually extracted: graph-theoretical measures \cite{wee2012identification,jie2014integration} and subgraph patterns \cite{kong2013discriminative,cao2015mining}. However, the expressive power of these explicit features is limited, and some tedious feature engineering works are needed. To explore a larger space of potentially informative features to represent brain networks, it is promising to learn latent representations from the brain network data through factorization techniques.
Despite its value and significance, the problem of brain network embedding based on constrained tensor factorization has not been studied in this context so far. There are three major difficulties as follows:
\begin{itemize}[leftmargin=*]
\item Brain networks are undirected graphs in most cases, and thus their corresponding connectivity matrices are symmetric. However, existing factorization techniques are usually designed for general tensors without taking into account such a graph property.
\item Conventional brain network analysis approaches focus on the graph data alone, while ignoring other side information, such as cognitive measures. These measures compose a complementary view for diagnostic purposes. An effective brain network embedding model should be able to use the valuable side information as guidance in the representation learning process.
\item Learning latent representations of brain networks and then training a classifier on them in a two-step manner would make these two procedures independent with each other and fail to introduce the supervision information to the representation learning process. In order to learn discriminative representations, it is desirable to fuse these two procedures together in an end-to-end framework.
\end{itemize}

\begin{figure}[t]
\centering
\begin{minipage}[l]{\columnwidth}
  \centering
  \includegraphics[width=1\textwidth]{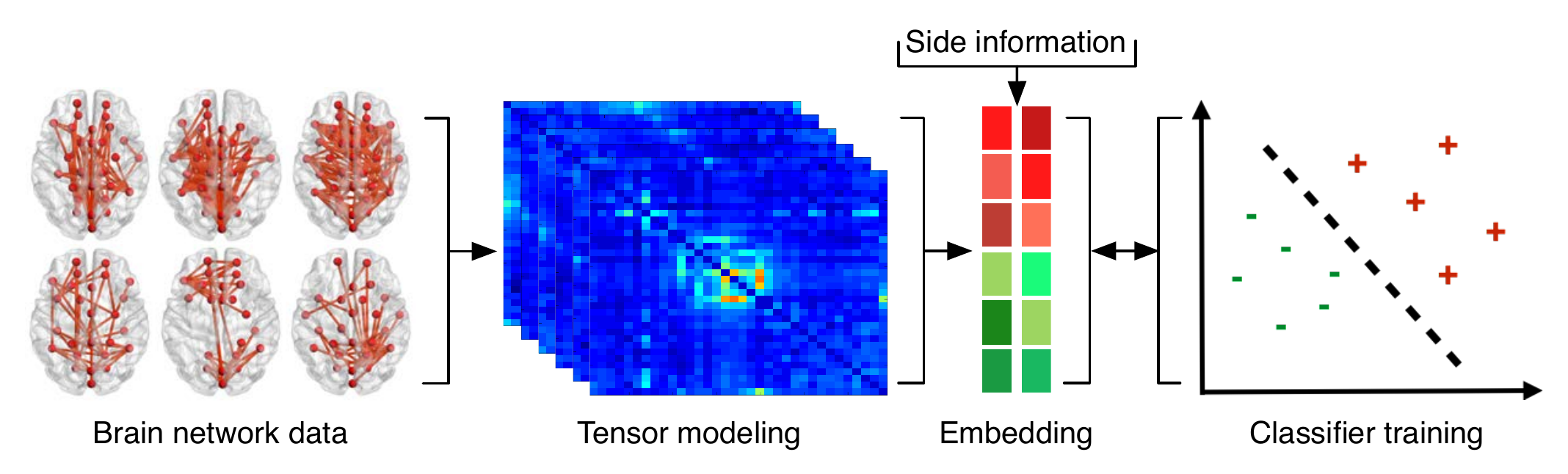}
\end{minipage}
\caption{Tensor-based brain network embedding framework.}
\label{fig:bne_intro}
\end{figure}

In this work, we develop a brain network embedding method based on constrained tensor factorization, named {\bne}, as illustrated in \ref{fig:bne_intro}. The contributions of the proposed method are threefold:
\begin{itemize}[leftmargin=*,noitemsep,topsep=0pt]
\item The brain network embedding problem is modeled as partially symmetric tensor factorization which is suitable for undirected graphs, \emph{e.g.}, electroencephalogram ({\eeg}) brain networks.
\item Self-report data are incorporated as guidance in the tensor factorization procedure to learn latent factors that are consistent with the side information, and an orthogonal constraint is introduced to obtain distinct factors.
\item Representation learning and classifier training are blended into a unified optimization problem, which allows the classifier parameters to interact with the original brain network data via latent factors and the supervision information to be introduced to the representation learning process, so that discriminative representations can be obtained.
\end{itemize}

We evaluate {\bne} on an anxiety disorder
dataset that contains {\eeg} brain networks. Experimental results illustrate the superior performance of the proposed method on graph classification with significant improvement over baselines. Furthermore, the derived factors are visualized which can be informative for investigating disease mechanisms under different emotion regulation tasks.

\section{Problem Formulation}
\label{sec:bne}

Let $\mathcal{D}=\{G_1,\cdots,G_n\}$ denote a graph dataset where $n$ is the number of instances. All graphs in the dataset share a given set of nodes $V$ which can correspond to a brain parcellation scheme. Suppose that the brain is parcellated via an atlas into $|V|=m$ regions, then each brain network $G_i$ can be represented by an adjacency matrix $\mathbf{A}_i \in \mathbb{R}^{m \times m}$.

We may assume without loss of generality that the first $l$ instances within $\mathcal{D}$ are labeled, and $\mathbf{Y} \in \mathbb{R}^{l \times c}$ is the class label matrix where $c$ is the number of class labels. Each instance belongs to one or more classes, where $\mathbf{Y}(i, j) = 1$ if $G_i$ belongs to the $j$-th class, otherwise $\mathbf{Y}(i, j) = 0$. For convenience, the labeled graph dataset is denoted as $\mathcal{D}_l = \{G_1, \cdots , G_l\}$, the unlabeled graph dataset is denoted as $\mathcal{D}_u = \{G_{l+1}, \cdots , G_n\}$, and $\mathcal{D} = \mathcal{D}_l \cup \mathcal{D}_u$.

The problem of brain network embedding can be described as to learn representations $\mathbf{S} \in \mathbb{R}^{n \times k}$ of the brain network data, where each brain network $G_i$ is represented in a $k$-dimensional space with coordinates $\mathbf{S}(i,:)$. It is desirable to let the latent representations be discriminative so that brain networks with different labels can be easily separated. Formally, given $\{\mathbf{S}(i,:)~|~i\in\mathcal{D}_l\}$, the labels of $\{\mathbf{S}(i,:)~|~i\in\mathcal{D}_u\}$ can be correctly classified. Learning such representations is a non-trivial task due to the following problems:
\begin{description}[leftmargin=*,noitemsep,topsep=0pt]
\item[(P1)] How can we model the symmetry property of brain networks?
\item[(P2)] How can we leverage the side information to facilitate the representation learning process?
\item[(P3)] How can we incorporate the classifier training procedure into the representation learning process to develop an end-to-end framework?
\end{description}

\section{Preliminaries}

In addition to Section~\ref{sec:tensor}, an overview of tensor notation and operators is given as follows which will be used throughout this work.

As we have introduced, an $m$th-order tensor can be represented as $\mathcal{X}=(x_{i_1,\cdots,i_m})\in\mathbb{R}^{I_{1}\times\cdots\times I_{m}}$, where $I_i$ is the dimension of $\mathcal{X}$ along the $i$-th mode. An $m$th-order tensor is a rank-one tensor if it can be defined as the tensor product of $m$ vectors. 

\begin{Definition}[Partially Symmetric Tensor]
A rank-one $m$th-order tensor $\mathcal{X}\in\mathbb{R}^{I_{1}\times\cdots\times I_{m}}$ is partially symmetric if it is symmetric on modes $i_1,\cdots,i_j\in\{1,\cdots,m\}$ and can be written as the tensor product of $m$ vectors: $\mathcal{X}=\mathbf{x}^{(1)}\circ\cdots\circ\mathbf{x}^{(m)}$ where $\mathbf{x}^{(i_1)}=\cdots=\mathbf{x}^{(i_j)}$.
\end{Definition}

\begin{Definition}[Mode-$k$ Product]
The mode-$k$ product $\mathcal{X}\times_k\mathbf{A}$ of a tensor $\mathcal{X} \in \mathbb{R}^{I_1\times\cdots\times I_m}$ and a matrix $\mathbf{A} \in \mathbb{R}^{J\times I_k}$ is of size $I_1\times\cdots\times I_{k-1}\times J\times I_{k+1}\times\cdots\times I_m$ and is defined by $(\mathcal{X}\times_k\mathbf{A})_{i_1,\cdots,i_{k-1},j,i_{k+1},\cdots,i_m}=\sum_{i_k=1}^{I_k}x_{i_1,\cdots,i_m}a_{j,i_k}$.
\end{Definition}

\begin{Definition}[Kronecker Product]
The Kronecker product of two matrices $\mathbf{A} \in \mathbb{R}^{I \times J}, \mathbf{B} \in \mathbb{R}^{K \times L}$ is of size $IK \times JL$ and is defined by
\begin{equation}
\begin{aligned}
\mathbf{A}\otimes\mathbf{B}=
\left(
\begin{array}{ccc}
 a_{11}\mathbf{B} & \cdots & a_{1J}\mathbf{B} \\
 \vdots & \ddots & \vdots \\
 a_{I1}\mathbf{B} & \cdots & a_{IJ}\mathbf{B}
\end{array}
\right)
\end{aligned}
\end{equation}
\end{Definition}

\begin{Definition}[Khatri-Rao Product]
The Khatri-Rao product of two matrices $\mathbf{A} \in \mathbb{R}^{I \times K}, \mathbf{B} \in \mathbb{R}^{J \times K}$ is of size $IJ \times K$ and is defined by $\mathbf{A}\odot\mathbf{B}=(a_1\otimes b_1,\cdots,a_K\otimes b_K)$ where $a_1,\cdots,a_K,b_1,\cdots,b_K$ are the columns of matrices $\mathbf{A}$ and $\mathbf{B}$, respectively.
\end{Definition}

\begin{Definition}[Mode-$k$ Matricization]
The mode-$k$ matricization of a tensor $\mathcal{X}\in\mathbb{R}^{I_{1}\times\cdots\times I_{m}}$ is denoted by $\mathbf{X}_{(k)}$ and arranges the mode-$k$ fibers to be the columns of the resulting matrix. The dimension of $\mathbf{X}_{(k)}$ is $\mathbb{R}^{I_k\times J}$, where $J=I_1\cdots I_{k-1}I_{k+1}\cdots I_m$. Each tensor element $(i_1,\cdots,i_m)$ maps to the matrix element $(i_k,j)$: $j=1+\sum_{p=1,p\neq k}^m(i_p-1)J_p~~\text{with}~~J_p=\prod_{q=1,q\neq k}^{p-1}I_q$.
\end{Definition}

\section{Proposed Method}

In this section, we first introduce {\bne}, the proposed brain network embedding solution based on constrained tensor factorization, and describe the optimization framework for solving the objective. Next, we present the time complexity of the solution and discuss some potential variations and extensions of {\bne}.

\subsection{Tensor Modeling}

\begin{figure}[t]
\centering
\begin{minipage}[l]{0.6\columnwidth}
  \centering
  \includegraphics[width=1\textwidth]{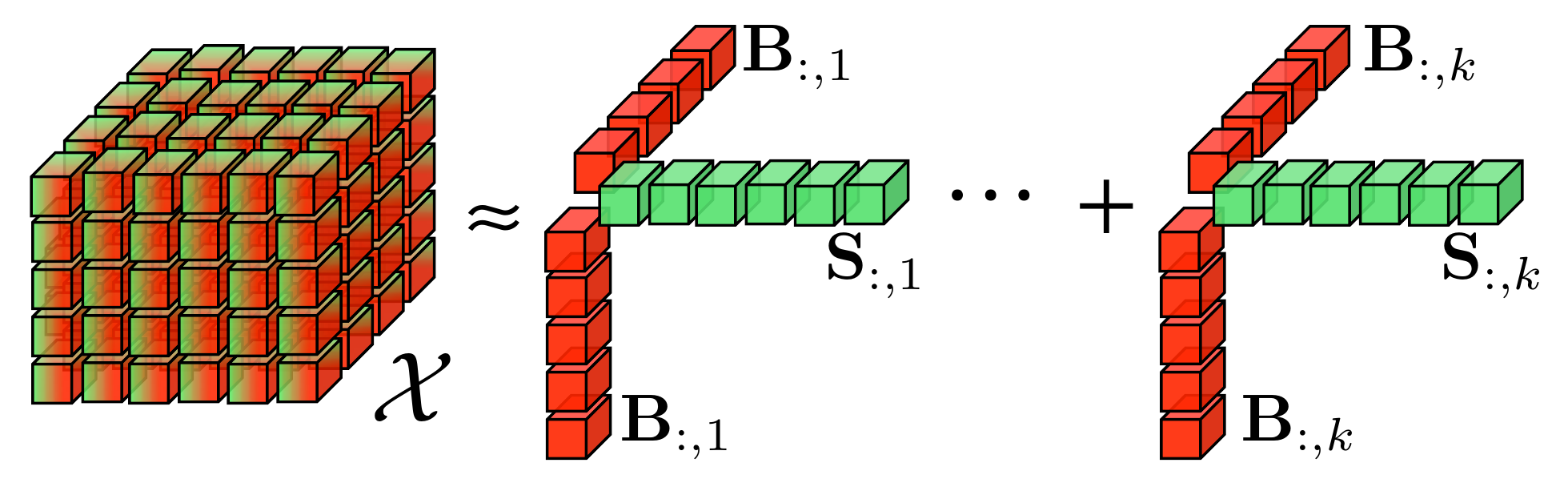}
\end{minipage}
\caption{{\cp} factorization.}
\label{fig:cp}
\end{figure}

We resort to tensor techniques for modeling the brain network data in this work. We first address the problem (P1) discussed in Section~\ref{sec:bne} by stacking brain networks of $n$ subjects, \emph{i.e.}, $\{\mathbf{A}_i\}_{i=1}^n$, as a partially symmetric tensor $\mathcal{X} \in \mathbb{R}^{m \times m \times n}$. We assume that the third-order tensor $\mathcal{X}$ can be decomposed into $k$ factors
\begin{equation}
\begin{aligned}
\label{eq:cp}
\mathcal{X}=\mathcal{C}\times_1\mathbf{B}\times_2\mathbf{B}\times_3\mathbf{S}
\end{aligned}
\end{equation}
where $\mathbf{B} \in \mathbb{R}^{m \times k}$ is the factor matrix for nodes, $\mathbf{S} \in \mathbb{R}^{n \times k}$ is the factor matrix for subjects, and $\mathcal{C} \in \mathbb{R}^{k \times k \times k}$ is the identity tensor, \emph{i.e.}, $\mathcal{C}(i_1,i_2,i_3)=\delta(i_1=i_2=i_3)$. Basically, \ref{eq:cp} is a CANDECOMP/PARAFAC ({\cp}) factorization \cite{kolda2009tensor} as shown in \ref{fig:cp}, where the third-order partially symmetric tensor $\mathcal{X}$ is approximated by $k$ rank-one tensors. The $f$-th factor tensor is the tensor product of three vectors, \emph{i.e.}, $\mathbf{B}(:,f)\circ\mathbf{B}(:,f)\circ\mathbf{S}(:,f)$.

For brain network analysis, auxiliary statistics are usually associated with subjects, \emph{e.g.}, demographic information and cognitive measures \cite{cao2014tensor,zhang2016identifying}. As noted in the previous work \cite{narita2012tensor}, the distance between subjects in the latent space should be consistent with that in the space of auxiliary features. That is to say, if two subjects are similar in the auxiliary space, they should also be close to each other in the latent space (\emph{i.e.}, a small distance between latent factors). Therefore, we address the problem (P2) discussed in Section~\ref{sec:bne} by defining the objective as to minimize the distance between latent factors of similar subjects based upon the side information. It can be mathematically formulated as
\begin{equation}
\begin{aligned}
\label{eq:side1}
\operatornamewithlimits{min}_{\mathbf{S}}~\sum_{i,j=1}^n\|\mathbf{S}(i,:)-\mathbf{S}(j,:)\|^2_\mathrm{F}\mathbf{Z}(i,j)
\end{aligned}
\end{equation}
where $\mathbf{Z}$ is a kernel matrix, and each element $\mathbf{Z}(i,j)$ represents the similarity between $G_i$ and $G_j$ in the space of auxiliary features. A linear kernel is used in this study. In this manner, the side information about subjects is effectively used as guidance to discover meaningful latent factors. We can rewrite \ref{eq:side1} as
\begin{equation}
\begin{aligned}
\label{eq:side2}
\operatornamewithlimits{min}_{\mathbf{S}}~\text{tr}(\mathbf{S}^\top\mathbf{L}_\mathbf{Z}\mathbf{S})
\end{aligned}
\end{equation}
where $\text{tr}(\cdot)$ is the trace of a matrix, $\mathbf{L}_\mathbf{Z}$ is a Laplacian matrix induced from the similarity matrix $\mathbf{Z}$, \emph{i.e.}, $\mathbf{L}_\mathbf{Z}=\mathbf{D}_\mathbf{Z}-\mathbf{Z}$, and $\mathbf{D}_\mathbf{Z}$ is the diagonal matrix whose elements are column sums of $\mathbf{Z}$, \emph{i.e.}, ${\mathbf{D}_\mathbf{Z}}(i,i)=\sum_{j}\mathbf{Z}(i,j)$. It is desirable to discover distinct latent factors to obtain more concise and interpretable results, and thus we include the orthogonal constraint $\mathbf{S}^\top\mathbf{S}=\mathbf{I}$.

In order to make the brain network representations ready for classification purposes, we assume that there is a mapping matrix $\mathbf{W} \in \mathbb{R}^{k \times c}$ which assigns labels to subjects based on the subject factor matrix $\mathbf{S}$. The relation can be captured by the ridge regression problem \cite{hoerl1970ridge} as
\begin{equation}
\begin{aligned}
\label{eq:trainingloss}
\operatornamewithlimits{min}_{\mathbf{W}}~\|\mathbf{D}\mathbf{S}\mathbf{W}-\mathbf{Y}\|^2_\mathrm{F}+\gamma\|\mathbf{W}\|^2_\mathrm{F}
\end{aligned}
\end{equation}
where $\mathbf{D}=[\mathbf{I}^{l\times l},~\mathbf{0}^{l\times(n-l)}] \in \mathbb{R}^{l \times n}$, and $\|\mathbf{W}\|^2_\mathrm{F}$ controls the capacity of $\mathbf{W}$ with the parameter $\gamma$ controlling its influence.

\begin{figure}[t]
\centering
\begin{minipage}[l]{0.6\columnwidth}
  \centering
  \includegraphics[width=1\textwidth]{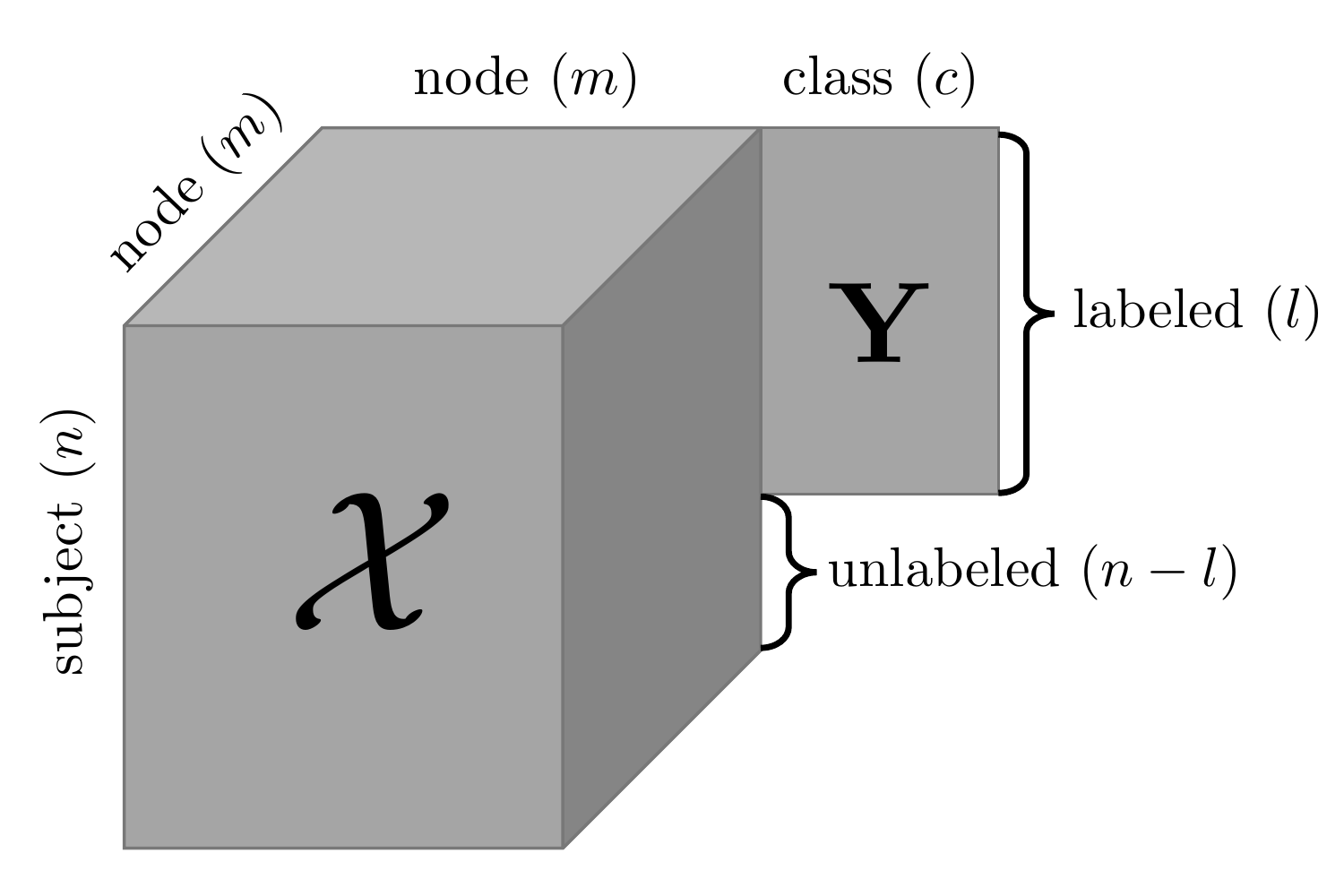}
\end{minipage}
\caption{Partially coupled matrix and tensor.}
\label{fig:coupled}
\end{figure}

An intuitive idea is to first obtain the latent factors of subjects as features and then train a classifier on them in a two-step manner. However, the advantage is established in \cite{bzdok2015semi} of directly searching for classification-relevant structures in the original data, rather than solving the supervised and unsupervised problems independently. To address the problem (P3) discussed in Section~\ref{sec:bne}, we propose to incorporate the classifier learning process (\emph{i.e.}, $\mathbf{W}$) into the framework of learning feature representations of subjects (\emph{i.e.}, $\mathbf{S}$). In this manner, the weight matrix $\mathbf{W}$ and the feature matrix $\mathbf{S}$ can interact with each other in an end-to-end learning framework. Note that it is similar to coupled matrix and tensor factorization \cite{acar2011all}, except that $\mathcal{X}$ and $\mathbf{Y}$ are only coupled along a part of the subject mode, as shown in \ref{fig:coupled}.

In summary, the proposed brain network embedding framework can be mathematically formulated as solving the optimization problem
\begin{equation}
\begin{aligned}
\label{eq:model}
\operatornamewithlimits{min}_{\mathbf{B},\mathbf{S},\mathbf{W}}~
&\underbrace{\|\mathcal{X}-\mathcal{C}\times_1\mathbf{B}\times_2\mathbf{B}\times_3\mathbf{S}\|^2_\mathrm{F}}_\text{factorization error} +\alpha\underbrace{\text{tr}(\mathbf{S}^\top\mathbf{L}_\mathbf{Z}\mathbf{S})}_\text{guidance}
+\beta\underbrace{\|\mathbf{D}\mathbf{S}\mathbf{W}-\mathbf{Y}\|^2_\mathrm{F}}_\text{classification loss}
+\gamma\underbrace{\|\mathbf{W}\|^2_\mathrm{F}}_\text{reg.} \\
\text{s.t.}~&\underbrace{\mathbf{S}^\top\mathbf{S}=\mathbf{I}}_\text{orthogonality}
\end{aligned}
\end{equation}
where $\alpha$, $\beta$, $\gamma$ are all positive parameters that control the contributions of side information guidance, classification loss, and regularization, respectively.

\subsection{Optimization Framework}

The model parameters that have to be estimated include $\mathbf{B}\in\mathbb{R}^{m \times k}$, $\mathbf{S}\in\mathbb{R}^{n \times k}$, and $\mathbf{W}\in\mathbb{R}^{k \times c}$. The optimization problem in \ref{eq:model} is not convex with respect to $\mathbf{B}$, $\mathbf{S}$, and $\mathbf{W}$ together. There is no closed-form solution for the problem. We now introduce an alternating scheme to solve the optimization problem. The key idea is to decouple the orthogonal constraint using an Alternating Direction Method of Multipliers ({\admm}) scheme \cite{boyd2011distributed}. We optimize the objective with respect to one variable, while fixing the others. The algorithm keeps updating the variables until convergence.

\noindent\textbf{Update the node factor matrix}. Firstly, we optimize $\mathbf{B}$ while fixing $\mathbf{S}$ and $\mathbf{W}$. Note that $\mathcal{X}$ is a partially symmetric tensor, and the objective function in \ref{eq:model} involves a fourth-order term with respect to $\mathbf{B}$ which is difficult to optimize directly. To obviate this problem, we use a variable substitution technique and minimize the objective function
\begin{equation}
\begin{aligned}
\label{eq:model_B}
\operatornamewithlimits{min}_{\mathbf{B}, \mathbf{P}}~
&\|\mathcal{X}-\mathcal{C}\times_1\mathbf{B}\times_2\mathbf{P}\times_3\mathbf{S}\|^2_\mathrm{F} \\
\text{s.t.}~&\mathbf{P}=\mathbf{B}
\end{aligned}
\end{equation}
where $\mathbf{P}$ is an auxiliary matrix. The augmented Lagrangian function is
\begin{equation}
\begin{aligned}
\label{eq:admm_B}
\mathcal{L}(\mathbf{B}, \mathbf{P})=
\|\mathcal{X}-\mathcal{C}\times_1\mathbf{B}\times_2\mathbf{P}\times_3\mathbf{S}\|^2_\mathrm{F}
+\text{tr}(\mathbf{U}^\top(\mathbf{P}-\mathbf{B}))
+\frac{\mu}{2}\|\mathbf{P}-\mathbf{B}\|^2_\mathrm{F}
\end{aligned}
\end{equation}
where $\mathbf{U}\in\mathbb{R}^{m \times k}$ is the Lagrange multiplier, and $\mu$ is the penalty parameter which can be adjusted efficiently according to \cite{lin2011linearized}. The optimization problem can be rewritten as
\begin{equation}
\begin{aligned}
\label{eq:optimize_B}
\operatornamewithlimits{min}_{\mathbf{B}}~
\|\mathbf{B}\mathbf{E}^\top-\mathbf{X}_{(1)}\|^2_\mathrm{F}
+\frac{\mu}{2}\|\mathbf{B}-\mathbf{P}-\frac{1}{\mu}\mathbf{U}\|^2_\mathrm{F}
\end{aligned}
\end{equation}
where $\mathbf{E}=\mathbf{S}\odot\mathbf{P}\in\mathbb{R}^{(m*n) \times k}$ ($\odot$ is Khatri-Rao product), and $\mathbf{X}_{(1)}\in\mathbb{R}^{m \times (m*n)}$ is the mode-1 matricization of $\mathcal{X}$. By setting the derivative of \ref{eq:optimize_B} with respect to $\mathbf{B}$ to zero, we obtain the closed-form solution
\begin{equation}
\begin{aligned}
\label{eq:update_B}
\mathbf{B}=(2\mathbf{X}_{(1)}\mathbf{E}+\mu\mathbf{P}+\mathbf{U})(2\mathbf{E}^\top\mathbf{E}+\mu\mathbf{I})^{-1}
\end{aligned}
\end{equation}

To efficiently compute $\mathbf{E}^\top\mathbf{E}$, we can leverage the property of the Khatri-Rao product of two matrices ($*$ is Hadamard product) \cite{kolda2009tensor}
\begin{equation}
\begin{aligned}
\label{eq:ete}
\mathbf{E}^\top\mathbf{E}=(\mathbf{S}\odot\mathbf{P})^\top(\mathbf{S}\odot\mathbf{P})=\mathbf{S}^\top\mathbf{S}*\mathbf{P}^\top\mathbf{P}
\end{aligned}
\end{equation}

The auxiliary matrix $\mathbf{P}$ can be optimized in a similar manner
\begin{equation}
\begin{aligned}
\label{eq:update_P}
\mathbf{P}=(2\mathbf{X}_{(2)}\mathbf{F}+\mu\mathbf{B}-\mathbf{U})(2\mathbf{F}^\top\mathbf{F}+\mu\mathbf{I})^{-1}
\end{aligned}
\end{equation}
where $\mathbf{F}=\mathbf{S}\odot\mathbf{B}\in\mathbb{R}^{(m*n) \times k}$, and $\mathbf{X}_{(2)} \in \mathbb{R}^{m \times (m*n)}$ is the mode-2 matricization of $\mathcal{X}$. Moreover, we optimize the Lagrange multiplier $\mathbf{U}$ using gradient ascent
\begin{equation}
\begin{aligned}
\label{eq:update_U}
\mathbf{U}\leftarrow\mathbf{U}+\mu(\mathbf{P}-\mathbf{B})
\end{aligned}
\end{equation}

\noindent\textbf{Update the subject factor matrix}. Next, we optimize $\mathbf{S}$ while fixing $\mathbf{B}$ and $\mathbf{W}$. We need to minimize the objective function
\begin{equation}
\begin{aligned}
\label{eq:optimize_S}
\mathcal{L}(\mathbf{S})=
&\|\mathbf{S}\mathbf{G}^\top-\mathbf{X}_{(3)}\|^2_\mathrm{F}
+\alpha\text{tr}(\mathbf{S}^\top\mathbf{L}_\mathbf{Z} \mathbf{S})
+\beta\|\mathbf{D}\mathbf{S}\mathbf{W}-\mathbf{Y}\|^2_\mathrm{F} \\
\text{s.t.}~&\mathbf{S}^\top\mathbf{S}=\mathbf{I}
\end{aligned}
\end{equation}
where $\mathbf{G}=\mathbf{P}\odot\mathbf{B}\in\mathbb{R}^{(m*m) \times k}$, and $\mathbf{X}_{(3)} \in \mathbb{R}^{n \times (m*m)}$ is the mode-3 matricization of $\mathcal{X}$. Such an optimization problem has been well studied and can be solved by many existing orthogonality preserving methods \cite{absil2009optimization,helmke2012optimization,wen2013feasible}. Here we employ the curvilinear search approach introduced in \cite{wen2013feasible}, for which we calculate the derivative of $\mathcal{L}(\mathbf{S})$ with respect to $\mathbf{S}$ as
\begin{equation}
\begin{aligned}
\label{eq:update_S}
\nabla_{\mathbf{S}} \mathcal{L}(\mathbf{S})= 
\mathbf{S}\mathbf{G}^\top\mathbf{G}-\mathbf{X}_{(3)}\mathbf{G}
+\alpha\mathbf{L}_\mathbf{Z} \mathbf{S}
+\beta\mathbf{D}^\top(\mathbf{D}\mathbf{S}\mathbf{W}-\mathbf{Y})\mathbf{W}^\top
\end{aligned}
\end{equation}

\noindent\textbf{Update the weight matrix}. Last, we optimize $\mathbf{W}$ while fixing $\mathbf{B}$ and $\mathbf{S}$. We need to minimize the objective function
\begin{equation}
\begin{aligned}
\label{eq:optimize_W}
\mathcal{L}(\mathbf{W}) = \|\mathbf{D}\mathbf{S}\mathbf{W}-\mathbf{Y}\|^2_\mathrm{F} + \gamma\|\mathbf{W}\|^2_\mathrm{F}
\end{aligned}
\end{equation}

Note that \ref{eq:optimize_W} is a regularized least squares problem with the closed-form solution
\begin{equation}
\begin{aligned}
\label{eq:update_W}
\mathbf{W} = (\mathbf{S}^\top\mathbf{D}^\top\mathbf{D}\mathbf{S}+\gamma\mathbf{I})^{-1}
\mathbf{S}^\top\mathbf{D}^\top\mathbf{Y}
\end{aligned}
\end{equation}

Based on the analysis above, we outline the optimization framework for brain network embedding in Algorithm~\ref{algo:admm}.
The implementation has been made available at {\github}\footnote{\url{https://github.com/caobokai/tBNE}}.

\begin{algorithm}[t]
\caption{\bne}
\label{algo:admm}
\begin{algorithmic}[1]
\REQUIRE $\mathcal{X}$ (stacked brain networks), $\mathbf{Z}$ (auxiliary features), $\mathbf{Y}$ (class labels), $k$ (rank of tensor factorization), $\alpha$ (weight of side information), $\beta$ (weight of classification loss), $\gamma$ (weight of regularization)
\ENSURE $\mathbf{B}$ (node factors), $\mathbf{S}$ (subject factors), $\mathbf{W}$ (classification weights)
\STATE $\mu_{max}=10^6,~\rho=1.15$
\STATE Initialize $\mathbf{B}, \mathbf{S}, \mathbf{W}\sim\mathcal{N}(0,1),~\mathbf{U}=\mathbf{0},~ \mu=10^{-6}$
\REPEAT
    \STATE Update $\mathbf{B}$ and $\mathbf{P}$ by \ref{eq:update_B} and \ref{eq:update_P}
    \STATE Update $\mathbf{U}$ by \ref{eq:update_U}
    \STATE Update $\mu$ by $\mu\leftarrow\text{min}(\rho\mu,\mu_{max})$
    \STATE Update $\mathbf{S}$ by \ref{eq:update_S} with the curvilinear search
    \STATE Update $\mathbf{W}$ by \ref{eq:update_W}
\UNTIL{convergence}
\end{algorithmic}
\end{algorithm}

\subsection{Time Complexity}

Each {\admm} iteration consists of simple matrix operations. Therefore, rough estimates of its computational complexity can be easily derived \cite{liavas2015parallel}.

The estimate for the update of $\mathbf{B}$ according to \ref{eq:update_B} is as follows:
$O(m^2nk)$ for the computation of the term $2\mathbf{X}_{(1)}\mathbf{E}+\mu\mathbf{P}+\mathbf{U}$;
$O((m+n)k^2)$ for the computation of the term $2\mathbf{E}^\top\mathbf{E}+\mu\mathbf{I}$ due to \ref{eq:ete} and $O(k^3)$ for its Cholesky decomposition;
$O(mk^2)$ for the computation of the system solution that gives the updated value of $\mathbf{B}$.
An analogous estimate can be derived for the update of $\mathbf{P}$.

Considering $l<n$ and $c$ is usually a small constant, the estimate for the update of $\mathbf{W}$ according to \ref{eq:update_W} is as follows:
$O(n^2k+nk^2)$ for the computation of the term $\mathbf{S}^\top\mathbf{D}^\top\mathbf{D}\mathbf{S}+\gamma\mathbf{I}$ and $O(k^3)$ for its Cholesky decomposition;
$O(nk)$ for the computation of the term $\mathbf{S}^\top\mathbf{D}^\top\mathbf{Y}$;
$O(k^2)$ for the computation of the system solution that gives the updated value of $\mathbf{W}$.

Overall, the update of model parameters $\mathbf{B}$, $\mathbf{P}$ and $\mathbf{W}$ requires $O(k^3+(m+n)k^2+(m^2n+n^2)k)$ arithmetic operations in total. Note that it excludes the update of $\mathbf{S}$ which depends on the orthogonality preserving method we use.

\subsection{Discussion}

The proposed method {\bne} is a general framework of modeling brain networks, and we introduce its potential extensions and related variations.

\noindent\textbf{Guidance}. The side information guidance used in {\bne} essentially regularizes the subject factor matrix in a row-wise manner. Another approach to incorporating the side information is to perform coupled matrix and tensor factorization \cite{acar2011all}. However, it would introduce additional model parameters, \emph{i.e.}, a factor matrix for auxiliary features. On the other hand, a column-wise guidance information can be added on factors if we have prior knowledge about community information of subjects or brain regions \cite{wang2015rubik}. Alternatively, such knowledge can also be modeled as an augmented space \cite{lian2014geomf}.

\noindent\textbf{Supervision}. Rather than integrating the process of training a classifier with tensor factorization, we can borrow the idea from \cite{kong2010semi} that latent factors should satisfy the must-link, cannot-link, and separability constraints. Intuitively, these constraints tend to discover latent factors that can distinguish different class labels among labeled subjects and separate unlabeled subjects from their distribution. 

\noindent\textbf{Multimodality}. A tensor can be constructed from brain networks of each modality, \emph{e.g.}, {\eeg}, {\fmri}, and {\dti}. We can perform joint tensor factorization to capture the consensus information across multiple modalities \cite{liu2018multi}.

\section{Experiments}

In this section, we first describe the {\eeg} brain network datasets and the compared methods. Next, we analyze the experimental results and investigate the parameter sensitivity. Last, we present the visualization of the obtained representations.

\subsection{Data Collection}

\begin{figure}[t]
\centering
\begin{minipage}[l]{0.6\columnwidth}
  \centering
  \includegraphics[width=1\textwidth]{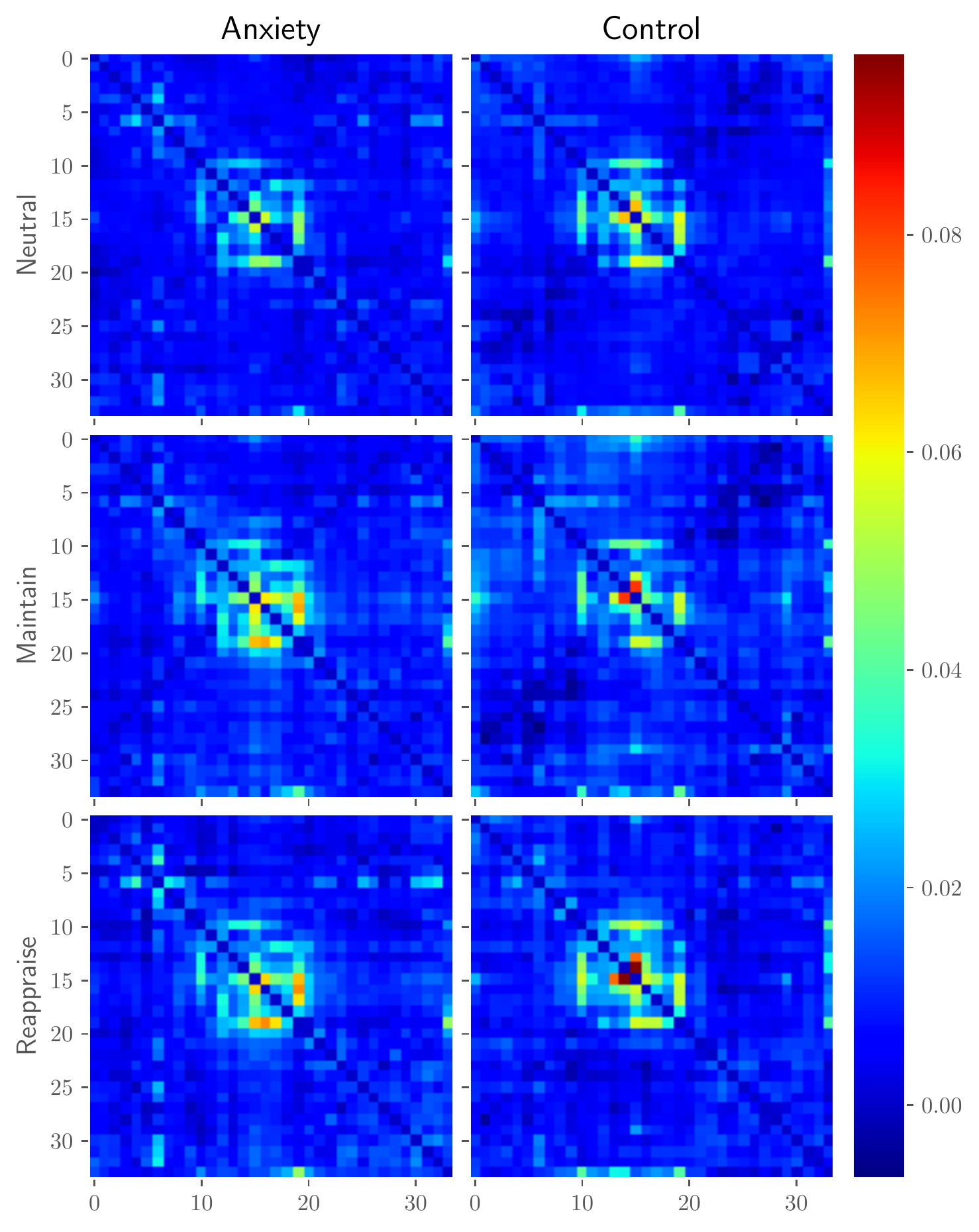}
\end{minipage}
\caption{Average brain networks on different diagnosises and tasks.}
\label{fig:heatmap}
\end{figure}

Data were collected from 37 patients with primary diagnoses of social anxiety disorder (\texttt{SAD}) or generalized anxiety disorder (\texttt{GAD}) and 32 healthy controls. Each subject underwent an emotion regulation task related to (1) viewing the neutral pictures, (2) viewing the negative pictures as they normally would, and (3) viewing the negative pictures while attempting to reduce their emotion response by reinterpreting the meaning of pictures. All {\eeg} data were recorded using the Biosemi system equipped with an elastic cap with 34 scalp channels. A detailed description about data acquisition and preprocessing is available in \cite{xing2016eeg}.

We partition the data into three datasets based on different tasks, denoted as {\nt}, {\mt} and {\rp}, respectively. Hence, in each dataset, there are $n=69$ subjects with their corresponding task-specific {\eeg} brain networks which contain $m=34$ nodes. The target is to distinguish patients from healthy controls under different emotion regulation tasks. The average brain networks are shown in \ref{fig:heatmap}, where the horizontal and vertical axes represent the node index, and the color of the cell represents the strength of the connection between two nodes. We can see that the connections between nodes from 11 to 20 are generally stronger than other node pairs.

In addition, self-report scales including Beck Depression Inventory, State-Trait Anxiety Inventory, and Leibowitz Social Anxiety Scale were obtained from all the subjects. Each scale is represented as a range of scores corresponding to the symptom severity rating of depression and anxiety.
In general, subjects with a higher score suggest a greater level of disease burden.
We use these scales as side information to guide the tensor factorization procedure.

\subsection{Compared Methods}

The compared methods are summarized as follows:
\begin{itemize}[leftmargin=*,noitemsep,topsep=0pt]
\setlist{nosep}
\item\textbf{\bne}: the proposed tensor factorization model for brain network embedding.
\item\textbf{\cmtf}: coupled matrix and tensor factorization where brain networks and side information are coupled along the subject mode \cite{acar2011all}.
\item\textbf{\rubik}: tensor factorization with orthogonality and sparsity constraints \cite{wang2015rubik}.
\item\textbf{\als}: tensor factorization using alternating least squares without any constraint \cite{comon2009tensor}.
\item\textbf{\gmsv}: a discriminative subgraph selection approach using side information \cite{cao2015mining}.
\item\textbf{\cc}: local clustering coefficients, one of the most popular graph-theoretical measures that quantify the cliquishness of nodes \cite{rubinov2010complex}.
\end{itemize}

For a fair comparison, we use ridge regression \cite{hoerl1970ridge} as in {\bne} as the base classifier for all the compared methods. All factorization based methods use the same stopping threshold, \emph{i.e.}, the difference between the explained variations of two consecutive estimations with the threshold value of $10^{-4}$. Moreover, to assure that all the compared methods have access to all data sources, especially to the side information, we conduct feature selection using Laplacian Score \cite{he2005laplacian} based on the side information for {\cc}. The number of selected features is determined by a hyperparameter $k$ which is equal to the number of latent factors in factorization models. In this manner, the number of (latent) features used for classification is the same for all the compared methods. In summary, $k$ is tuned in the same range of $1,\cdots,20$, the regularization parameter $\gamma$ is tuned in the same range of $2^{-6},\cdots,2^{6}$ for all the compared methods, and other model-specific parameters are set as default, \emph{e.g.}, $\alpha=\beta=0.1$ in {\bne} and $\lambda_q=0.1$ in {\rubik}. In the experiments, 10-fold cross validation is performed, and the average accuracy with the best parameter configuration is reported.

\subsection{Classification Performance}

\begin{table}[t]
\centering
\caption{Graph classification accuracy.}
\label{tab:bne_mainresult}
\newcolumntype{x}[1]{>{\centering\arraybackslash}p{#1}}
\begin{tabular}{||l|x{2cm}|x{2cm}|x{2cm}||}
\hline
\multirow{2}*{Methods} & \multicolumn{3}{c||}{Datasets}\\
\cline{2-4}
&{\nt} &{\mt} &{\rp}\\									\hline\hline
{\bne}      &0.7833 &0.7548 &0.7524 \\
{\cmtf}     &0.5810 &0.7095 &0.6381 \\
{\rubik}    &0.6405 &0.6833 &0.6667 \\
{\als}      &0.6119 &0.6667 &0.6524 \\
{\gmsv}     &0.6500 &0.6548 &0.5952 \\
{\cc}       &0.5357 &0.6667 &0.5357 \\
\hline
\end{tabular}
\end{table}

Experimental results in \ref{tab:bne_mainresult} show the classification performance of the compared methods under three different emotion regulation tasks.
A significant improvement of $20.51\%$, $6.38\%$ and $12.85\%$ by {\bne} over the best baseline performance can be observed on {\nt}, {\mt} and {\rp} datasets, respectively.
Although clustering coefficients have been widely used to identify Alzheimer's disease \cite{wee2012identification,jie2014integration}, they appear to be less useful for distinguishing anxiety patients from normal controls. Note that they are filtered using Laplacian Score \cite{he2005laplacian} based on the side information. On the other hand, {\gmsv} achieves a better performance than {\cc} on {\nt} and {\rp} datasets by extracting connectivity patterns within brain networks that are consistent with the side information guidance.

In general, factorization based models demonstrate themselves with better accuracies. Based on the low-rank assumption, a low-dimensional latent factor of each subject is obtained by factorizing the stacked brain network data of all the subjects.
{\als} is a direct application of the alternating least squares technique to the tensor factorization problem without incorporating any domain knowledge.
{\rubik} is a constrained tensor factorization method by regularizing the subject factors to be orthogonal and enforcing sparsity.
{\cmtf} incorporates the side information by collectively factorizing the brain network tensor and the side information matrix with shared subject factors.
{\bne} achieves the best performance through the employment of factorizing a partially symmetric tensor, introducing the side information guidance and fusing the processes of tensor factorization and classifier learning.

\begin{figure}[t]
\centering
\begin{minipage}[l]{0.8\columnwidth}
  \centering
  \includegraphics[width=1\textwidth]{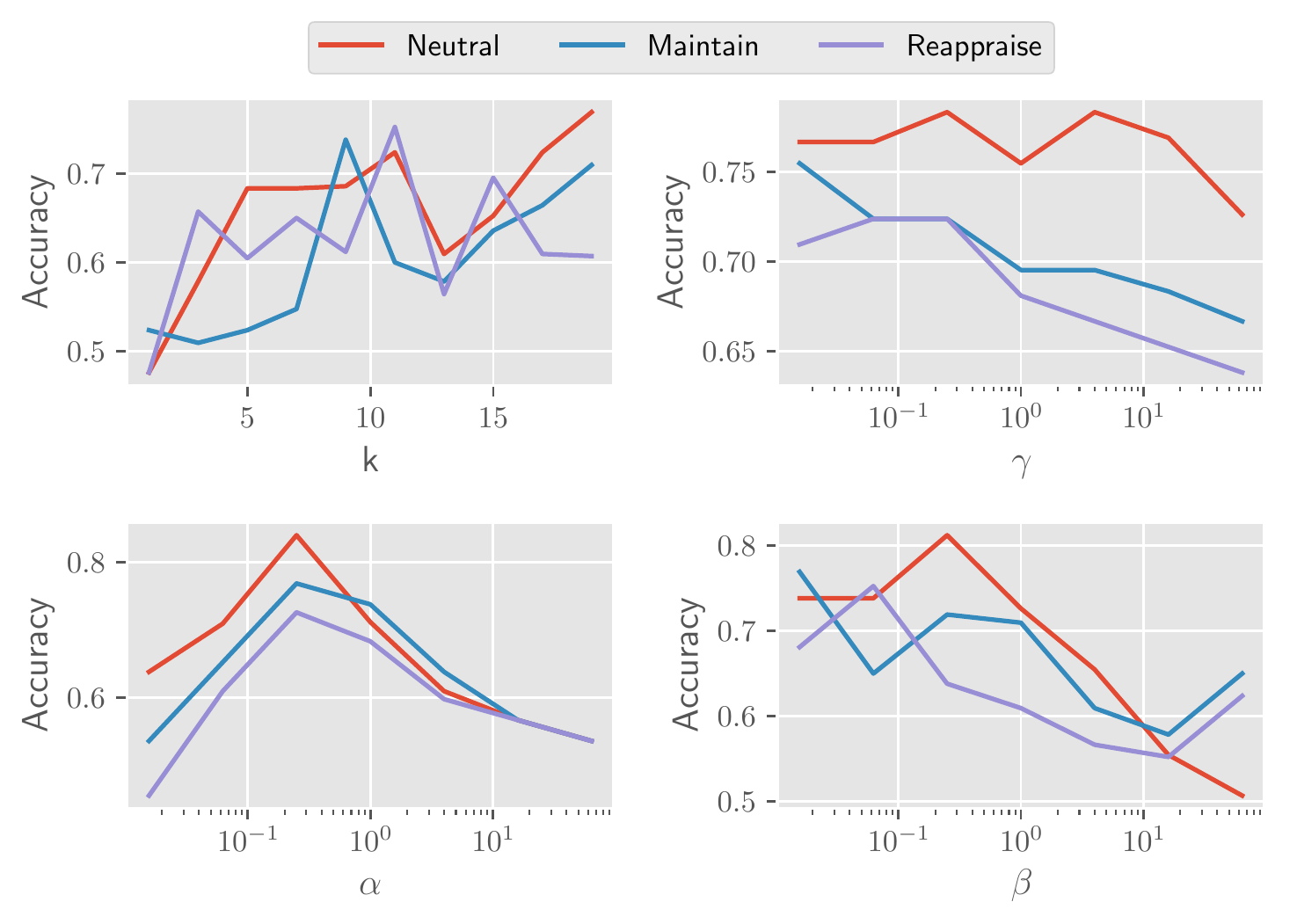}
\end{minipage}
\caption{Sensitivity analysis of hyperparameters in {\bne}.}
\label{fig:sensitivity}
\end{figure}

\subsection{Parameter Sensitivity}

In the experiments above, the degree of freedom of all the compared methods is 2, because of the hyperparameters $k$ and $\gamma$, and other model-specific parameters are fixed. In order to evaluate how changes to the parameterization of {\bne} affect its performance on classification tasks, we study the influence of the hyperparameters $k$ and $\gamma$, as well as $\alpha$ and $\beta$ in the proposed {\bne} model.

\begin{figure}[t]
\centering
\subfigure[{\nt}.]{
\begin{minipage}[l]{0.6\columnwidth}
  \centering
  \includegraphics[width=1\textwidth]{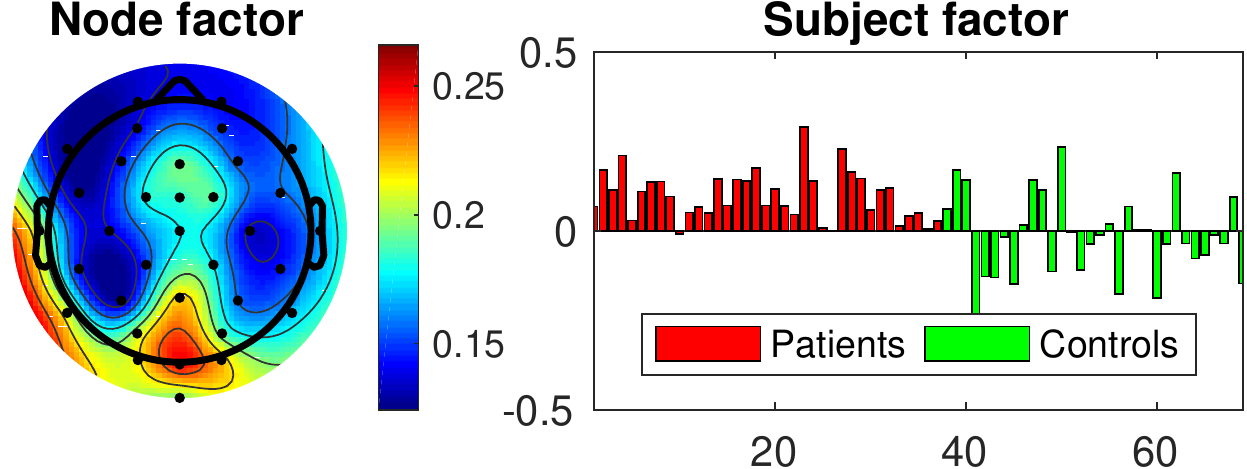}
\end{minipage}
}
\subfigure[{\mt}.]{
\begin{minipage}[l]{0.6\columnwidth}
  \centering
  \includegraphics[width=1\textwidth]{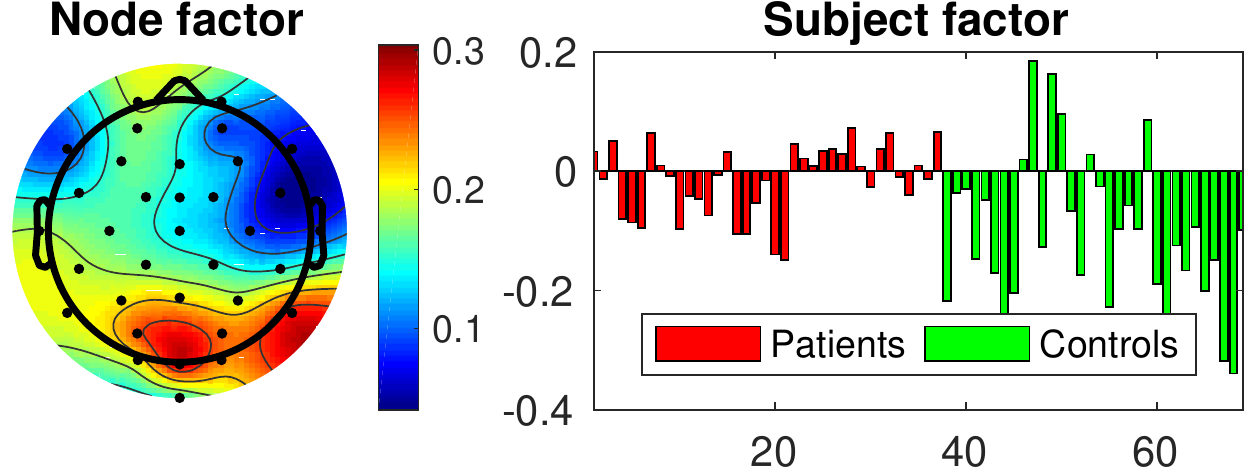}
\end{minipage}
}
\subfigure[{\rp}.]{
\begin{minipage}[l]{0.6\columnwidth}
  \centering
  \includegraphics[width=1\textwidth]{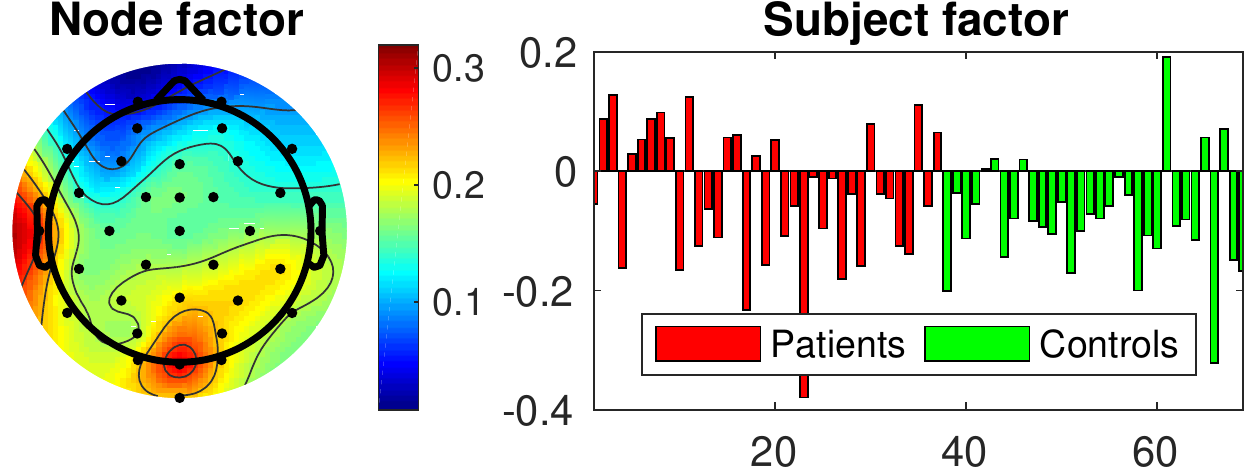}
\end{minipage}
}
\caption{Anxiety-related latent factors on different cognitive tasks.}
\label{fig:factor}
\end{figure}

In \ref{fig:sensitivity}, the performance of {\bne} with different $k$ shows that a very small $k$ would not be a wise choice in general, and the best performance can usually be achieved around $k=10$. For the regularization parameter $\gamma$, we observe that {\bne} is insensitive to $\gamma$ in a relatively large range, and a small $\gamma$ would be preferred (\emph{e.g.}, $\gamma\le2^{-2}$). It makes sense because a large $\gamma$ will let the regularization term override the effect of other terms and thus dominate the objective. Considering the influence of $\alpha$ in \ref{fig:sensitivity}, the performance of {\bne} in \ref{tab:bne_mainresult} can be further improved when $\alpha=2^{-2}$. Neither a small nor a large $\alpha$ would be preferred, because both of the brain network data and the side information guidance are important to learn discriminative representations of brain networks. It is critical to set a desirable $\alpha$ for the trade-off between the two different data sources, in order to acquire the complementary information from them. The sensitivity analysis of {\bne} with respect to $\beta$ generally shows that a good choice of $\beta$ can be found around $2^{-2}$.

\subsection{Factor Analysis}

A $k$-factor {\bne} model extracts $\mathbf{B}(:,i)$ and $\mathbf{S}(:,i)$, for $i=1,\cdots,k$, where these factors indicate the signatures of sources in brain regions and the subject domain, respectively. In \ref{fig:factor}, we show the largest factors in terms of magnitude that are learned from {\nt}, {\mt} and {\rp} datasets, respectively. In the left panel, dots indicate the spatial layout of electrodes (\emph{i.e.}, nodes) on the scalp, and factor values of electrodes are demonstrated on a colormap using {\eeglab} \cite{delorme2004eeglab}. The right panel shows the factor strengths for each subject (both patients and controls). A domain expert may examine a brain activity pattern in the left panel and the difference between subjects in such a pattern in the right panel. For instance, we can observe that regardless of emotion regulation tasks, the first factor always exhibits a pattern of occipital dominance, while {\rp} additionally includes a left temporal/parietal involvement, likely reflecting the integration between fronto-parietal ``cognitive" control networks and temporal limbic regions instrumental for emotion processing.

\section{Related Work}

The recent development of brain network analysis has enabled characterization of brain disorders at a whole-brain connectivity level and provided a new direction for brain disease classification \cite{kong2014brain,cao2015review}. 
Wee et al.~extracted weighted local clustering coefficients from brain networks to quantify the prevalence of clustered connectivity around brain regions for disease diagnosis on Alzheimer's disease \cite{wee2012identification}. Jie et al.~used the Weisfeiler-Lehman subtree kernel \cite{shervashidze2011weisfeiler} to measure the topological similarity between paired {\fmri} brain networks \cite{jie2014integration}. Compared with graph kernel approaches, subgraph pattern mining approaches are more interpretable. Kong et al.~proposed a discriminative subgraph feature selection method based on dynamic programming by modeling normalized brain networks as weighted graphs \cite{kong2013discriminative}.

There are also increasing research efforts of incorporating constraints in tensor factorization. Carroll et al.~described a least squares fitting procedure with linear constraints for tensor data \cite{carroll1980candelinc}. Narita et al.~provided a framework to utilize relationships among data as auxiliary information to improve the quality of tensor factorization \cite{narita2012tensor}. Davidson et al.~proposed a constrained alternating least squares framework for network analysis of {\fmri} data \cite{davidson2013network}. 
Wang et al.~introduced knowledge guided tensor factorization for computational phenotyping \cite{wang2015rubik}. However, some of the constraints are specifically designed for one domain, and thus it is not straightforward to have them applied to other areas, \emph{e.g.}, brain network analysis.

Rather than embedding the prior knowledge as guidance or constraints, another approach to fusing heterogeneous information sources is coupled factorization where matrices and tensors sharing some common modes are jointly factorized \cite{ermics2015link}. Acar et al.~proposed a gradient-based optimization approach for joint analysis of matrices and higher order tensors \cite{acar2011all}. Scalable solutions to the coupled factorization problem are presented in \cite{beutel2014flexifact,papalexakis2014turbo}. However, these frameworks are not directly applicable to partially symmetric tensor factorization, and they do not leverage any domain knowledge for brain network embedding.

\chapter{Multi-view sequence prediction}
\label{chapter:deepmood}

(This chapter was previously published as ``DeepMood: Modeling Mobile Phone Typing Dynamics for Mood Detection \cite{cao2017deepmood}'', in \textit{Proceedings of the 23rd ACM SIGKDD International Conference on Knowledge Discovery and Data Mining (KDD)}, 2017, ACM. DOI: \url{https://doi.org/10.1145/3097983.3098086}.)

\section{Introduction}

Mobile phones, in particular ``smartphones'', have become near ubiquitous with 2 billion smartphone users worldwide. This presents new opportunities in the treatment of psychiatric illness by allowing us to study the manifestations of psychiatric illness in an unobtrusive manner and at a level of detail that was not previously possible. Continuous collection of automatically generated smartphone data that reflect illness activity could facilitate real-time monitoring and early intervention \cite{ankers2009objective,bopp2010longitudinal,faurholt2016behavioral}.

While mobile phones are used for a variety of tasks, the most widely and frequently used feature is text messaging. To the best of our knowledge, no previous studies \cite{american2013diagnostic,puiatti2011smartphone,frost2013supporting,gruenerbl2014using,schleusing2011monitoring,valenza2014wearable} have investigated the relationship between mobile phone typing dynamics and mood states. In this work, we aim to determine the feasibility of inferring mood disturbance and severity from such data. In particular, we seek to investigate the relationship between digital footprints and mood in bipolar affective disorder which is considered to be the most expensive behavioral healthcare diagnosis \cite{peele2003insurance}, costing more than twice as much as depression per affected individual \cite{laxman2008impact}. The expense on inpatient care is nearly twice as much as that on outpatient care for people with bipolar disorder, which suggests that the financial impact of this illness could be relieved by early intervention \cite{peele2003insurance}.

\begin{figure}[t]
\centering
\begin{minipage}[l]{\columnwidth}
  \centering
  \includegraphics[width=1\textwidth]{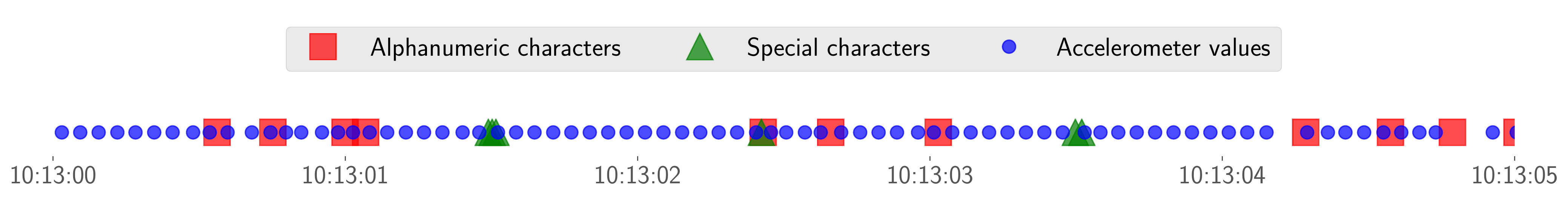}
\end{minipage}
\caption{Example of time series data collected on smartphones.}
\label{fig:timeline}
\end{figure}

We study the mobile phone typing dynamics metadata on a session-level. A session is defined as beginning with a keypress which occurs after 5 or more seconds have elapsed since the last keypress and continuing until 5 or more seconds elapse between keypresses\footnote{5-second is an arbitrary threshold which can be changed and tuned easily.}. The duration of a session is typically less than one minute. In this manner, each user contributes many samples, one per phone usage session, which could benefit data analysis and model training. Each session is composed of sequential features that are represented in multiple views or modalities (\emph{e.g.}, alphanumeric characters, special characters, and accelerometer values), each of which has different timestamps and densities, as shown in \ref{fig:timeline}.
Modeling the multi-view time series data on such a fine-grained session-level brings up several formidable challenges:
\begin{itemize}[leftmargin=*,noitemsep,topsep=0pt]
\item\textbf{Unaligned views}: An intuitive idea for fusing multi-view time series is to align them with each unique timestamp. However, features defined in one view would be missing for data points collected in another view. For example, a data point in the view of alphanumeric characters should have no features that are defined for special characters.
\item\textbf{Dominant views}: One may also attempt to perform the fusion by concatenating the multi-view time series per session. However, the length of a view or the density of timestamps in a session may vary a lot across different views, because the metadata are collected from different sources or sensors. For example, accelerometer values collected in the background have 16 times more data points than character-related metadata that follow a person's typing behaviours in our data collection. Dense views could dominate a concatenated feature space and potentially override the effects of sparse but important views.
\item\textbf{View interactions}: The multi-view time series from typing dynamics contain complementary information reflecting a person's mental health. The relationship between digital footprints and mood states can be highly nonlinear. An effective fusion strategy is needed to explore feature interactions across multiple views.
\end{itemize}

\begin{figure}[t]
\centering
\begin{minipage}[l]{\columnwidth}
  \centering
  \includegraphics[width=1\textwidth]{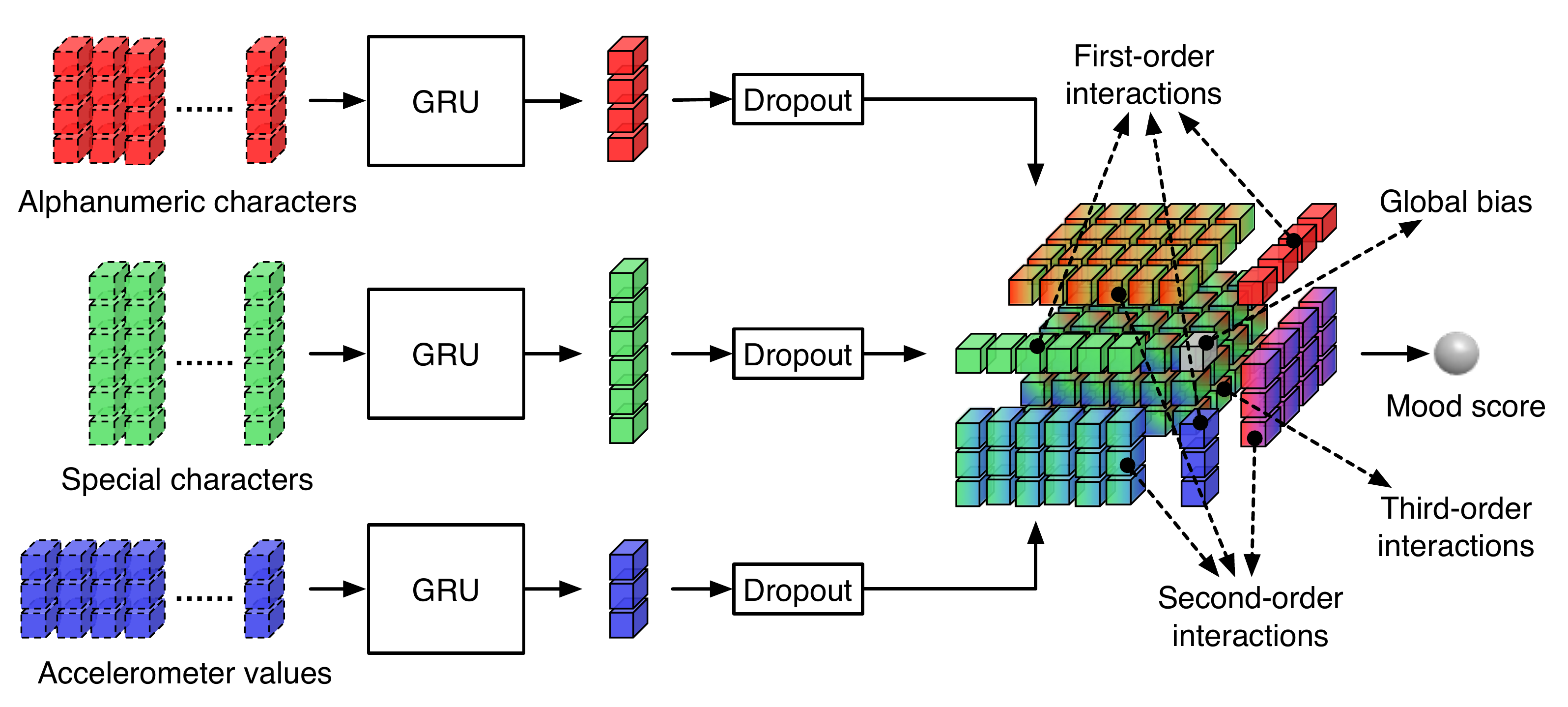}
\end{minipage}
\caption{{\deepmood} architecture with an {\mvm} layer.}
\label{fig:architecture}
\end{figure}

In this work, we first obtain interesting insights related to the digital footprints on mobile phones by analyzing the relationship between the typing dynamics patterns and mood in bipolar affective disorder.
Because of the aforementioned problems brought by early fusion strategies (\emph{i.e.}, aligning views with timestamps or concatenating views per session), we develop an end-to-end deep architecture, named {\deepmood}, to model mobile phone typing dynamics, as illustrated in \ref{fig:architecture}. Specifically, {\deepmood} is a two-stage approach to modeling multi-view time series data based on late fusion. In the first stage, each view of the time series is separately modeled by a Recurrent Neural Network ({\rnn}) \cite{mikolov2010recurrent,sutskever2011generating}. The multi-view representations are then fused in the second stage by exploring interactions across the output vectors from each view, where three different fusion layers are developed. We conduct experiments showing that 90.31\% prediction accuracy on the depression score can be achieved based on session-level typing dynamics, which reveals the potential of using mobile phone metadata to predict mood disturbance and severity.

\section{Problem Formulation}

A labeled instance in the form of multi-view time series is denoted as $\{\mathcal{X}^{(1)}, \cdots, \mathcal{X}^{(m)},\mathbf{y}\}$, where $\mathbf{y}\in\mathbb{R}^c$ is the label, $c$ is the number of classes, and $m$ is the number of views. The $p$-th view is denoted as $\mathcal{X}^{(p)}=\{\mathbf{x}^{(p)}_1,\cdots,\mathbf{x}^{(p)}_{l_p}\}$, where $l_p$ is the sequence length of the $p$-th view which are usually different across instances. For the same instance, $l_p$ and $l_q$ of two different views may also be very different, which leads to the problem of \emph{dominant views}: one view can dominate another after concatenation. The features of the $k$-th data points (in a chronological order) in the $p$-th view are denoted as $\mathbf{x}^{(p)}_k\in\mathbb{R}^{d_p}$, where $d_p$ is the feature dimension of the $p$-th view which is the same for all instances. However, the timestamp associated with the $k$-th data points $\mathbf{x}^{(p)}_k$ and $\mathbf{x}^{(q)}_k$ in two views may not be exactly the same, which leads to the problem of \emph{unaligned views}: features defined in one view would be missing for data points collected in another view after alignment with each unique timestamp. Note that we omit the index for instances here for simplicity. The problem of multi-view sequence prediction is to learn a model from labeled multi-view time series instances and to predict the label $\mathbf{y}^*$ of an unseen instance $\{ \mathcal{X}^{(1)*}, \cdots, \mathcal{X}^{(m)*} \}$.

\section{Data Analysis}
\label{sec:deepmood_data}

The data used in this work were collected from the {\biaffect}\footnote{\url{http://www.biaffect.com}} study which is the winner of the Mood Challenge for ResearchKit\footnote{\url{http://www.moodchallenge.com}}. During a preliminary data collection phase, for a period of 8 weeks, 40 users were provided a Galaxy Note 4 mobile phone which they were instructed to use as their primary phone during the study. This phone was loaded with a custom keyboard that replaced the standard Android OS keyboard. The keyboard collected metadata consisting of keypress entry time and accelerometer movement, and uploaded them to the study server. In order to protect users' privacy, individual character data with the exceptions of the backspace key and space bar were not collected.

In this work, we study the collected metadata for users including bipolar subjects and normal controls who had provided at least one week of metadata. There are 7 users with \emph{bipolar I} disorder that is characterized by severe mood episodes from mania to depression, 5 users with \emph{bipolar II} disorder which is a milder form of mood episodes of hypomania that alternate with periods of severe depression, and 8 users with no diagnosis per \texttt{DSM-IV TR} criteria \cite{kessler2005lifetime}.

Users were administered the Hamilton Depression Rating Scale ({\hdrs}) \cite{williams1988structured} and Young Mania Rating Scale ({\ymrs}) \cite{young1978rating} once a week which are used as the ground-truth to assess the level of depressive and manic symptoms in bipolar disorder. However, a face-to-face clinical evaluation is required for the use of these rating scales which may also be unreliable due to many methodological issues \cite{demitrack1998problem,psaty2010minimizing,faurholt2016behavioral}. Thus, it motivates us to explore more objective methods with real-time data for assessing affective symptoms.

\begin{figure}[t]
\centering
\begin{minipage}[l]{0.5\columnwidth}
  \centering
  \includegraphics[width=1\textwidth]{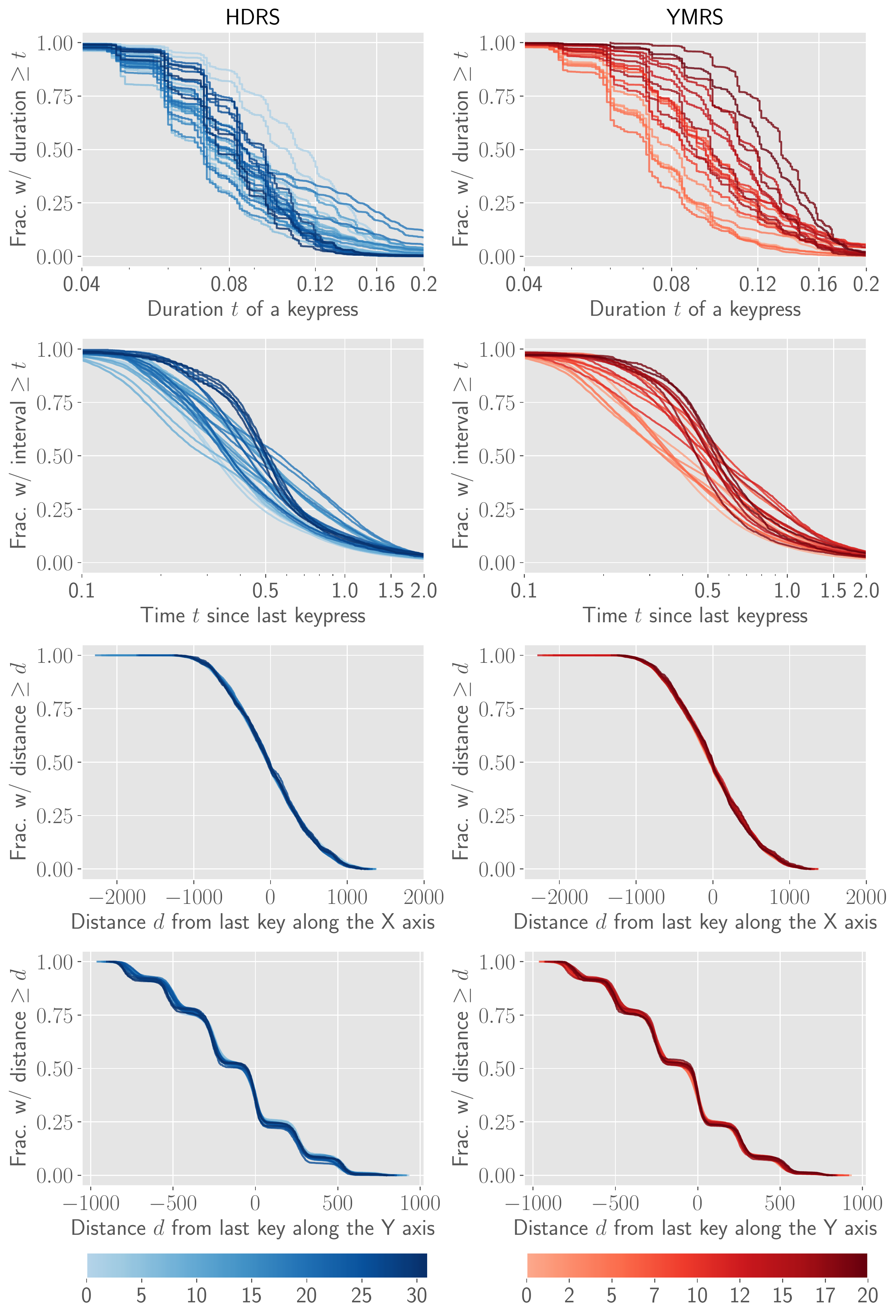}
\end{minipage}
\caption{{\ccdf}s of features with alphanumeric characters.}
\label{fig:cdf_alphanum}
\end{figure}

\noindent\textbf{Alphanumeric characters}.
Due to privacy reasons, we only collected metadata for keypresses on alphanumeric characters, including \emph{duration of a keypress}, \emph{time since last keypress}, and \emph{distance from last key} along two axes. Firstly, we aim to assess the correlation between the duration of a keypress and mood states. The complementary cumulative distribution functions ({\ccdf}s) of the duration of a keypress are displayed at the first row in \ref{fig:cdf_alphanum}. Data points with different mood scores are colored differently, and the range of scores corresponds to the colorbar. In general, the higher the score, the darker the color and the more severe the depressive or manic symptoms. According to the Kolmogorov-Smirnov test on two samples, for all the pairs of distributions, we can reject the null hypothesis that two samples are drawn from the same distribution with significance level $\alpha=0.01$. As expected, we are dealing with a heavy-tailed distribution: (1) most keypresses are very fast with median 85ms, (2) but a non-negligible number have a longer duration with 5\% using more than 155ms. Interestingly, samples with mild depression tend to have a shorter duration than normal ones, while those with severe depression stand in the middle. Samples in manic symptoms seem to hold a key longer than normal ones.

Next, we ask how the time since last keypress correlates with mood states. We show the {\ccdf}s of the time since last keypress at the second row in \ref{fig:cdf_alphanum}. Based on the Kolmogorov-Smirnov test, for 98.06\% in {\hdrs} and 99.52\% in {\ymrs} of the distribution pairs, we can reject the null hypothesis that two samples are drawn from the same distribution with significance level $\alpha=0.01$. Not surprisingly, this distribution is heavily skewed, with most time intervals being very short with median 380ms. However, there is a significant fraction of keypresses with a much longer interval where 5\% have more than 1.422s. We can observe that the values of time since last keypress from the normal group (with light blue/red) approximate a uniform distribution on the log scale in the range from 0.1s to 2.0s. On the contrary, this metric from samples with mood disturbance (with dark blue/red) shows a more skewed distribution with a few values on the two tails and majority centered between 0.4s and 0.8s. In other words, healthy people show a good range of reactivity that gets lost in mood disturbance where the range is more restricted.

\ref{fig:cdf_alphanum} also shows the {\ccdf}s of the distance from last key along two axes which can be considered as a sort of very rough proxy of the semantic content of users' typing.
No distinctions can be observed across different mood states, because there are no dramatic differences in the manner in which depressive or manic users type compared to controls.

\begin{figure}[t]
\centering
\begin{minipage}[l]{0.6\columnwidth}
  \centering
  \includegraphics[width=1\textwidth]{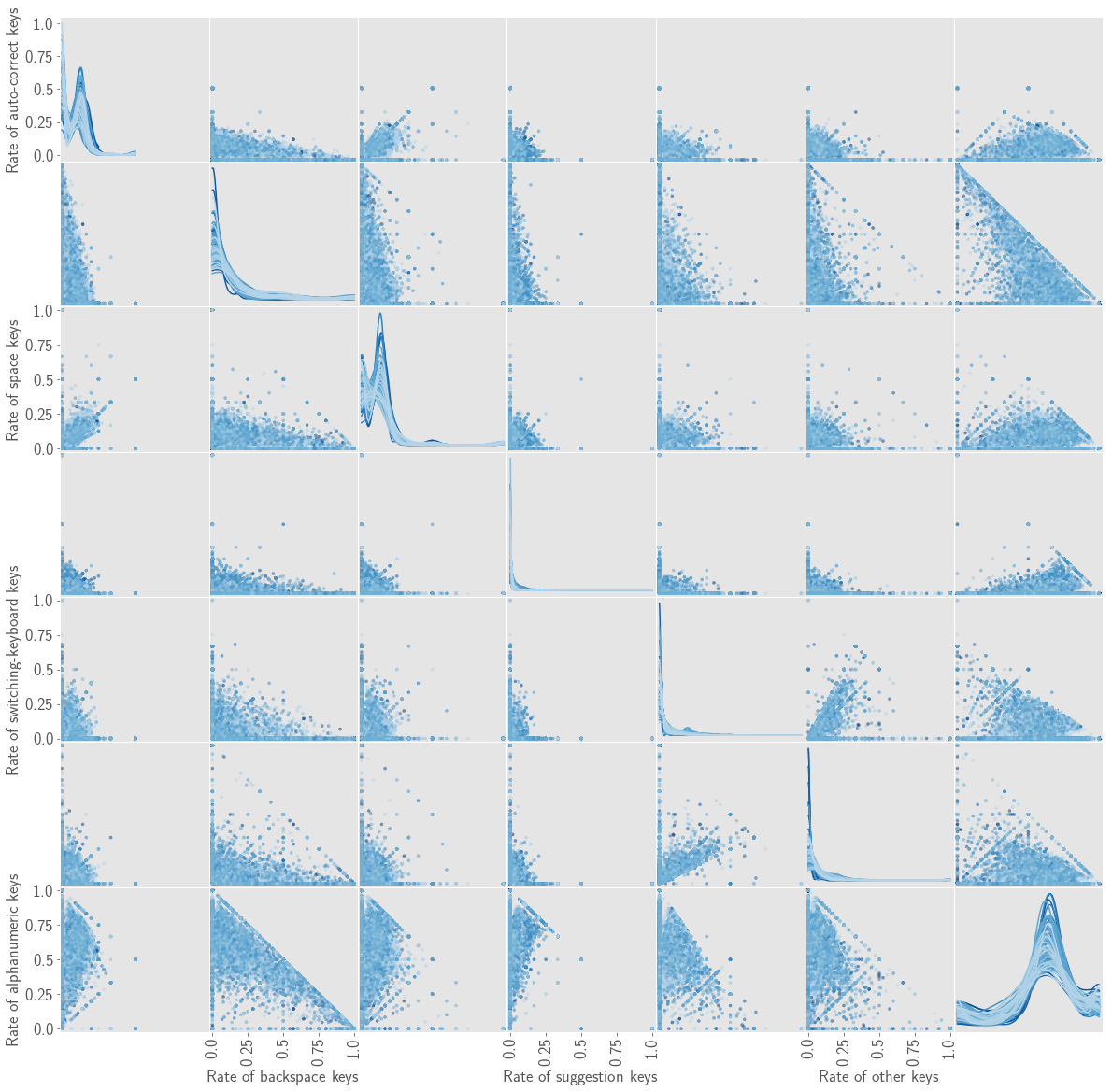}
\end{minipage}
\caption{Scatter plot between rates of different keys per session.}
\label{fig:scatter}
\end{figure}

\noindent\textbf{Special characters}.
In this view, we use one-hot-encoding for typing behaviors other than alphanumeric characters, including \emph{auto-correct}, \emph{backspace}, \emph{space}, \emph{suggestion}, \emph{switching-keyboard} and \emph{other}. They are usually sparser than alphanumeric characters. \ref{fig:scatter} shows a scatter plot between rates of these special characters as well as alphanumeric ones in a session where the color of a dot/line corresponds to the {\hdrs} score. Although no obvious distinctions can be found between mood states, we can observe some interesting patterns: the rate of alphanumeric keys is negatively correlated with the rate of backspace (from the subfigure at the 2nd row, 7th column), while the rate of switching-keyboard is positively correlated with the rate of other keys (from the subfigure at the 5th row, 6th column). On the diagonal there are kernel density estimations, where the rate of alphanumeric characters is generally high in a session, followed by auto-correct, space, backspace, \emph{etc.} 

\begin{figure}[t]
\centering
\begin{minipage}[l]{0.5\columnwidth}
  \centering
  \includegraphics[width=1\textwidth]{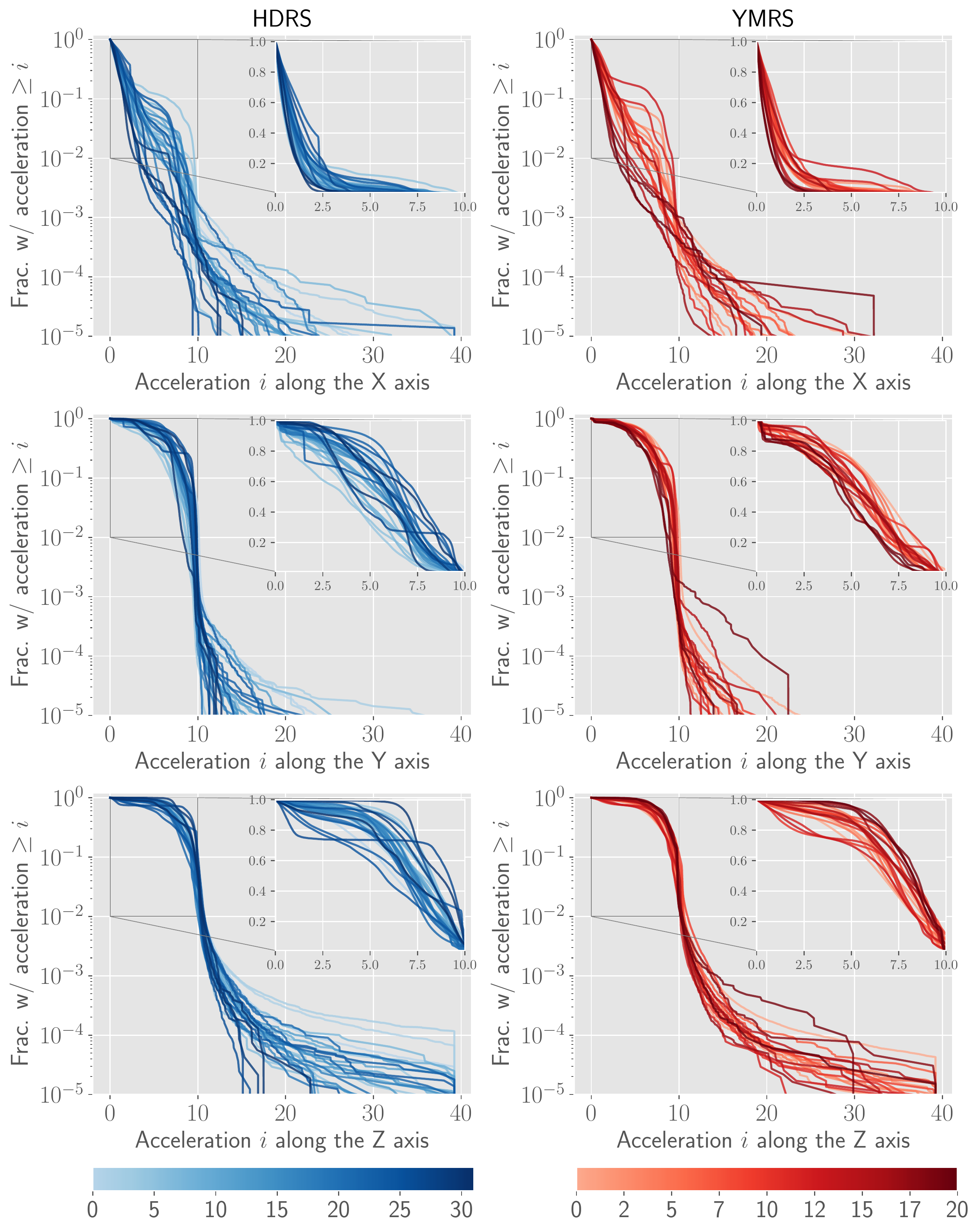}
\end{minipage}
\caption{{\ccdf}s of absolute acceleration along three axes.}
\label{fig:cdf_accel}
\end{figure}

\noindent\textbf{Accelerometer values}.
Accelerometer values are recorded every 60ms in the background during an active session regardless of a user's typing speed, thereby making them much denser than alphanumeric characters. The {\ccdf}s of the absolute accelerometer values along three axes are displayed in \ref{fig:cdf_accel}. Data points with different mood scores are colored differently, and the higher the score, the more severe the depressive or manic symptoms. According to the Kolmogorov-Smirnov test on two samples, for all the pairs of distributions, we can reject the null hypothesis that two samples are drawn from the same distribution with significance level $\alpha=0.01$. Note that the vertical axis of the non-zoomed plots is on a log scale. We observe a heavy-tailed distribution for all three acceleration axes and for both {\hdrs} and {\ymrs}.
By zooming into the ``head'' of the distribution on a regular scale, we can see different patterns on the absolute acceleration along different axes. There is a nearly uniform distribution of absolute acceleration along the Y axis in the range from 0 to 10, while the majority along the X axis lie between 0 and 2, and the majority along the Z axis lie between 6 and 10. An interesting observation is that compared with normal ones, samples with mood disturbance tend to have a larger acceleration along the Z axis and a smaller acceleration along the Y axis. Hence, we suspect that users in a normal mood state prefer to hold their phone towards to themselves, while users in depressive or manic symptoms are more likely to lay their phone with an angle towards to the horizon, given that the data were collected only when the phone was in a portrait position.

See \ref{tab:stats} for more information about the statistics of the dataset. Note that the length of a sequence is measured in terms of the number of data points in a sample rather than the duration in time.

\begin{table}[t]
\centering
\caption{Statistics of the {\biaffect} dataset.}
\label{tab:stats}
\newcolumntype{x}[1]{>{\centering\arraybackslash}p{#1}}
\begin{tabular}{||l|x{2cm}|x{2cm}|x{2cm}||}
\hline
Statistics & {\ch} & {\nonch} & {\accel} \\
\hline\hline
\# data points      & 836,027 & 538,520 & 14,237,503 \\
\# sessions         & 34,993 & 33,385 & 37,647 \\
Mean length         & 24 & 16 & 378 \\
Median length       & 14 & 9 & 259 \\
Maximum length      & 538 & 437 & 90,193 \\
\hline
\end{tabular}
\end{table}

\section{Proposed Method}
\label{sec:method}

In this section, we propose {\deepmood} as an end-to-end deep architecture to model mobile phone typing dynamics. Specifically, {\deepmood} provides a late fusion framework of modeling multi-view time series data. It first models each view of the time series data separately using Gated Recurrent Unit ({\gru}) \cite{cho2014learning} which is a simplified version of Long Short-Term Memory ({\lstm}) \cite{hochreiter1997long}. As the {\gru} extracts a latent feature representation out of each time series, where the notions of timestamp alignment and sequence length are removed from the latent space, this avoids the problem of directly dealing with the heterogeneity of the original multi-view time series data. The intermediate representations produced from multiple {\gru}s are then integrated by three alternative fusion layers, following the idea of Multi-View Machine ({\mvm}) \cite{cao2016multi}, Factorization Machine ({\fm}) \cite{rendle2012factorization}, or in a conventional Fully Connected ({\fc}) fashion. These fusion layers effectively integrate the complementary information in the multi-view time series and finally produce a prediction on the mood score. The {\deepmood} architecture is illustrated in \ref{fig:architecture}.

\subsection{Modeling One View}

Each view in the metadata is essentially a time series, and its length can vary a lot across instances which largely depends on the duration of a session in our study. For simplicity, we omit the superscript for views here. In order to model the sequential dependencies in each time series, we adopt the {\rnn} architecture \cite{mikolov2010recurrent,sutskever2011generating} which keeps hidden states over a sequence of elements and updates the hidden state $\mathbf{h}_k$ by the current input $\mathbf{x}_k$ as well as the previous hidden state $\mathbf{h}_{k-1}$ where $k>1$ with a recurrent function $\mathbf{h}_k = f(\mathbf{x}_k, \mathbf{h}_{k-1})$. The simplest form of an {\rnn} is
\begin{equation}
\begin{aligned}
\mathbf{h}_k = \sigma(\mathbf{W}\mathbf{x}_k+\mathbf{U}\mathbf{h}_{k-1})
\end{aligned}
\end{equation}
where $\mathbf{W} \in \mathbb{R}^{d_h \times d_p}$, and $\mathbf{U} \in \mathbb{R}^{d_h \times d_h}$ are model parameters that need to be learned, $d_p$ and $d_h$ are the input dimension and the number of recurrent units, respectively. $\sigma(\cdot)$ is a nonlinear transformation function such as \texttt{tanh}, \texttt{sigmoid}, and rectified linear unit (\texttt{ReLU}). Since {\rnn}s in such a form would fail to learn long term dependencies due to the vanishing gradient problem \cite{bengio1994learning,hochreiter1998vanishing}, they are not suitable to learn dependencies from a long input sequence in practice.

To make the learning procedure more effective over long sequences, the {\gru} \cite{cho2014learning} is proposed as a variation of the {\lstm} unit \cite{hochreiter1997long}. The {\gru} has been attracting great attentions since it overcomes the vanishing gradient problem in traditional {\rnn}s and is more efficient than the {\lstm} in some tasks \cite{chung2014empirical}. The {\gru} is designed to learn from previous timestamps with long time lags of unknown size between important timestamps via memory units that enable the network to learn to both update and forget hidden states based on new inputs.

A typical {\gru} is formulated as
\begin{equation}
\begin{aligned}
\mathbf{r}_k &= \texttt{sigmoid} (\mathbf{W}_r\mathbf{x}_k + \mathbf{U}_r\mathbf{h}_{k - 1}) \hfill \\
\mathbf{z}_k &= \texttt{sigmoid} (\mathbf{W}_z\mathbf{x}_k + \mathbf{U}_z\mathbf{h}_{k - 1}) \hfill \\
\tilde{\mathbf{h}}_k &= \texttt{tanh} (\mathbf{W}\mathbf{x}_k + \mathbf{U}(\mathbf{r}_k * \mathbf{h}_{k - 1})) \hfill \\
\mathbf{h}_k &= \mathbf{z}_k * \mathbf{h}_{k - 1} + (1 - \mathbf{z}_k) * \tilde {\mathbf{h}}_k \hfill \\
\end{aligned}
\end{equation}
where $*$ is the element-wise multiplication operator (Hadamard product). A reset gate $\mathbf{r}_k$ allows the {\gru} to forget the previously computed state $\mathbf{h}_{k-1}$, an update gate $\mathbf{z}_k$ balances between the previous state $\mathbf{h}_{k-1}$ and the candidate state $\tilde{\mathbf{h}}_k$, and the hidden state $\mathbf{h}_k$ can be considered as a compact representation of the input sequence from $\mathbf{x}_1$ to $\mathbf{x}_k$.

\subsection{Late Fusion on Multiple Views}

Let $\mathbf{h}^{(p)}\in\mathbb{R}^{d_h}$ denote the output vectors at the end of a sequence from the $p$-th view. We can consider $\mathcal{H}=\{\mathbf{h}^{(p)}\}_{p=1}^m$ as an intermediate representation of multi-view time series. We pursue a late fusion strategy to integrate the output vectors in $\mathcal{H}$. This avoids the issues of alignment and different lengths of time series in different views when performing early fusion directly on the input data.

\begin{figure}[t]
\centering
\subfigure[{\fc} layer.]{
\label{fig:fig_nn}
\begin{minipage}[l]{0.45\columnwidth}
  \centering
  \includegraphics[width=1\textwidth]{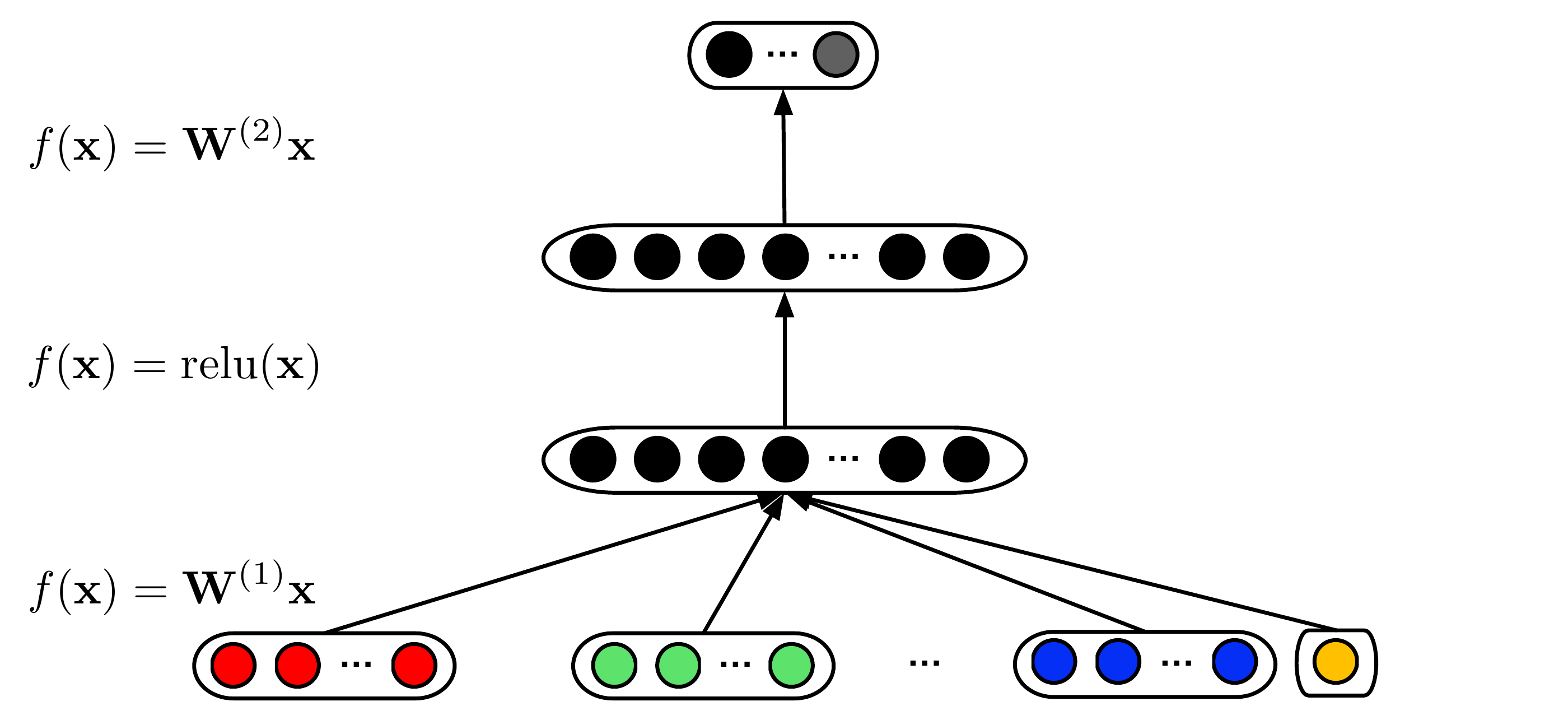}
\end{minipage}
}
\subfigure[{\fm} layer.]{
\label{fig:fig_fm}
\begin{minipage}[l]{0.45\columnwidth}
  \centering
  \includegraphics[width=1\textwidth]{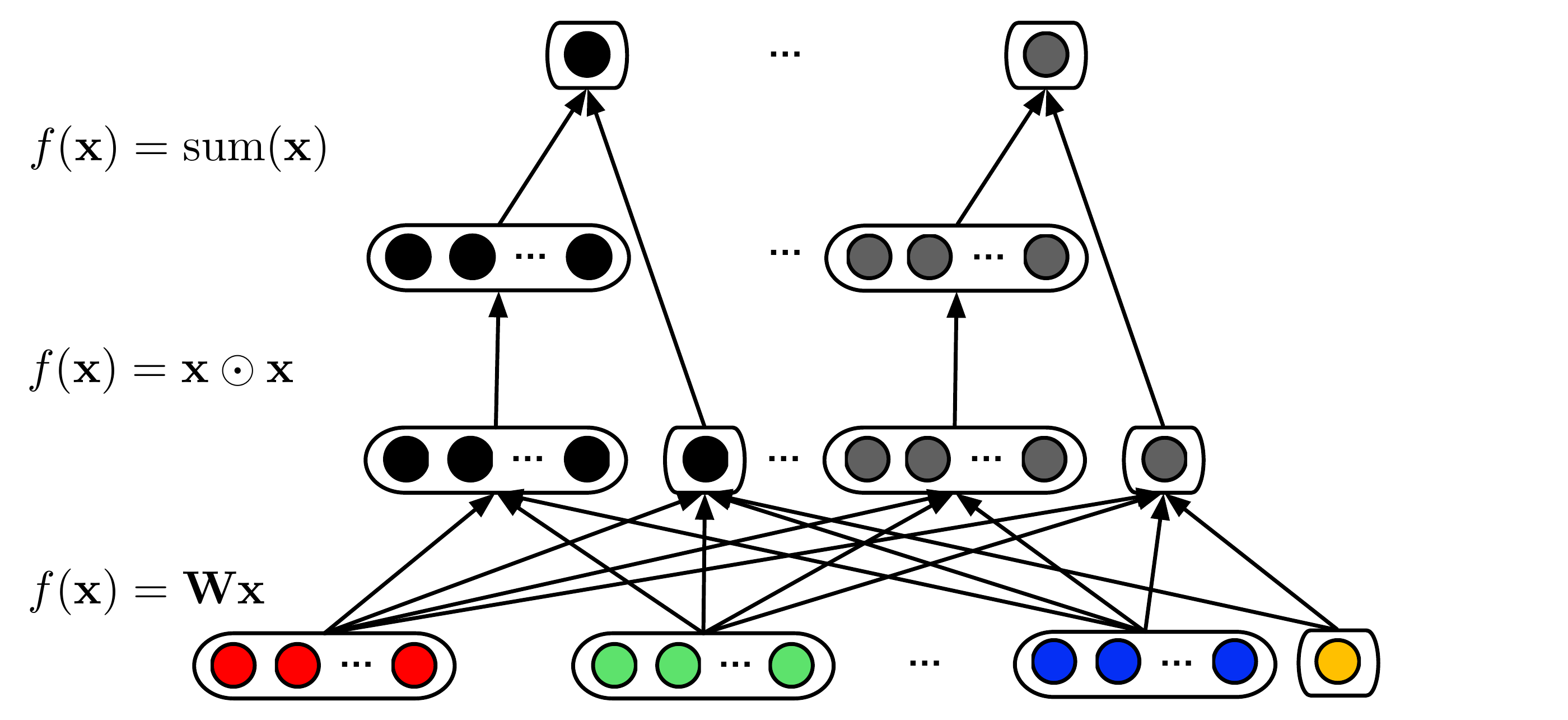}
\end{minipage}
}
\subfigure[{\mvm} layer.]{
\label{fig:fig_mvm}
\begin{minipage}[l]{0.45\columnwidth}
  \centering
  \includegraphics[width=1\textwidth]{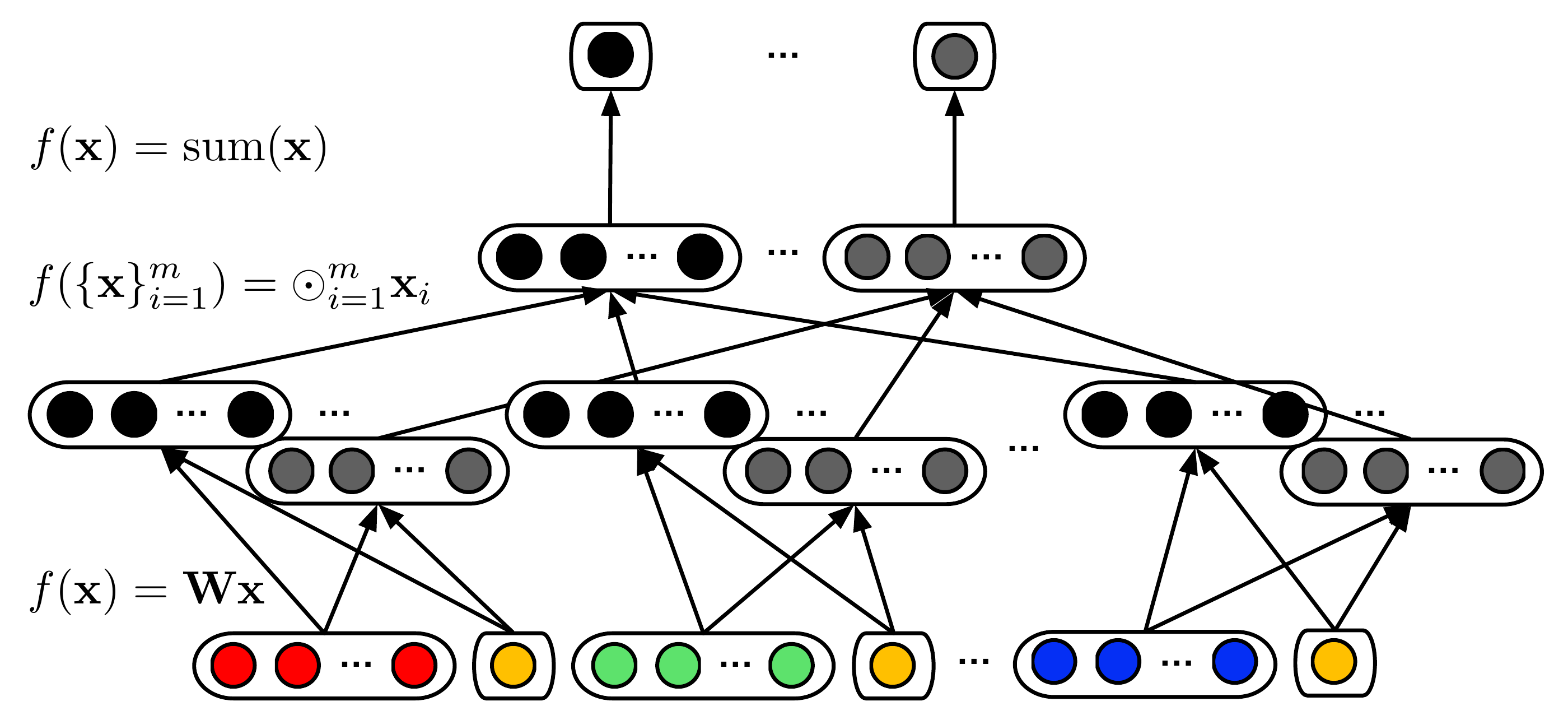}
\end{minipage}
}
\caption{Three strategies for multi-view data fusion.}
\label{fig:fig_graph}
\end{figure}

In the following we develop alternative methods for performing late fusion. These include not only a straightforward approach through an {\fc} layer to concatenate the features from different views, but also novel approaches to capturing interactions across the features of multiple views by exploring the concept of {\fm} \cite{rendle2012factorization} to capture the second-order feature interactions as well as the concept of {\mvm} \cite{cao2016multi} to capture higher order feature interactions, as shown in \ref{fig:fig_graph}.

\noindent\textbf{{\fc} layer}.
In order to generate a prediction on the mood score, a straightforward approach is to first concatenate features from multiple views together, \emph{i.e.}, $\mathbf{h} = [\mathbf{h}^{(1)}; \mathbf{h}^{(2)}; \cdots; \mathbf{h}^{(m)}] \in \mathbb{R}^{d_c}$, where $d_c$ is the total number of multi-view features, and typically $d_c=md_h$ for one-directional {\rnn}s and $d_c=2md_h$ for bidirectional {\rnn}s. We then feed forward $\mathbf{h}$ into one or several fully connected neural network layers with a nonlinear function $\sigma(\cdot)$ in between.
\begin{equation}
\begin{aligned}
\mathbf{q} &= \texttt{ReLU}(\mathbf{W}^{(1)}[\mathbf{h}; 1]) \\
\hat{\mathbf{y}} &= \mathbf{W}^{(2)}\mathbf{q}
\end{aligned}
\end{equation}
where $\mathbf{W}^{(1)} \in \mathbb{R}^{d_k \times (d_c+1)}, \mathbf{W}^{(2)} \in \mathbb{R}^{c \times d_k}$, $d_k$ is the number of hidden units, and the constant signal ``1'' is to model the global bias. Note that here we consider only one hidden layer between the {\gru} output layer and the final output layer as shown in Figure~\ref{fig:fig_nn} where red, green and blue represent the intermediate feature representations from different views, and yellow represents a constant signal ``1''.

\noindent\textbf{{\fm} layer}.
Rather than implicitly capturing nonlinearity through the transformation function, we consider modeling feature interactions in an explicit manner as shown in Figure~\ref{fig:fig_fm}.
\begin{equation}
\begin{aligned}
\label{eq:fm}
\mathbf{q}_a &= \mathbf{U}_a\mathbf{h} \\
b_a &= \mathbf{w}_a^\top[\mathbf{h}; 1] \\
\hat{y}_a &= \texttt{sum}([\mathbf{q}_a*\mathbf{q}_a;b_a])
\end{aligned}
\end{equation}
where $\mathbf{U}_a \in \mathbb{R}^{d_k \times d_c}, \mathbf{w}_a \in \mathbb{R}^{d_c+1}$, and $a$ denotes the $a$-th class. By denoting $\bar{\mathbf{h}} = [\mathbf{h}; 1]$, we can rewrite the decision function of $\hat{y_a}$ in \ref{eq:fm} as
\begin{equation}
\begin{aligned}
\hat{y}_a
&= \sum_{f=1}^{d_k} \left(\sum_{i=1}^{d_c} \mathbf{U}_a(f, i)\mathbf{h}(i)\right)^2 + \sum_{i=1}^{d_c+1} \mathbf{w}_a(i)\bar{\mathbf{h}}(i) \\
&= \sum_{f=1}^{d_k} \left(\sum_{i=1}^{d_c} \mathbf{U}_a(f, i)\mathbf{h}(i)\right) \left(\sum_{j=1}^{d_c} \mathbf{U}_a(f, j)\mathbf{h}(j)\right) + \sum_{i=1}^{d_c+1} \mathbf{w}_a(i)\bar{\mathbf{h}}(i) \\
&= \sum_{f=1}^{d_k} \sum_{i=1}^{d_c} \sum_{j=1}^{d_c} \mathbf{U}_a(f, i)\mathbf{U}_a(f, j)\mathbf{h}(i)\mathbf{h}(j) + \sum_{i=1}^{d_c+1} \mathbf{w}_a(i)\bar{\mathbf{h}}(i) \\
&= \sum_{i=1}^{d_c} \sum_{j=1}^{d_c} \left<\mathbf{U}_a(:, i),\mathbf{U}_a(:, j)\right>\mathbf{h}(i)\mathbf{h}(j) + \sum_{i=1}^{d_c} \mathbf{w}_a(i)\mathbf{h}(i) + \mathbf{w}_a(d_c+1)
\end{aligned}
\end{equation}

One can easily see that this is similar to the two-way {\fm} \cite{rendle2012factorization} except that the subscript $j$ ranges from $i+1$ to $d$ in the original form.

\noindent\textbf{{\mvm} layer}.
In contrast to modeling up to the second-order feature interactions as in the {\fm} layer, we further explore feature interactions up to the $m$th-order on the intermediate representations from $m$ views as shown in Figure~\ref{fig:fig_mvm}.
\begin{equation}
\begin{aligned}
\label{eq:mvm}
\mathbf{q}_a^{(p)} &= \mathbf{U}_a^{(p)}[\mathbf{h}^{(p)}; 1] \\
\hat{y}_a &= \texttt{sum}([\mathbf{q}_a^{(1)}*\cdots*\mathbf{q}_a^{(m)}])
\end{aligned}
\end{equation}
where $\mathbf{U}_a^{(p)} \in \mathbb{R}^{d_k \times (d_h+1)}$ is the factor matrix of the $p$-th view for the $a$-th class. By denoting $\bar{\mathbf{h}}^{(p)} = [\mathbf{h}^{(p)}; 1]$, we can verify that \ref{eq:mvm} is equivalent to {\mvm} \cite{cao2016multi}.
\begin{equation}
\begin{aligned}
\hat{y}_a
&= \sum_{f=1}^{d_k} \prod_{p=1}^m \left(\sum_{i_p=1}^{d_h+1} \mathbf{U}_a^{(p)}(f, i_p)\bar{\mathbf{h}}^{(p)}(i_p)\right) \\
&= \sum_{f=1}^{d_k} \sum_{i_1=1}^{d_h+1} \cdots \sum_{i_m=1}^{d_h+1} \left(\prod_{p=1}^m \mathbf{U}_a^{(p)}(f, i_p)\bar{\mathbf{h}}^{(p)}(i_p)\right) \\
&= \sum_{i_1=1}^{d_h+1} \cdots \sum_{i_m=1}^{d_h+1} \left(\sum_{f=1}^{d_k} \prod_{p=1}^m \mathbf{U}_a^{(p)}(f, i_p)\right) \left(\prod_{p=1}^m \bar{\mathbf{h}}^{(p)}(i_p)\right)
\end{aligned}
\end{equation}

As shown in \ref{fig:architecture}, the full-order feature interactions across multiple views are modeled in a tensor and factorized in a collective manner.
Note that a dropout layer \cite{hinton2012improving} is applied before feeding the output from {\gru} to the fusion layer which is a regularization method designed to prevent co-adaptation of feature detectors in deep neural networks. 

Following the computational graph, it is straightforward to compute gradients for model parameters in both the {\mvm} layer and the {\fm} layer, as we do for the conventional {\fc} layer. Therefore, the error messages generated from the loss function on the final mood score can be back-propagated through these fusion layers all the way to the very beginning, \emph{i.e.}, $\mathbf{W}_r$, $\mathbf{U}_r$, $\mathbf{W}_z$, $\mathbf{U}_z$, $\mathbf{W}$, and $\mathbf{U}$ in {\gru} for each input view. In this manner, we can say that {\deepmood} is an end-to-end learning framework for mood detection.

\section{Experiments}
\label{sec:deepmood_exp}

In order to evaluate the performance of {\deepmood} on mood detection, we conduct experiments on the metadata of typing dynamics as introduced in Section~\ref{sec:deepmood_data}. We investigate a session-level prediction problem. That is to say, we use features of alphanumeric characters, special characters and accelerometer values in a session to predict the mood score of a user.

\subsection{Compared Methods}

\begin{table}[t]
\centering
\caption{Parameter configuration in {\deepmood}.}
\label{tab:para}
\begin{tabular}{||l|c||}
\hline
Parameters & Values \\
\hline\hline
\# recurrent units ($d_h$)  & 4, 8, 16 \\
\# hidden units ($d_k$)     & 4, 8, 16 \\
\# epochs                   & 500 \\
Batch size                  & 256 \\
Learning rate               & 0.001 \\
Dropout rate                & 0.1 \\
Maximum sequence length     & 100 \\
Minimum sequence length     & 10 \\
\hline
\end{tabular}
\end{table}

We truncate sessions that contain more than 100 keypresses, and we remove sessions if any of their views contain less than 10 keypresses. It leaves us with 14,613 total samples which are then split in a chronological order for training and validation on a per user basis. Each user contributes first 80\% of her sessions for training and the rest for validation. The compared methods are summarized as follows:
\begin{itemize}[leftmargin=*,noitemsep,topsep=0pt]
\item\textbf{\dmvm}, \textbf{\dfm}, and \textbf{\dnn}: the proposed {\deepmood} model with an {\mvm} layer, an {\fm} layer, and an {\fc} layer for data fusion, respectively. The implementation is completed using {\keras} \cite{chollet2015keras} with {\tensorflow} \cite{tensorflow2015-whitepaper} as the backend. The code has been made available at {\github}\footnote{\url{https://github.com/caobokai/DeepMood}}. Specifically, a bidirectional {\gru} is applied on each view of the metadata. {\rmsprop} \cite{tieleman2012lecture} is used as the optimizer. The number of recurrent units and the number of hidden units are selected on the validation set. We empirically set other parameters, including the number of epochs, batch size, learning rate, and dropout rate. Detailed configurations of the hyperparameters are summarized in \ref{tab:para}.
\item\textbf{\xgb}: a tree boosting system\footnote{\url{https://github.com/dmlc/xgboost}} \cite{chen2016xgboost}. We concatenate the sequence data with the maximum length 100 (padding 0 for short ones) of each feature as the input.
\item\textbf{\svm} and \textbf{\lr}: two linear models implemented in {\scikit}\footnote{\url{http://scikit-learn.org}}. For classification, {\svm} refers to \emph{Linear Support Vector Classification}, and {\lr} refers to \emph{Logistic Regression}. For regression, {\svm} refers to \emph{Linear Support Vector Regression}, and {\lr} refers to \emph{Ridge Regression}. The input setting is the same as {\xgb}.
\end{itemize}

In general, {\dmvm}, {\dfm} and {\dnn} can be categorized as late fusion approaches, while {\xgb}, {\svm} and {\lr} are early fusion strategies for the multi-view sequence prediction problem. Note that the number of model parameters in the {\mvm} layer and the {\fm} layer is $c d_k (d_c+m)$ and $c d_k d_c + c(d_c+1)$, respectively, thereby leading to approximately the same model complexity $O(c d_k d_c)$ due to $m \ll d_c$.
For the {\fc} layer, the number of model parameters is $d_k(d_c+c+1)$ that results in a model complexity $O(d_k d_c)$ due to $c \ll d_c$. For a fair comparison, we control the model complexity of different fusion layers at the same level by setting $d_k$ in the {\fc} layer as $c$ times as that in the {\mvm} layer and the {\fm} layer.

Experiments on the depression score {\hdrs} are conducted as a binary classification task where $c=2$. According to the recommended severity ranges for the {\hdrs} 
\cite{zimmerman2013severity}, we consider sessions with the {\hdrs} score between 0 and 7 (inclusive) as negative samples (normal) and those with {\hdrs} greater than or equal to 8 as positive samples (from mild depression to severe depression). On the other hand, the mania score {\ymrs} is more complicated without a widely adopted threshold. Therefore, {\ymrs} is directly used as the label for a regression task where $c=1$. Accuracy and {\fone} score are used to evaluate the classification task, and root-mean-square error ({\rmse}) is used for the regression task.

\begin{table}[t]
\centering
\caption{Prediction performance for mood detection.}
\label{tab:mainresult}
\newcolumntype{x}[1]{>{\centering\arraybackslash}p{#1}}
\begin{tabular}{||l|x{2cm}|x{2cm}|x{2cm}||}
\hline
\multirow{2}*{Methods} & \multicolumn{2}{c|}{Classification} & Regression \\
\cline{2-4} & Accuracy & {\fone} & {\rmse} \\
\hline\hline
{\dmvm}	& 0.9031 & 0.9070 & 3.5664 \\
{\dfm}	& 0.9021 & 0.9029 & 3.6767 \\
{\dnn}	& 0.8868 & 0.8929 & 3.7874 \\
{\xgb}	& 0.8555 & 0.8562 & 3.9634 \\
{\svm}	& 0.7323 & 0.7237 & 4.1257 \\
{\lr}	& 0.7293 & 0.7172 & 4.1822 \\
\hline
\end{tabular}
\end{table}

\subsection{Prediction Performance}

Experimental results are shown in \ref{tab:mainresult}. It is found that the late fusion based {\deepmood} models achieve the best prediction performance on the dichotomized {\hdrs} score, especially {\dmvm} and {\dfm} with 90.31\% and 90.21\%, respectively. It demonstrates the feasibility of using passive typing dynamics from mobile phone metadata to predict the disturbance and severity of mood states. In addition, it is found that {\svm} and {\lr} are not a good fit to this task, or sequence prediction in general. {\xgb} performs reasonably well as an ensemble method, but {\dmvm} still outperforms it by a significant margin 5.56\%, 5.93\% and 10.02\% in terms of accuracy, {\fone} score and {\rmse}, respectively. Among the {\deepmood} variations, the improvement of {\dmvm} and {\dfm} over {\dnn} reveals the potential of replacing a conventional {\fc} layer with an {\mvm} layer or an {\fm} layer for data fusion in a deep framework, because {\mvm} and {\fm} model the higher order feature interactions in an explicit manner.

\begin{figure}[t]
\centering
\begin{minipage}[l]{0.6\columnwidth}
  \centering
  \includegraphics[width=1\textwidth]{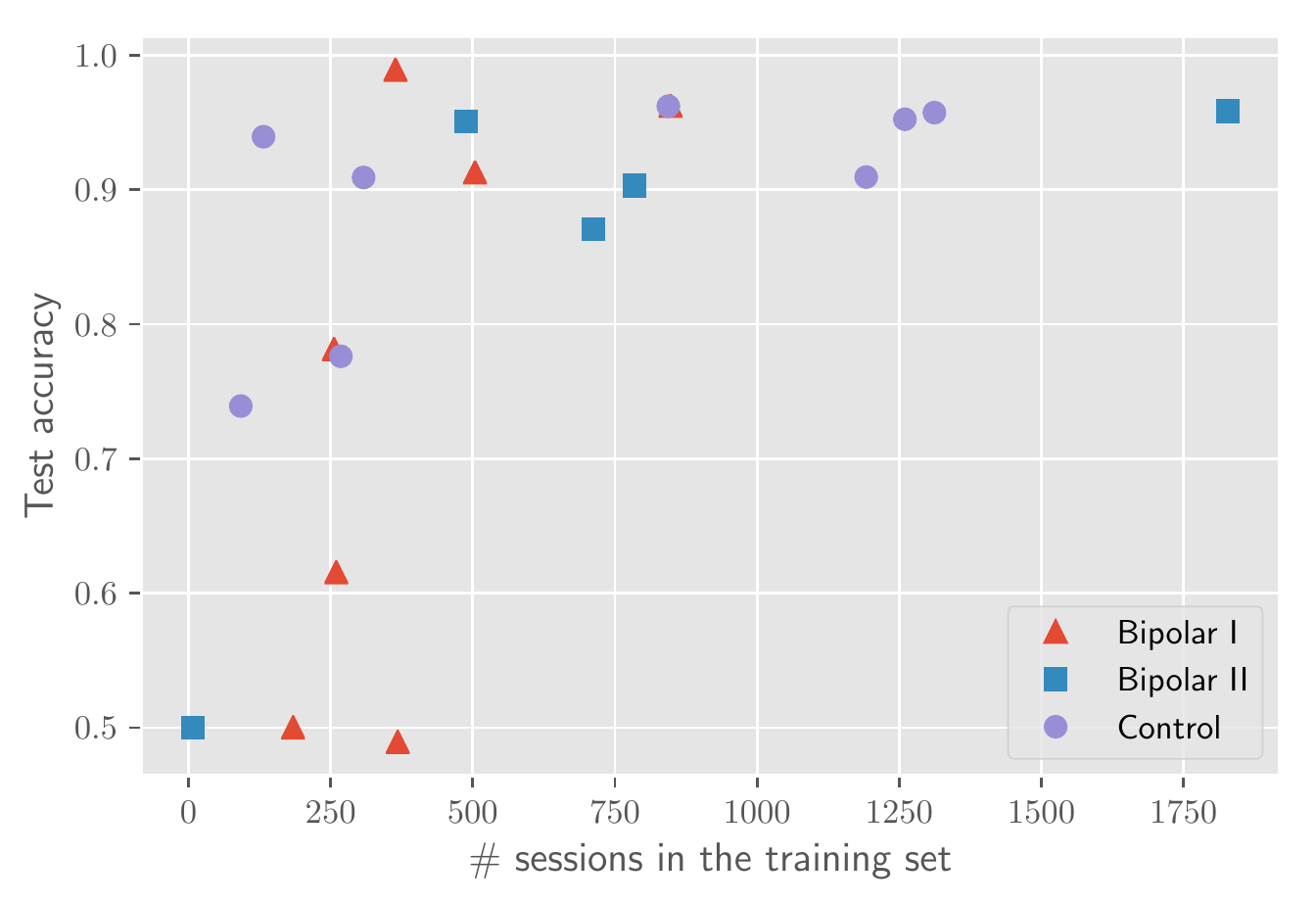}
\end{minipage}
\caption{Prediction performance of {\dmvm} per individual.}
\label{fig:acc_per_subj}
\end{figure}

In practice, it is important to understand how the model works for each user when monitoring her mood states. Therefore, we investigate the prediction performance of {\dmvm} on each of the 20 users in our study. Results are shown in \ref{fig:acc_per_subj} where each dot represents a user with the number of her contributed sessions in the training set and the corresponding prediction accuracy. We can observe that the proposed model can steadily produce accurate predictions ($\ge$87\%) of a user's mood states when she provides more than 400 valid typing sessions in the training phase. Note that the prediction we make in this work is on a per session basis which is typically less than one minute. We can expect more accurate results on a daily level by ensembling sessions occurring during one day.

\begin{figure}[t]
\centering
\begin{minipage}[l]{0.6\columnwidth}
  \centering
  \includegraphics[width=1\textwidth]{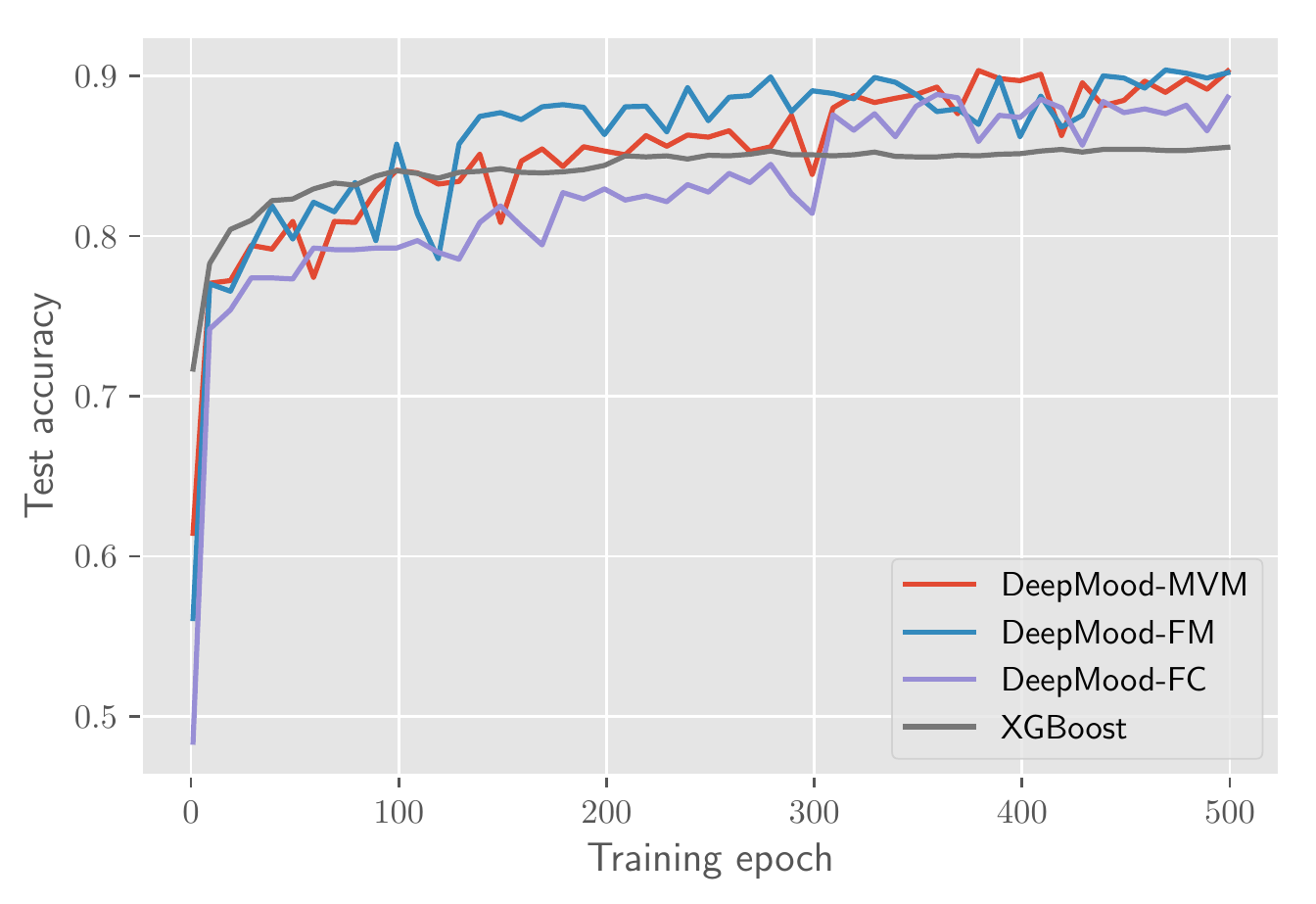}
\end{minipage}
\caption{Learning curves for mood detection.}
\label{fig:epoch}
\end{figure}

\subsection{Convergence Efficiency}

We show more details about the learning procedure of the proposed {\deepmood} model with different fusion layers and that of {\xgb}. \ref{fig:epoch} illustrates how the accuracy on the validation set changes over epochs. It is found that different fusion layers have different convergence performance in the first 300 epochs, and afterwards they steadily outperform {\xgb}. Among the {\deepmood} models, we can observe that {\dmvm} and {\dfm} converge more efficiently than {\dnn} in the first 300 epochs, and they are able to reach a better local minima of the loss function at the end. This again shows the importance of the fusion layer in a deep framework. It is also found that the generalizability of {\xgb} on the sequence prediction task is limited, although its training error could perfectly converge to 0 at an early stage.

\subsection{Importance of Different Views}

To better understand the role that different views play in the buildup of mood detection by {\deepmood}, we examine separate models trained with or without each view. Since {\dmvm} is designed for heterogeneous data fusion, \emph{i.e.}, data with at least two views, we train {\dmvm} on every pairwise views. Moreover, we train {\dfm} on every single view. Experimental results are shown in \ref{tab:subsetresult}. Firstly, we observe that alphanumeric characters and accelerometer values have a significantly better prediction performance on mood states than special characters, and alphanumeric characters are the best individual predictors of mood states. It validates a high correlation between the mood disturbance and typing patterns including duration of a keypress, time interval since the last keypress, as well as accelerometer values.

\begin{table}[t]
\centering
\caption{Prediction performance with different views of typing dynamics.}
\label{tab:subsetresult}
\newcolumntype{x}[1]{>{\centering\arraybackslash}p{#1}}
\begin{tabular}{||l|x{2cm}|x{2cm}|x{2cm}||}
\hline
\multirow{2}*{Methods} & \multicolumn{2}{c|}{Classification} & Regression \\
\cline{2-4} & Accuracy & {\fone} & {\rmse} \\
\hline\hline
{\dmvm} w/o {\ch}	& 0.8125 & 0.8164 & 3.9833 \\
{\dmvm} w/o {\nonch}	& 0.9008 & 0.9034 & 3.8166 \\
{\dmvm} w/o {\accel}	& 0.8318 & 0.8253 & 3.9499 \\
\hline
{\dmvm} w/ all	& 0.9031 & 0.9070 & 3.5664 \\
\hline\hline
{\dfm} w/ {\ch}	& 0.8322 & 0.8224 & 3.9515 \\
{\dfm} w/ {\nonch}	& 0.6260 & 0.5676 & 4.1040 \\
{\dfm} w/ {\accel}	& 0.8015 & 0.8089 & 3.9722 \\
\hline
{\dfm} w/ all	& 0.9021 & 0.9011 & 3.6767 \\
\hline
\end{tabular}
\end{table}

\section{Related Work}

This work is studied in the context of supervised sequence prediction. Xing et al.~provided a brief survey on the sequence prediction problem where sequence data are categorized into five subtypes: simple symbolic sequences, complex symbolic sequences, simple time series, multivariate time series, and complex event sequences; sequence classification methods are grouped into three subtypes: feature based methods, sequence distance based methods, and model based methods \cite{xing2010brief}. Feature based methods first transform a sequence into a feature vector and then apply conventional classification models \cite{lesh1999mining,aggarwal2002effective,leslie2004fast,ji2007mining,ye2009time}. Distance based methods include $k$-nearest neighbor classifier \cite{keogh2000scaling,keogh2003need,ratanamahatana2004making,wei2006semi,xi2006fast,ding2008querying} and {\svm} with local alignment kernels \cite{lodhi2002text,she2003frequent,sonnenburg2005large} by measuring the similarity between a pair of sequences. Model based methods assume that sequences in a class are generated by an underlying probability distribution, including Naive Bayes \cite{cheng2005protein}, Markov Model \cite{yakhnenko2005discriminatively} and Hidden Markov Model \cite{srivastava2007hmm}.

However, most of the works focus on simple symbolic sequences and simple time series, with a few on complex symbolic sequences and multivariate time series. The problem of classifying complex event sequence data (a combination of multiple numerical measurements and categorical fields) still needs further investigation which motivates this work. In addition, most of the methods are devoted to shallow models with feature engineering. Inspired by the great success of deep {\rnn}s in the applications to many sequence tasks, including speech recognition \cite{graves2013speech} and natural language processing \cite{mikolov2010recurrent,bahdanau2014neural}, in this work, we develop a deep architecture to model complex event sequences of mobile phone typing dynamics.

Rendle pioneered the concept of feature interactions in {\fm} \cite{rendle2010factorization}.
Juan et al.~presented Field-aware Factorization Machine (\texttt{FFM}) to allow each feature to interact differently with another feature depending on its field \cite{juan2016field}.
Lu et al.~studied the multi-view feature interactions in the context of multi-task learning \cite{lu2017multilinear} and overlapping groups of views \cite{lu2018learning}.
Novikov et al.~proposed Exponential Machine (\texttt{ExM}) where the weight tensor is represented in a factorized format called Tensor Train \cite{novikov2016exponential}. 
Zhang et al.~used {\fm} to initialize the embedding layer in a deep model \cite{zhang2016deep}.
Qu et al.~developed a product layer on the top of the embedding layer to increase the model capacity \cite{qu2016product}.
Other extensions of {\fm} to deep architectures include Neural Factorization Machine (\texttt{NFM}) \cite{he2017neural} and Attentional Factorization Machine (\texttt{AFM}) \cite{xiao2017attentional}.
In order to effectively model feature interactions, a variety of models have been developed in the industry as well. Microsoft studied feature interactions in deep models, including Deep Semantic Similarity Model (\texttt{DSSM}) \cite{huang2013learning}, Deep Crossing \cite{shan2016deep} and Deep Embedding Forest \cite{zhu2017deep}. They use features as raw as possible without manually crafted combinatorial features and let deep neural networks take care of the rest. Alibaba proposed Deep Interest Network (\texttt{DIN}) to learn user embeddings as a function of ad embeddings \cite{zhou2017deep}. Google used deep neural networks to learn from heterogeneous signals for YouTube recommendations \cite{covington2016deep}.
In addition, \texttt{Wide \& Deep} was developed for app recommender systems in Google Play where the wide component includes cross features that are good at memorization and the deep component includes embedding layers for generalization \cite{cheng2016wide}. Guo et al.~further proposed to use {\fm} as the wide component in \texttt{Wide \& Deep} with shared embeddings in the deep component \cite{guo2017deepfm}. Wang et al.~developed \texttt{Deep \& Cross} to learn explicit cross features of bounded degrees \cite{wang2017deep}. In this work, we model feature interactions on intermediate representations of multi-view time series data to fuse heterogeneous signals in a deep framework.

\chapter{Conclusion}

In this thesis, we have explored the idea of broad learning for healthcare. Towards this direction, we thoroughly studied four different research problems: multi-view feature selection, subgraph pattern mining, brain network embedding, and multi-view sequence prediction. The effectiveness of the proposed models was corroborated by extensive experiments on real-world datasets. The contributions of our works are fourfold:

\begin{enumerate}[leftmargin=*,noitemsep,topsep=0pt]

\item We studied the problem of multi-view feature selection. Tensor product was utilized to bring multiple views together in a joint space, and a dual method named {\mvfs} was presented to perform feature selection. Empirical studies on an {\hiv} dataset demonstrated that features selected by {\mvfs} could achieve a better classification performance and were relevant to disease diagnosis. Such a solution has broad applicability to many biomedical and healthcare problems. Capability for simultaneous analysis of multiple feature groups has transformative potential for yielding new insights concerning risk and protective relationships, for clarifying disease mechanisms, for aiding diagnostics and clinical monitoring, for biomarker discovery, for identification of new treatment targets, and for evaluating effects of intervention.

\item We presented an approach to selecting discriminative subgraph features using side views. It was shown that by leveraging the side information from clinical and other measures that are available along with the {\fmri} and {\dti} brain networks, the proposed method named {\gmsv} could achieve a superior performance on the problem of feature selection for graph classification, and the selected subgraph patterns were relevant to disease diagnosis. {\gmsv} has important applications to neurological disorder identification through brain networks, as well as other domains where one can find graph data with associated side information. For example, in bioinformatics, chemical compounds can be represented by graphs based on their inherent molecular structures, and each of them has properties such as drug repositioning, side effects, ontology annotations. Leveraging all the information to find out discriminative subgraph patterns can be transformative for drug discovery.

\item We developed a novel constrained tensor factorization model named {\bne} for brain network embedding. Specifically, brain networks were stacked as a partially symmetric tensor which was then factorized in conjunction with the side information guidance, the orthogonal constraint, and the classifier learning procedure. It allowed us to obtain discriminative and distinct latent factors which include brain network representations. In the experiments on an anxiety disorder dataset that contains {\eeg} brain networks and self-report data, we demonstrated the superior performance of {\bne} on the graph classification task. It is also possible to incorporate different types of guidance and supervision and learn a joint representation from multimodal brain network data.

\item We validated that mobile phone typing dynamics metadata could be used to predict the presence of mood disorders. The proposed end-to-end deep learning architecture named {\deepmood} was able to significantly outperform the state-of-the-art approaches in terms of prediction accuracy. Experimental results showed that late fusion was more effective than aligning or concatenating the multi-view time series data at an early stage, and more sophisticated fusion layers also helped. The ability to passively collect data that can be used to infer the presence and severity of mood disturbance may enable providers to provide interventions to more patients earlier in their mood episodes. Models such as {\deepmood} presented in this work may also lead to deeper understanding of the effects of mood disturbance in the daily activities of people with mood disorders.

\end{enumerate}

In addition to the four tasks elaborated in this thesis, my colleagues and I have been working on other bioinformatics and healthcare related problems as well, including drug discovery in heterogeneous bioinformatics networks \cite{cao2014collective,kong2013meta,kong2013multi}, brain network clustering \cite{liu2018multi,ma2016multi}, and deep learning on brain network and neuroimaging data \cite{zhang2016identifying,wang2017structural,zheng2017novel}. We hope that our efforts could facilitate large-scale data-driven artificial intelligence applications to the healthcare domain, including computer-aided diagnosis, precision medicine, and mobile health. 

For future work, we think that federated learning is critical for deploying intelligent healthcare services on mobile devices while preserving user privacy, meta learning for transferring and sharing medical knowledge across users, and reinforcement learning for providing effective interventions to users. Most importantly, broad learning is a promising research direction with the target to find a general framework of fusing heterogeneous signals for synergistic knowledge discovery. Although neural network models are capable of approximating many complicated functions in theory, there are practical difficulties in the learning process (\emph{e.g.}, local minima). Therefore, analogous to convolution layers for computer vision and recurrent layers for sequence modeling, it is critical to develop suitable neural network modules (\emph{e.g.}, fusion layers or interaction operators) for fusing heterogeneous data sources in healthcare applications and web services.

\appendices
\newpage
\appendix



\section*{Supplementary material}

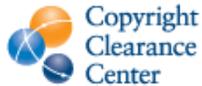 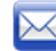

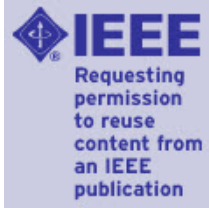

| | |
|---|---|
| **Title:** | Tensor-Based Multi-view Feature Selection with Applications to Brain Diseases |
| **Conference Proceedings:** | Data Mining (ICDM), 2014 IEEE International Conference on |
| **Author:** | Bokai Cao |
| **Publisher:** | IEEE |
| **Date:** | Dec. 2014 |
| Copyright © 2014, IEEE |

## Thesis / Dissertation Reuse

**The IEEE does not require individuals working on a thesis to obtain a formal reuse license, however, you may print out this statement to be used as a permission grant:**

*Requirements to be followed when using any portion (e.g., figure, graph, table, or textual material) of an IEEE copyrighted paper in a thesis:*

1) In the case of textual material (e.g., using short quotes or referring to the work within these papers) users must give full credit to the original source (author, paper, publication) followed by the IEEE copyright line © 2011 IEEE.
2) In the case of illustrations or tabular material, we require that the copyright line © [Year of original publication] IEEE appear prominently with each reprinted figure and/or table.
3) If a substantial portion of the original paper is to be used, and if you are not the senior author, also obtain the senior author's approval.

*Requirements to be followed when using an entire IEEE copyrighted paper in a thesis:*

1) The following IEEE copyright/ credit notice should be placed prominently in the references: © [year of original publication] IEEE. Reprinted, with permission, from [author names, paper title, IEEE publication title, and month/year of publication]
2) Only the accepted version of an IEEE copyrighted paper can be used when posting the paper or your thesis on-line.
3) In placing the thesis on the author's university website, please display the following message in a prominent place on the website: In reference to IEEE copyrighted material which is used with permission in this thesis, the IEEE does not endorse any of [university/educational entity's name goes here]'s products or services. Internal or personal use of this material is permitted. If interested in reprinting/republishing IEEE copyrighted material for advertising or promotional purposes or for creating new collective works for resale or redistribution, please go to http://www.ieee.org/publications_standards/publications/rights/rights_link.html to learn how to obtain a License from RightsLink.

If applicable, University Microfilms and/or ProQuest Library, or the Archives of Canada may supply single copies of the dissertation.

BACK    CLOSE WINDOW

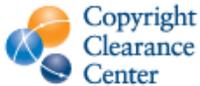

RightsLink®

Home | Create Account | Help | 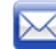

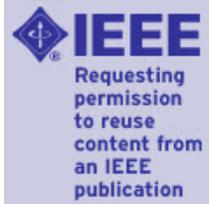

**IEEE**
Requesting permission to reuse content from an IEEE publication

**Title:** Mining Brain Networks Using Multiple Side Views for Neurological Disorder Identification

**Conference Proceedings:** Data Mining (ICDM), 2015 IEEE International Conference on

**Author:** Bokai Cao

**Publisher:** IEEE

**Date:** Nov. 2015

Copyright © 2015, IEEE

## Thesis / Dissertation Reuse

**The IEEE does not require individuals working on a thesis to obtain a formal reuse license, however, you may print out this statement to be used as a permission grant:**

*Requirements to be followed when using any portion (e.g., figure, graph, table, or textual material) of an IEEE copyrighted paper in a thesis:*

1) In the case of textual material (e.g., using short quotes or referring to the work within these papers) users must give full credit to the original source (author, paper, publication) followed by the IEEE copyright line © 2011 IEEE.
2) In the case of illustrations or tabular material, we require that the copyright line © [Year of original publication] IEEE appear prominently with each reprinted figure and/or table.
3) If a substantial portion of the original paper is to be used, and if you are not the senior author, also obtain the senior author's approval.

*Requirements to be followed when using an entire IEEE copyrighted paper in a thesis:*

1) The following IEEE copyright/ credit notice should be placed prominently in the references: © [year of original publication] IEEE. Reprinted, with permission, from [author names, paper title, IEEE publication title, and month/year of publication]
2) Only the accepted version of an IEEE copyrighted paper can be used when posting the paper or your thesis on-line.
3) In placing the thesis on the author's university website, please display the following message in a prominent place on the website: In reference to IEEE copyrighted material which is used with permission in this thesis, the IEEE does not endorse any of [university/educational entity's name goes here]'s products or services. Internal or personal use of this material is permitted. If interested in reprinting/republishing IEEE copyrighted material for advertising or promotional purposes or for creating new collective works for resale or redistribution, please go to http://www.ieee.org/publications_standards/publications/rights/rights_link.html to learn how to obtain a License from RightsLink.

If applicable, University Microfilms and/or ProQuest Library, or the Archives of Canada may supply single copies of the dissertation.

BACK | CLOSE WINDOW

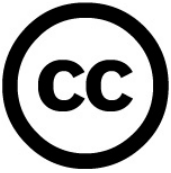

# Creative Commons License Deed

## Attribution 4.0 International (CC BY 4.0)

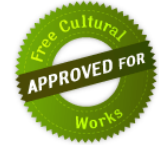

This is a human-readable summary of (and not a substitute for) the license.

# You are free to:

**Share** — copy and redistribute the material in any medium or format

**Adapt** — remix, transform, and build upon the material

for any purpose, even commercially.

The licensor cannot revoke these freedoms as long as you follow the license terms.

# Under the following terms:

**Attribution** — You must give appropriate credit, provide a link to the license, and indicate if changes were made. You may do so in any reasonable manner, but not in any way that suggests the licensor endorses you or your use.

**No additional restrictions** — You may not apply legal terms or technological measures that legally restrict others from doing anything the license permits.

# Notices:

You do not have to comply with the license for elements of the material in the public domain or where your use is permitted by an applicable exception or limitation.

No warranties are given. The license may not give you all of the permissions necessary for your intended use. For example, other rights such as publicity, privacy, or moral rights may limit how you use the material.

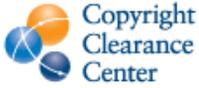 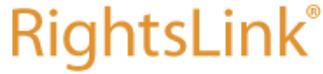

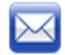
**SPRINGER NATURE**

**Title:**        Semi-supervised Tensor Factorization for Brain Network Analysis

**Author:**     Bokai Cao, Chun-Ta Lu, Xiaokai Wei et al

**Publication:** Springer eBook

**Publisher:**   Springer Nature

**Date:**        Jan 1, 2016

Copyright © 2016, Springer Nature

## Author Use

Authors of Springer Nature content do not require permission to use content from their article in most cases as stated in the author's guidelines.

Authors wishing to use their article for commercial purposes must request permission in the normal way.

For journal content questions, please contact Springer Nature's permissions department at journalpermissions@springernature.com

For book content questions, please contact Springer Nature's permissions department at bookpermissions@springernature.com

BACK | CLOSE WINDOW

**In order for SIAM to include your paper in the 2017 SIAM International Conference on Data Mining proceedings, the following Copyright Transfer Agreement must be agreed to during the paper upload process.**

### COPYRIGHT TRANSFER AGREEMENT

Title of Paper:

Author(s):

Copyright to this paper is hereby irrevocably assigned to SIAM for publication in the **2016 SIAM International Conference on Data Mining (SDM17)**, April 27 - 29, 2017 at The Westin Galleria Houston, Houston, Texas, USA. SIAM has sole use for distribution in all forms and media, such as microfilm and anthologies, except that the author(s) or, in the case of a "work made for hire," the employer will retain:

> The right to use all or part of the content of the paper in future works of the author(s), including author's teaching, technical collaborations, conference presentations, lectures, or other scholarly works and professional activities as well as to the extent the fair use provisions of the U.S. Copyright Act permit. If the copyright is granted to SIAM, then the proper notice of the SIAM's copyright should be provided.

> The right to post the final draft of the paper on noncommercial pre-print serves like arXiv.org.

> The right to post the final version of the paper on the author's personal web site and on the web server of the author's institution, provided the proper notice of the SIAM's copyright is included and that no separate or additional fees are collected for access to or distribution of the paper.

> The right to refuse permission to third parties to republish all or part of the paper or translation thereof.

It is affirmed that neither this paper nor portions of it have been published elsewhere and that a copyright transfer agreement has not been signed permitting the publication of a similar paper in a journal or elsewhere. For multi-author works, the signing author agrees to notify all co-authors of his/her action.

**Transfer of Copyright to the Publisher**

*SIAM strongly recommends this option. This transfer of copyright provides SIAM the legal basis not only to publish and to distribute the work, but also to pursue infringements of copyright (such as plagiarism and other forms of unauthorized use) and to grant permissions for the legitimate uses of the work by third parties. This option should not be selected if the work was prepared by a government office or employee as part of his or her official duties.*

[__]  By selecting the box at left, the Author hereby irrevocably assigns, conveys and transfers the copyright to the Work to SIAM. SIAM shall have sole rights of distribution and/or publication of the work in all forms and media, throughout the world, except for those rights given to the Author above.

[__]  By selecting the box at left, the Author DOES NOT assign, convey and transfer the


\centerline{\Large ACM Copyright Letter\footnote{\url{http://authors.acm.org/main.html}}}
Authors can reuse any portion of their own work in a new work of their own (and no fee is expected) as long as a citation and DOI pointer to the Version of Record in the ACM Digital Library are included.

Authors can include partial or complete papers of their own (and no fee is expected) in a dissertation as long as citations and DOI pointers to the Versions of Record in the ACM Digital Library are included. Authors can use any portion of their own work in presentations and in the classroom (and no fee is expected).





\nocite{*}
\bibformc
\bibliography{uictest}
\newpage
\vita

\begin{longtable}{p{.22\textwidth}>{\baselineskip=15pt}p{.72\textwidth}}
\textbf{NAME} & Bokai Cao \\ 
\textbf{EDUCATION}
& \textbf{Ph.D., Computer Science}, University of Illinois at Chicago, 2018.\\
& \textbf{B.Eng., Computer Science}, Renmin University of China, 2013.\\
& \textbf{B.Sc., Mathematics}, Renmin University of China, 2013.\\
\textbf{PUBLICATIONS}
& Chun-Ta Lu, Lifang He, Hao Ding, Bokai Cao, and Philip S. Yu. Learning from multi-view multi-way data via structural factorization machines. WWW, 2018.\\ 
& Ye Liu, Lifang He, Bokai Cao, Philip S. Yu, Ann B. Ragin, and Alex D. Leow. Multi-view multi-graph embedding for brain network clustering analysis. AAAI, 2018.\\ 
& Jingyuan Zhang, Chun-Ta Lu, Bokai Cao, Yi Chang, and Philip S. Yu. Connecting emerging relationships from news via tensor factorization. Big Data, 2017.\\ 
& Lei Zheng, Bokai Cao, Nianzu Ma, Vahid Noroozi, and Philip S. Yu. Hierarchical collaborative embedding for context-aware recommendations. Big Data, 2017.\\ 
& Bokai Cao, Mia Mao, Siim Viidu, and Philip S. Yu. HitFraud: A broad learning approach for collective fraud detection in heterogeneous information networks. ICDM, 2017.\\ 
& Xiaokai Wei, Bokai Cao, and Philip S. Yu. Unsupervised feature selection with heterogeneous side information. CIKM, 2017.\\ 
& Xiaokai Wei, Sihong Xie, Bokai Cao, and Philip S. Yu. Rethinking unsupervised feature selection: From pseudo labels to pseudo must-links. ECML/PKDD, 2017.\\ 
& Lichao Sun, Yuqi Wang, Bokai Cao, Philip S. Yu, Witawas Srisa-an, and Alex D. Leow. Sequential keystroke behavioral biometrics for user identification via multi-view deep learning. ECML/PKDD, 2017.\\ 
& Bokai Cao, Lei Zheng, Chenwei Zhang, Philip S. Yu, Andrea Piscitello, John Zulueta, Olu Ajilore, Kelly Ryan, and Alex D. Leow. DeepMood: Modeling mobile phone typing dynamics for mood detection. KDD, 2017.\\ 
& Shen Wang, Lifang He, Bokai Cao, Chun-Ta Lu, Philip S. Yu, and Ann B. Ragin. Structural deep brain network mining. KDD, 2017.\\ 
& Bokai Cao, Mia Mao, Siim Viidu, and Philip S. Yu. Collective fraud detection capturing inter-transaction dependency. KDD Workshop on Anomaly Detection in Finance, 2017.\\ 
& Xiaokai Wei, Bokai Cao, and Philip S. Yu. Multi-view unsupervised feature selection by cross-diffused matrix alignment. IJCNN, 2017.\\ 
& Bokai Cao, Lifang He, Xiaokai Wei, Mengqi Xing, Philip S. Yu, Heide Klumpp, and Alex D. Leow. t-BNE: Tensor-based brain network embedding. SDM, 2017.\\ 
& Xiaokai Wei, Linchuan Xu, Bokai Cao, and Philip S. Yu. Cross view link prediction by learning noise-resilient representation consensus. WWW, 2017.\\ 
& Lei Zheng, Jingyuan Zhang, Bokai Cao, Philip S. Yu, and Ann B. Ragin. A novel ensemble approach on regionalized neural networks for brain disorder prediction. SAC, 2017.\\ 
& Chun-Ta Lu, Lifang He, Weixiang Shao, Bokai Cao, and Philip S. Yu. Multilinear factorization machines for multi-task multi-view learning. WSDM, 2017.\\ 
& Xiaokai Wei, Bokai Cao, Weixiang Shao, Chun-Ta Lu, and Philip S. Yu. Community detection with partially observable links and node attributes. Big Data, 2016.\\ 
& Bokai Cao, Chun-Ta Lu, Xiaokai Wei, Philip S. Yu, and Alex D. Leow. Semi-supervised tensor factorization for brain network analysis. ECML/PKDD, 2016.\\ 
& Guixiang Ma, Lifang He, Bokai Cao, Jiawei Zhang, and Philip S. Yu. Multi-graph clustering based on interior-node topology with applications to brain networks. ECML/PKDD, 2016.\\ 
& Xiaokai Wei, Bokai Cao, and Philip S. Yu. Nonlinear joint unsupervised feature selection. SDM, 2016.\\ 
& Jingyuan Zhang, Bokai Cao, Sihong Xie, Chun-Ta Lu, Philip S. Yu, and Ann B. Ragin. Identifying connectivity patterns for brain diseases via multi-side-view guided deep architectures. SDM, 2016.\\ 
& Bokai Cao, Hucheng Zhou, Guoqiang Li, and Philip S. Yu. Multi-view machines. WSDM, 2016.\\ 
& Xiaokai Wei, Bokai Cao, and Philip S. Yu. Unsupervised feature selection on networks: a generative view. AAAI, 2016.\\ 
& Bokai Cao, Xiangnan Kong, Jingyuan Zhang, Philip S. Yu, and Ann B. Ragin. Identifying HIV-induced subgraph patterns in brain networks with side information. Brain Informatics, 2015.\\ 
& Bokai Cao, Xiangnan Kong, and Philip S. Yu. A review of heterogeneous data mining for brain disorder identification. Brain Informatics, 2015.\\ 
& Bokai Cao, Xiangnan Kong, Casey Kettering, Philip S. Yu, and Ann B. Ragin. Determinants of HIV-induced brain changes in three different periods of the early clinical course: A data mining analysis. NeuroImage: Clinical, 2015.\\ 
& Bokai Cao, Xiangnan Kong, Jingyuan Zhang, Philip S. Yu, and Ann B. Ragin. Mining brain networks using multiple side views for neurological disorder identification. ICDM, 2015.\\ 
& Bokai Cao, Francine Chen, Dhiraj Joshi, and Philip S. Yu. Inferring crowd-sourced venues for tweets. Big Data, 2015.\\ 
& Bokai Cao, Liang Zhan, Xiangnan Kong, Philip S. Yu, Nathalie Vizueta, Lori L. Altshuler, and Alex D. Leow. Identification of discriminative subgraph patterns in fMRI brain networks in bipolar affective disorder. BIH, 2015.\\ 
& Bokai Cao, Lifang He, Xiangnan Kong, Philip S. Yu, Zhifeng Hao, and Ann B. Ragin. Tensor-based multi-view feature selection with applications to brain diseases. ICDM, 2014.\\ 
& Bokai Cao, Xiangnan Kong, and Philip S. Yu. Collective prediction of multiple types of links in heterogeneous information networks. ICDM, 2014.\\ 
& Xiangnan Kong, Bokai Cao, Philip S. Yu, Ying Ding, and David J. Wild. Meta path-based collective classification in heterogeneous information networks. arXiv, 2013.\\
& Xiangnan Kong, Bokai Cao, and Philip S. Yu. Multi-label classification by mining label and instance correlations from heterogeneous information networks. KDD, 2013.
\end{longtable}

\end{document}